\documentclass[final,3p,times,authoryear]{elsarticle}
\usepackage{hyperref}
\usepackage{url}
\usepackage{amsmath}
\usepackage{amssymb}
\usepackage{mathtools}
\usepackage{upgreek}
\usepackage{soul}
\usepackage{graphicx}
\usepackage{subcaption}
\usepackage{lineno}
\usepackage{subcaption}
\usepackage{tikz}
\usepackage{graphicx}
\usepackage{xcolor}
\newcommand{\Czero}[1]{C_0([0,T];{\mathbb{R}^#1})}

\newcommand{\sigmagamma}[1][]{\sigma_{\it{\Gamma}}\left(#1\right)}
\newcommand{\sigmapi}[1][]{\sigma_{\it{\Pi}}\left(#1\right)}

\newcommand{\Phigamma}[1]{\Phi_{\it{\Gamma}}\left[#1\right]}

\newcommand{\norminf}[1]{{\left\lVert #1 \right\rVert}_{L^{\infty}}}
\newcommand{\norminft}[2]{{\left\lVert #2 \right\rVert}_{L^{\infty}[0, #1]}}

\newcommand{\wGammain}{w_{{\it{\Gamma},{\mathrm{in}}}}}
\newcommand{\wGamma}{w_{\it{\Gamma}}}
\newcommand{\wGammaout}{w_{{\it{\Gamma},{\mathrm{out}}}}}

\newcommand{\wPiin}{w_{{\it{\Pi},{\mathrm{in}}}}}
\newcommand{\wPi}{w_{\it{\Pi}}}
\newcommand{\wPiout}{w_{{\it{\Pi},{\mathrm{out}}}}}
\newcommand{\hPi}{h_{\it{\Pi}}}

\newcommand{\itGamma}{{\it{\Gamma}}}
\newcommand{\itPi}{{\it{\Pi}}}

\DeclarePairedDelimiter{\floor}{\lfloor}{\rfloor}
\DeclarePairedDelimiter{\ceil}{\lceil}{\rceil}

\definecolor{mybackground}{RGB}{255, 255, 255}%%White

\pagecolor{mybackground}

\journal{Mechanical Systems and Signal Processing}

\begin{document}

\begin{frontmatter}



\title{Upper Approximation Bounds for Neural Oscillators} %% Article title

\author[label1,label2]{Zifeng Huang}
\ead{zifeng.huang@irz.uni-hannover.de}

\author[label3]{Konstantin M. Zuev}
\ead{kostia@caltech.edu}

\author[label2,label4]{Yong Xia\corref{cor1}}
\ead{ceyxia@polyu.edu.hk}

\author[label1,label5,label6]{Michael Beer}
\ead{beer@irz.uni-hannover.de}

%% Corresponding author footnote
\cortext[cor1]{Corresponding author.}

%% Affiliations
\affiliation[label1]{organization={Institute for Risk and Reliability, Leibniz University Hannover},
            addressline={Callinstraße 34},
            city={Hannover},
            postcode={30167},
            country={Germany}}

\affiliation[label2]{organization={Department of Civil and Environmental Engineering, The Hong Kong Polytechnic University},
            addressline={Kowloon},
            city={Hong Kong},
            country={China}}

\affiliation[label3]{organization={Department of Computing and Mathematical Sciences, California Institute of Technology},
            city={Pasadena},
            state={California},
            country={United States}}

\affiliation[label4]{organization={Guangdong-Hong Kong Joint Research Laboratory for Marine Infrastructure, The Hong Kong Polytechnic University},
            addressline={Kowloon},
            city={Hong Kong},
            country={China}}

\affiliation[label5]{organization={Department of Civil and Environmental Engineering, University of Liverpool},
            city={Liverpool},
            postcode={L69 3GH},
            country={United Kingdom}}

\affiliation[label6]{organization={International Joint Research Center for Resilient Infrastructure \& International Joint Research Center for Engineering Reliability and Stochastic Mechanics, Tongji University},
            city={Shanghai},
            postcode={200092},
            country={China}}

%% Abstract
\begin{abstract}
Neural oscillators, originating from second-order ordinary differential equations (ODEs), have demonstrated strong performance in stably learning causal mappings between long-term sequences or continuous temporal functions, as well as in accurately approximating physical systems. However, theoretically quantifying the capacities of their neural network architectures remains a significant challenge. In this study, the neural oscillator consisting of a second-order ODE followed by a multilayer perceptron (MLP) is considered. Its upper approximation bound for approximating causal and uniformly continuous operators between continuous temporal function spaces and that for approximating uniformly asymptotically incrementally stable second-order dynamical systems are derived. The established proof method of the approximation bound for approximating the causal continuous operators can also be directly applied to state-space models consisting of a linear time-continuous complex recurrent neural network followed by an MLP. Theoretical results reveal that the approximation error of the neural oscillator for approximating the second-order dynamical systems scales polynomially with the reciprocals of the widths of two utilized MLPs, thus overcoming the curse of parametric complexity. The {\color{black} convergence rates} of two established approximation error bounds are validated through {\color{black} four} numerical cases. These results provide a robust theoretical foundation for the effective application of the neural oscillator in science and engineering.
\end{abstract}


% \begin{highlights}
% \item This study establishes upper approximation bounds for a neural oscillator architecture in approximating causal and uniformly continuous operators and in approximating second-order dynamical systems.

% \item The approximation bound for the neural oscillator in approximating second-order dynamical systems scales polynomially with the reciprocals of the widths of two employed MLPs, thereby overcoming the curse of parametric complexity.

% \item The results established in this study provide a robust theoretical foundation for the effective application of the neural oscillator in science and engineering.
% \end{highlights}

%% Keywords
\begin{keyword}
Neural oscillator \sep
Upper approximation bounds \sep
Causal continuous operator \sep
Stable dynamical system

% %% PACS codes here, in the form: \PACS code \sep code

% %% MSC codes here, in the form: \MSC code \sep code
% %% or \MSC[2008] code \sep code (2000 is the default)
\end{keyword}

\end{frontmatter}

% \linenumbers


\section{Introduction}
\label{sec1}
Modeling the mapping relationships of long sequences or continuous temporal functions is a significant challenge in machine learning. A large variety of neural network architectures have been proposed to address sequential and functional mapping tasks, including recurrent neural networks (RNNs) \citep{rumelhart1986learning} and their variants \citep{hochreiter1997long,cho2014learning}, attention-based architectures \citep{vaswani2017attention}, state-space (SS) models based on the first-order ordinary differential equations (ODEs) \citep{gu2020hippo,gu2021efficiently,gu2022parameterization}, and neural oscillators based on the second-order ODEs \citep{rusch2020coupled,rusch2021unicornn,lanthaler2023neural,rusch2024oscillatory,zifeng2025}. SS models and neural oscillators possess the fast inference capability of RNNs together with the competitive performance of the attention-based architectures in long-range sequence learning \citep{gu2023mamba,rusch2024oscillatory}. Furthermore, neural oscillators have rigorously demonstrated to mitigate the problem of vanishing or exploding gradients \citep{rusch2020coupled,rusch2021unicornn}. They can provide stable dynamics in learning long-range interactions \citep{rusch2024oscillatory}. 

{\color{black} Regarding the applications of neural oscillators, \citet{rusch2024oscillatory} utilized two neural oscillator models to predict heart rates using real-world data from wrist-worn devices, and showed that the neural oscillators outperformed some SS models. \citet{stolzle2024input} employed an input-to-state stable neural oscillator to characterize the latent dynamic equation of an autoencoder framework, and  the proposed model was utilized to learn complex nonlinear dynamics of a continuum soft robot directly from raw pixel images. An integral-saturated proportional–integral–derivative controller with potential-energy-shaping feedforward compensation was developed to generate corrective actuation commands that drive the robot toward desired trajectories, using raw pixel images as the sole sensing modality. \citet{kapoor2024neural} employed the neural oscillators to enhance the 
generalization of physics-informed machine learning beyond the training 
domain. By exploiting the inherent causality and temporal sequential 
characteristics of partial differential equation (PDE) solutions, this work fused a physics-informed 
machine learning model with the neural oscillators to process the model outputs 
as sequential data and predict solutions in unseen temporal regions. 
The effectiveness of this framework was demonstrated across multiple 
time-dependent nonlinear PDEs. \citet{zifeng2025} numerically demonstrated that neural oscillators can directly learn the nonlinear derivative terms governing complex ODE systems.} A recent overview on SS models and neural oscillators was provided by \citet{tiezzi2025back}.

Despite the practical success of SS models and neural oscillators, advancing their theoretical understanding, particularly in quantifying the capacities of their network architectures, remains a significant challenge. For the universal approximation properties of SS models, \citet{JMLR:v23:21-0368} demonstrated that linear RNNs can universally approximate linear, causal, continuous, regular, and time-homogeneous operators between continuous temporal function spaces along with an approximation bound. Based on the result of \citet{JMLR:v23:21-0368} and the Volterra series \citep{boyd1984analytical}, \citet{wang2023state} proved that given sufficiently large model sizes, an SS model consisting of a time-continuous linear RNN followed by a polynomial nonlinear projection can approximate any time-homogeneous causal continuous operator between continuous temporal function spaces. Similarly, \citet{muca2024theoretical} demonstrated that a selective SS model comprising a linear controlled differential equation followed by a multilayer perceptron (MLP) can universally approximate causal continuous operators of continuous temporal functions. However, neither result provides a quantitative relationship between the approximation error and the number of parameters required in the SS models. \citet{orvieto2024universality} derived an upper approximation bound for an SS model consisting of a {\color{black} discrete-time} linear RNN followed by an MLP. However, this result is only limited to the mapping relationships between finite-length sequences and does not apply to infinite-length sequences or continuous temporal functions. In addition to above results, the approximation property of the original RNN for approximating the first-order dynamical systems \citep{hanson2020universal} and {\color{black} that} for approximating fading memory causal operators of {\color{black} discrete-time} sequences \citep{grigoryeva2018echo,grigoryeva2018universal,gonon2023approximation} have been studied.

For {\color{black} the} neural oscillators, a fundamental lemma proposed by \citet{lanthaler2023neural} provides a novel insight that the time-varying sine transform coefficients of {\color{black} time-continuous input functions} can be used to construct a universal neural network architecture for the causal continuous operators between continuous temporal function spaces. Based on this fundamental lemma, the neural oscillator architecture consisting of a second-order ODE followed by an MLP was proven to be a universal approximator for such causal continuous operators \citep{lanthaler2023neural,rusch2024oscillatory}. However, these theoretical results are only qualitative, and the quantitative approximation bounds of {\color{black} the} neural oscillators remain undeveloped.

In this study, the neural oscillator consisting of a second-order ODE followed by an MLP is considered. Its upper approximation bound for approximating causal and uniformly continuous operators between continuous temporal function spaces and that for approximating uniformly asymptotically incrementally stable second-order dynamical systems are derived. In Section 2, the network architecture of the neural oscillator is briefly introduced, and the newly established theoretical approximation bounds of the neural oscillator are presented informally to facilitate understanding. The formal statements of two theorems that establish the approximation bounds for the neural oscillator, accompanied by the supporting lemmas essential for their proofs, are provided in Section 3. In Section 4, the {\color{black} convergence rates} of the two established approximation error bounds are validated through {\color{black} four} numerical cases. The main contributions of this study are summarized as follows:
\begin{enumerate}
    \item In Section \ref{subsec3.1}, Lemma 1 establishes an error bound for approximating continuous temporal functions using time-varying sine transform coefficients. Theorem 1 derives an upper approximation bound for the neural oscillator in approximating causal and uniformly continuous operators between continuous temporal function spaces. The proof method of Theorem 1 established in this study can also be directly employed to derive the approximation bound for the SS models consisting of a linear time-continuous complex RNN followed by an MLP.
    
    \item In Section \ref{subsec3.2}, Theorem 2 establishes an upper approximation bound for the neural oscillator in approximating the uniformly asymptotically incrementally stable second-order dynamical systems. The approximation error scales polynomially with the reciprocals of the widths of two utilized MLPs. This theorem provides a robust theoretical foundation for the effective application of the neural oscillator in science and engineering.
    
    \item During the mathematical derivation in Section \ref{subsec3.2}, Lemma 4 enhances the proof of a fundamental Lemma 3.1.2 on differential equation flow by \citet{van2007filtering}. The results in Lemma 6 and \ref{appD} demonstrate that two methods, one based on the differential equation flow and the other one on the Grönwall's inequality \citep{ames1997inequalities}, are equivalent in establishing the bound of the difference between the solutions of two second-order ODEs driven by the same input functions. The proof of Theorem 2 establishes a new method to ensure that all possible solutions of a neural ODE, whose MLP is employed to approximate the derivative {\color{black} term} function of a target ODE, remain within a domain where the error between the MLP of the neural ODE and the derivative {\color{black} term} function of the target ODE can be quantified.
\end{enumerate}

{\bf Notations:} In this study, the $\it{L^r}$-norm of an \textit{n}-dimensional vector $\mathbf{a} = [a_1, a_2,..., a_n]^\top, 1 \leq r \leq +\infty$, is $|\mathbf{a}|_r = [\sum_{i=1}^n |a_i|^r]^{1/r}$ for $r < \infty$, or $|\mathbf{a}|_r = \max_{1 \leq i \leq n}|a_i|$ for $r = +\infty$. $|\mathbf{a}|$ is short for $|\mathbf{a}|_1$. For a matrix $\mathbf{M} = [m_{ij}]_{p \times q}$, its $L^{r,s}$-norm, $1 \leq r,s \leq +\infty$, is $\lvert\mathbf{M}\rvert_{r,s} = \lvert\left[\lvert\mathbf{m}_{1}\rvert_{r}, \lvert\mathbf{m}_{2}\rvert_{r},\dots,\lvert\mathbf{m}_{q}\rvert_{r}\right]^\top\rvert_{s}$, where $\mathbf{m}_{j}$ is the $j^\text{th}$ column vector of $\mathbf{M}$. The $L^r$-norm of $\mathbf{M}$ is $|\mathbf{M}|_r = \sup_{|\mathbf{a}|_r = 1}|\mathbf{Ma}|_r/|\mathbf{a}|_r$. Re($\cdot$) and Im($\cdot$) represent the real and imaginary parts of a complex number, respectively. $\floor{\cdot}$ and $\ceil{\cdot}$ represent the floor and ceiling functions, respectively. $\mathbb{Z}$ represents the integer space. $\mathbb{R}$ represents the real number space. $\mathbb{R}^{p{\times}q}$ represents the $p \times q$-dimensional real matrix space. $f'(t) = \mathrm{d}f(t)/\mathrm{d}t$, $f''(t) = \mathrm{d}^{2}f(t)/\mathrm{d}t^{2}$, and $f'''(t) = \mathrm{d}^{3}f(t)/\mathrm{d}t^{3}$. $C_0([0, T]; \mathbb{R}^p)$ represents a \textit{p}-dimensional vector-valued real continuous function space defined over $[0, T]$ {\color{black} under the} $L^{\infty}$-norm ${\lVert \cdot \rVert}_{L^{\infty}}$, e.g., ${\lVert\mathbf{u}(t) \rVert}_{L^{\infty}}$ $= \sup_{t\in[0, T]}|\mathbf{u}(t)|$ for $\mathbf{u}(t) = [u_1(t), u_2(t),..., u_p(t)]^\top \in \Czero{p}$. In addition, {\color{black} the} $L^\infty{[0, t_0]}$-norm ${\lVert \cdot \rVert}_{L^{\infty}[0, t_0]}$ is ${\lVert \mathbf{u}(t) \rVert}_{L^{\infty}[0, t_0]} = \sup_{t\in [0, t_0]}|\mathbf{u}(t)|$ and {\color{black} the} $L^2$-norm ${\lVert \cdot \rVert}_{L^2}$ is ${\lVert \mathbf{u}(t) \rVert}_{L^2} = \sqrt{\int_{0}^{T}|\mathbf{u}(t)|_{2}^{2}\mathrm{d}t}$. $f(n) = O\left[g(n)\right]$ if there exist positive constants $c$ and $n_0$ such that $f(n) \leq c \cdot g(n)$ for all $n \geq n_0$. {\color{black} $f(n) = {\it{\Omega}}\left[g(n)\right]$ if there exist positive constants $c$ and $n_0$ such that $f(n) \geq c \cdot g(n)$ for all $n \geq n_0$.}

\section{Neural Oscillator}
\label{sec2}
{\color{black} For a given} $\mathbf{u}(t) = [u_1(t), u_2(t),..., u_p(t)]^{\top} \in C_{0}([0, T],\mathbb{R}^p)$, a neural oscillator architecture mapping $\mathbf{u}(t)$ to $\mathbf{y}(t) = [y_1(t), y_2(t),..., y_q(t)]^{\top}$ over $t \in [0, T]$ is
\begin{equation}
    \left\{
    \begin{aligned}
    &\mathbf{x}''(t) = {\it{\Gamma}}[\mathbf{x}(t), \mathbf{x}'(t), \mathbf{u}(t)] \\
    &\mathbf{y}(t) = {\it{\Pi}}[\mathbf{x}(t), \mathbf{u}(0), t]
    \end{aligned}
     \right.
    \label{eq:1}
\end{equation} with $\mathbf{x}(0) = \mathbf{x}'(0) = \mathbf0$, where ${\it{\Gamma}}(\cdot)$ and ${\it{\Pi}}(\cdot)$ are two MLPs
\begin{equation}
   {\it{\Gamma}}\left[ {{\bf{x}}(t),{\bf{x'}}(t),{\bf{u}}(t)} \right] = {{\bf{W}}_2}{\sigma _{\it{\Gamma}}}\left\{ {{{\bf{W}}_1}{{\left[ {{{\bf{x}}^{\top}}(t),{{{\bf{x'}}}^{\top}}(t),{{\bf{u}}^{\top}}(t)} \right]}^{\top}} + {{\bf{b}}_1}} \right\} + {{\bf{b}}_2},
   \label{eq:2}
\end{equation}
\begin{equation}
   {\it{\Pi}}\left[ {{\bf{x}}(t),{\bf{u}}(0),t} \right] = {{\bf{\tilde W}}_{{h_{\it{\Pi}} }}}{\sigma _{\it{\Pi}}}\left( { \ldots {\sigma _{\it{\Pi}} }\left\{ {{{{\bf{\tilde W}}}_1}{{\left[ {{{\bf{x}}^{\top}}(t),{{\bf{u}}^{\top}}(0),t} \right]}^{\top}} + {{{\bf{\tilde b}}}_1}} \right\}} \right) + {{\bf{\tilde b}}_{{h_{\it{\Pi}}}}},
   \label{eq:3}
\end{equation}
$\mathbf{x}(t) = [x_1(t), x_2(t),..., x_r(t)]^\top$ is an intermediate \textit{r}-dimensional real-valued state function, $\mathbf{W}_i$, \textit{i} = 1 and 2, $\tilde{\mathbf{W}}_j$, \textit{j} = 1, 2,..., $h_{\it{\Pi}}$, are real-valued {\color{black} trainable} weight matrices, $\mathbf{b}_i$ and $\tilde{\mathbf{b}}_j$ are real-valued {\color{black} trainable} bias vectors, $h_{\it{\Pi}}$ is the depth of ${\it{\Pi}}(\cdot)$, and $\sigmagamma[\cdot]$ and $\sigmapi[\cdot]$ are the activation functions of ${\it{\Gamma}}(\cdot)$ and ${\it{\Pi}}(\cdot)$, respectively. The neural oscillator architecture in Eq.~\eqref{eq:1} can be denoted as $\mathbf{y}(t) = {\it{\Pi}}\circ{\Phi}_{\it{\Gamma}}[\mathbf{u}(\tau)](t) = {\it{\Pi}}\left\{{{\Phi}_{\it{\Gamma}}[{\mathbf{u}(\tau)}](t),\mathbf{u}(0),t}\right\}$, where $\circ$ is the composition operator and $\mathbf{x}(t) = \Phigamma{\mathbf{u}(\tau)}(t)$ represents the casual operator characterized by the second-order ODE in Eq.~\eqref{eq:1}.

The initial value \textbf{u}(0) added in ${\it{\Pi}}(\cdot)$ eliminates the need for the assumption of $\mathbf{u}(0) = \mathbf{0}$ in the previous universal approximation theorems for {\color{black} the} neural oscillators, as presented in \citet{lanthaler2023neural} and \citet{rusch2024oscillatory}. The influence of time in non-autonomous causal operators is characterized by a term $0.25t^2$ in \citet{lanthaler2023neural}, which arises as the solution to a second-order ODE $f''(t) = 0.5$. However, this has the possibility of causing unbounded derivatives of the MLP ${\it{\Pi}}(\cdot)$. For example, if ${\it{\Pi}}(t^2) = g(t)$ and $g'(t)$ is continuous and bounded over $t \in [0,T]$ with $g'(0) \neq 0$, then $g'(t)=2t{\it{\Pi}}'(t^2)$ and ${\it{\Pi}}'(x)$ tends to infinity as \textit{x} approaches zero. The potential unbounded derivatives of ${\it{\Pi}}(\cdot)$ may induce challenges for subsequent theoretical analysis. To address this problem, in this study, a term \textit{t} is directly added to ${\it{\Pi}}(\cdot)$ to ensure that the derivative of ${\it{\Pi}}(\cdot)$ with respect to \textit{t} can be bounded.

\begin{figure}[t]
    \centering
    \includegraphics[width=0.6\linewidth]{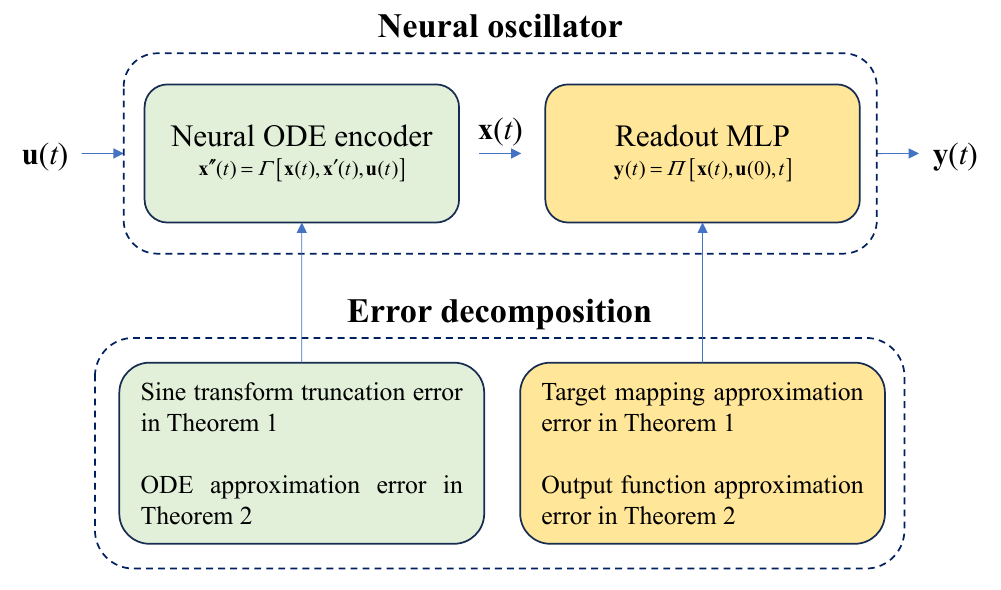}
    
    \caption{{\color{black}The two-module structure and error propagation pathways of the neural oscillator.}}
    \label{fig:1}
\end{figure}

{\color{black} As illustrated in Figure \ref{fig:1}, the neural oscillator is a two-module structure. The first module is a second-order neural ODE encoder characterized by $\itGamma(\cdot)$ that encodes the input function $\mathbf{u}(t)$ into the intermediate state function $\mathbf{x}(t)$. The second module is a readout MLP $\itPi(\cdot)$ that maps  $\mathbf{x}(t)$, together with the initial value $\mathbf{u}(0)$ of $\mathbf{u}(t)$ and time \textit{t}, to the output function $\mathbf{y}(t)$. As established in Theorem 1 of this study, when $\sigma_{\it{\Gamma}}(\cdot)$ is the Rectified Linear Unit (ReLU) $\sigma_{\mathrm{ReLU}}(x) = \mathrm{max}(0,x)$ \citep{nair2010rectified}, the neural oscillator can function as a universal approximator of causal continuous operators. In Theorem 1, the total approximation error is decomposed into two sequential errors. The first arises from truncating the sine transform coefficients of $\mathbf{u}(t)$, and the second stems from approximating the target mapping from $\mathbf{u}(0)$, \textit{t}, and the sine transform coefficients of $\mathbf{u}(t)$ to $\mathbf{y}(t)$ by $\itPi(\cdot)$. Moreover, for the second-order dynamical systems in science and engineering \citep{strogatz2024nonlinear}, the neural oscillator can leverage its MLPs $\itGamma(\cdot)$ and $\itPi(\cdot)$ to directly approximate the derivative term and output functions of these systems with efficient neural network sizes, as demonstrated in Theorem 2 of this study. In this situation, the total approximation error again consists of two sequential errors. The first is induced by approximating a target second-order ODE by the neural ODE characterized by $\itGamma(\cdot)$, and the second is caused by approximating a target output function by $\itPi(\cdot)$.}

{\color{black} Traditional neural ODE models, including the first-order and second-order neural ODEs \citep{chen2018neural,norcliffe2020second}, leverage the relationships between their initial conditions and solutions at specific time instants to universally approximate finite-dimensional continuous functions \citep{celledoni2023dynamical}. The neural oscillator augments a second-order neural ODE with a readout MLP. It acts as a universal approximator for causal continuous operators by first encoding input functions into a finite set of sine transform coefficients, and subsequently mapping these coefficients to output functions. This universal approximation property is similar to those of neural operators \citep{kovachki2023neural}.}

In the following content of this section, the underlying assumptions and problem setup are outlined. Subsequently, the established theoretical approximation bounds of the neural oscillator in Eq.~\eqref{eq:1} are presented in an informal manner. The formal statements of the derived theoretical results are presented in Theorems 1 and 2 in Section \ref{sec3}, together with the supporting lemmas required for their proofs.

{\bf Assumption 1:} $C_0([0,T];{\mathbb{R}}^p)$ with {\color{black} the} $L^\infty$-norm is a Banach space. Its metric is defined as ${\lVert{\mathbf{u}_1(t)-\mathbf{u}_2(t)}\rVert}_{L^{\infty}}$ with $\mathbf{u}_1(t)$ and $\mathbf{u}_2(t)$ $\in$ $C_0([0,T];\mathbb{R}^p)$.

{\bf Assumption 2:} \textit{K} is a compact subset of ${C_0([0,T];\mathbb{R}^p)}$ and it is closed. By virtue of the Arzelà-Ascoli theorem \citep{beattie2013convergence}, all functions belonging to \textit{K} are uniformly bounded. There exists a constant $B_K$ such that all elements of $K$ are bounded by $B_K$, that is ${\lVert{u_i(t)}\rVert}_{L^\infty} \leq B_K$ for all $\mathbf{u}(t) \in K$ and $1 \leq i \leq p$, where $u_i(t)$ is the $i^\mathrm{th}$ element of $\mathbf{u}(t)$. It follows from the Arzelà-Ascoli theorem that the functions belonging to $K$ are equicontinuous, and there exists a continuous modulus of continuity $\mathbf{\upphi}_K(t) = [\phi_{K,1}(t),\phi_{K,2}(t),...,\phi_{K,p}(t)]^{\top}$ such that for arbitrary $0 \leq t_1 \leq t_2 \leq T$ and all $\mathbf{u}(t) \in K$, {\color{black} it is satisfied that} $|u_i(t_1) - u_i(t_2)|\leq\phi_{K,i}(t_2-t_1)$, thereby $|\mathbf{u}(t_1) - \mathbf{u}(t_2)|\leq \left|\mathbf{\upphi}_{K}(t_2-t_1)\right|$, where $\phi_{K,i}(t): [0,+\infty) \to [0,+\infty)$ is monotonic with $\phi_{K,i}(t) \to 0$ as $t \to 0$.

{\bf Assumption 3:} ${{\Phi}} = [\Phi_1,\Phi_2...., \Phi_q]: C_0([0,T];{\mathbb{R}}^p) \to C_0([0,t];{\mathbb{R}}^q)$ is a causal and uniformly continuous operator. For $\forall{t} \in [0,T]$, $\Phi[\mathbf{u}(\tau)](t)$ only depends on $\mathbf{u}(\tau)$ over $\tau \in [0,t]$, where $\mathbf{u}(t) \in C_0([0,T];{\mathbb{R}}^p)$. For $\forall{\varepsilon} > 0$, there exists a distance $d_{\varepsilon} > 0$, such that ${\lVert{\mathbf{u}_1(t)-\mathbf{u}_2(t)}\rVert}_{L^{\infty}} < d_{\varepsilon} \Rightarrow {\lVert{\Phi[\mathbf{u}_1(\tau)](t) - \Phi[\mathbf{u}_2(\tau)](t)}\rVert}_{L^{\infty}} < \varepsilon$ for arbitrary $\mathbf{u}_1(t)$ and $\mathbf{u}_2(t)$ $\in C_0([0,t];{\mathbb{R}}^p)$.

{\bf Assumption 4:} Since ${{\Phi}} = [\Phi_1,\Phi_2...., \Phi_q]$ is a causal and uniformly continuous operator mapping from $C_0([0,T];{\mathbb{R}}^p)$ to $C_0([0,T];{\mathbb{R}}^q)$ and $K$ is a compact subset of $C_0([0,t];{\mathbb{R}}^p)$, the image set $\Phi(K)$ of $K$ by $\Phi$ is a compact subset of $C_0([0,T];{\mathbb{R}}^q)$. There exists a constant $B_{\Phi(K)}$ such that all elements of $\Phi(K)$ are bounded by $B_{\Phi(K)}$, that is ${\lVert{v_j(t)}\rVert}_{L^\infty} \leq B_{\Phi(K)}$ for all $\mathbf{v}(t) = [v_1(t), v_2(t),..., v_q(t)]^\top \in \Phi(K)$ and $1 \leq j \leq q$.

{\bf Assumption 5:} For the causal and uniformly continuous operator ${{\Phi}} = [\Phi_1,\Phi_2...., \Phi_q]$ mapping from $C_0([0,T];{\mathbb{R}}^p)$ to $C_0([0,T];{\mathbb{R}}^q)$, for each $\Phi_j$, it is assumed that the first-order derivatives of all temporal functions in the image set $\Phi_j\left\{ \Czero{p} \right\}$ of $\Czero{p}$ by $\Phi_j$ are bounded by a finite constant $D_{\Phi_j}$. 

{\bf Assumption 6:} Since ${{\Phi}} = [\Phi_1,\Phi_2...., \Phi_q]$ is a causal and uniformly continuous operator mapping from $C_0([0,T];{\mathbb{R}}^p)$ to $C_0([0,T];{\mathbb{R}}^q)$, its $j^{\mathrm{th}}$ element $\Phi_j$ has a monotonic continuous modulus of continuity $\phi_{\Phi_j}(\cdot): [0,+\infty) \to [0, +\infty)$ with $\phi_{\Phi_j}(r) \to 0$ as $r \to 0$, such that $\norminf{\Phi_j[\mathbf{u}_1(\tau)](t) - \Phi_j[\mathbf{u}_2(\tau)](t)}$ $\leq \phi_{\Phi_j}[\norminf{\mathbf{u}_1(t) - \mathbf{u}_2(t)}]$ for arbitrary $\mathbf{u}_1(t)$ and $\mathbf{u}_2(t)$ $\in \Czero{p}$. $\phi_{\Phi}(\cdot)$ $=\sum_{j=1}^{q}{\phi_{\Phi_j}(\cdot)}: [0,+\infty) \to [0,+\infty)$ with $\phi_{\Phi}(r) \to 0$ as $r \to 0$ is a monotonic modulus of continuity of $\Phi$, such that $\norminf{\Phi[\mathbf{u}_1(\tau)](t) - \Phi[\mathbf{u}_2(\tau)](t)} \leq \sum_{j=1}^{q}\norminf{\Phi_j[\mathbf{u}_1(\tau)](t) - \Phi_j[\mathbf{u}_2(\tau)](t)} \leq \phi_{\Phi}[\norminf{\mathbf{u}_1(t) - \mathbf{u}_2(t)}]$ for arbitrary $\mathbf{u}_1(t)$ and $\mathbf{u}_2(t)$ $\in \Czero{p}$.

{\bf Setting 1:} Given two real continuous function spaces $\Czero{p}$ and $\Czero{q}$ with {\color{black} the} $L^{\infty}$-norm $\norminf{\cdot}$, a compact set $K \subset \Czero{p}$, and a causal and uniformly continuous operator $\Phi:\Czero{p} \to \Czero{q}$, for each $\mathbf{u}(t) \in K$, its corresponding image function $\mathbf{v}(t) \in \Czero{q}$ is calculated by $\Phi$
\begin{equation}
   {\bf{v}}(t) = \Phi \left[ {{\bf{u}}(\tau )} \right](t) = {\left\{ {{\Phi _1}\left[ {{\bf{u}}(\tau )} \right](t),{\Phi _2}\left[ {{\bf{u}}(\tau )} \right](t), \ldots ,{\Phi _q}\left[ {{\bf{u}}(\tau )} \right](t)} \right\}^\top}.
   \label{eq:4}
\end{equation}

{\bf Informal Statement of Theorem 1:} {\it Given a compact subset $K \subset \Czero{p}$ and a causal and uniformly continuous operator $\Phi: \Czero{p}$ $\to \Czero{q}$ satisfying Assumptions 1 to 6 with $\mathrm{\upphi}_K(t) = L_K[t, t,\dots,t]^\top$ and $L_K > 0$, for every integer $M_{\it{\Gamma}}$ {\color{black} implicitly} larger than an independent deterministic threshold and every positive integer $H_{\it{\Pi}}$, there exist two MLPs $\it{\Gamma}(\cdot)$ and $\it{\Pi}(\cdot)$ employing $\sigma_{\mathrm{ReLU}}(\cdot)$, such that for an arbitrary $\mathbf{u}(t) \in K$, the corresponding solution $\mathbf{y}(t) = [y_1(t), y_2(t),...,y_q(t)]^{\top}$ from Eq.~\eqref{eq:1} subject to $\mathbf{u}(t)$ with the initial conditions $\mathbf{x}(0) = {\mathbf{x}}'(0) = \mathbf{0}$ satisfies
\begin{equation}
    \left| {\Phi \left[ {{\bf{u}}(\tau )} \right](t) - {\bf{y}}(t)} \right| \le {\varepsilon _{\bf{y}}}
    \label{eq:5}
\end{equation}
for all $t \in [0,T]$, where
{\color{black} \begin{equation}
    {\varepsilon _{\bf{y}}} = {\phi _\Phi}\left\{O\left[ \frac{p{L_K}T\left(\ln{M_\itGamma}\right)^{2}}{M_{\it{\Gamma}}}\right]\right\} + O\left\{\frac{qp^{3}M_{\it{\Gamma}}^5}{H_{\it{\Pi}}^{{1 \mathord{\left/
    {\vphantom {1 {\left[ {p\left( {{M_{\it{\Gamma}} } + 1} \right) + 1} \right]}}} \right.
    \kern-\nulldelimiterspace} {\left[ {p\left( {{M_{\it{\Gamma}} } + 1} \right) + 1} \right]}}}}\right\}
    \label{eq:6}
\end{equation}}
and ${\phi_{\Phi}}(\cdot)$ is the modulus of continuity of $\Phi$ in Assumption 6. $\it{\Gamma}(\cdot)$ has one hidden layer and the widths of its input, hidden, and output layers are {\color{black} $\wGammain = 2pM_{\itGamma}+p$, $\wGamma = 2pM_{\itGamma}$, and $\wGammaout = pM_{\itGamma}$}, respectively. For ${\it{\Pi}}(\cdot)$, the widths of its input, hidden, and output layers are {\color{black} $\wPiin = pM_{\itGamma}+p+1$, $\wPi = q\left(pM_{\itGamma}+p+4 \right)$}, and $\wPiout = q$, respectively, and its depth is $\hPi = H_{\it{\Pi}} + 1$.} 

{\bf Informal Statement of Theorem 2:} {\it Let K be a compact subset of $\Czero{p}$, a second-order dynamical system
\begin{equation}
    \left\{
    \begin{aligned}
        &{\hat{\bf x}}''(t) = g\left[ {{\hat{\bf x}}(t),{\hat{\bf x}}'(t),{\bf{u}}(t)} \right]\\
        &{\hat{\bf y}}(t) = h\left[ {{\hat{\bf x}}(t)} \right]
    \end{aligned} 
    \right.,
    \label{eq:7}
\end{equation}
with ${\hat{\mathbf{x}}}(0) = {{\hat{\mathbf{x}}}}'(0) = \mathbf{0}$, is uniformly asymptotically incrementally stable for all $\mathbf{u}(t) \in K$ on a domain $\left[-B_{\mathbf{x}}, B_{\mathbf{x}}\right]^{r} \times \left[-B_{{\mathbf{x}}'}, B_{{\mathbf{x}}'}\right]^{r} \times \left[-B_K, B_K\right]^{p} \subset \mathbb{R}^{2r+p}$, where ${\hat{\mathbf{x}}}(t) = \left[{\hat{x}}_1(t),{\hat{x}}_2(t),...,{\hat{x}}_r(t)\right]^{\top}$, ${\hat{\mathbf{y}}}(t) = \left[{\hat{y}}_1(t),{\hat{y}}_2(t),...,{\hat{y}}_q(t)\right]^{\top}$, $B_{{\mathbf{x}}} = \alpha B_{\beta{g}} + B_{\hat{\mathbf{x}}}$, $B_{{\mathbf{x}
}'} = \alpha B_{\beta{g}} + B_{\hat{\mathbf{x}}'}$, $\alpha$ is a positive coefficient, $B_{\hat{\mathbf{x}}}$ and $B_{\hat{\mathbf{x}}'}$ are the bounds of ${\hat{\mathbf{x}}}(t)$ and ${\hat{\mathbf{x}}}'(t)$, respectively, $B_K$ is the bound of all $\mathbf{u}(t) \in K$ in Assumption 2, and $B_{\beta{g}}$ is the stability bound of the system in Eq.~\eqref{eq:7}. {\color{black} $g(\cdot) = \left[g_1(\cdot), g_2(\cdot),\dots,g_r(\cdot)\right]^{\top}$ satisfies that for every compact domain $\Omega \subset $ ${\mathbb{R}}^{2r+p}$, there exists a Barron function extension $\tilde{g}_{\Omega}(\cdot) = g(\cdot)$ on $\Omega$ and the spectral Barron norms of all extensions are uniformly bounded. $h(\cdot) = \left[h_1(\cdot), h_2(\cdot),\dots,h_q(\cdot)\right]^{\top}$ satisfies that for every compact domain $\Omega \subset $ ${\mathbb{R}}^{r}$, there exists a Barron function extension $\tilde{h}_{\Omega}(\cdot) = h(\cdot)$ on $\Omega$ and the spectral Barron norms of all extensions are uniformly bounded.} Then, for arbitrary positive errors $\varepsilon_1 \leq \alpha/r$ and $\varepsilon_2$, there exist two one-hidden-layer MLPs ${\it{\Gamma}}(\cdot)$ and ${\it{\Pi}}(\cdot)$ employing $\sigma_{\mathrm{ReLU}}(\cdot)$, such that for all $\mathbf{u}(t) \in K$, the corresponding solutions $\mathbf{y}(t)$ to Eq.~\eqref{eq:1} and ${\hat{\mathbf{y}}}(t)$ to Eq.~\eqref{eq:7} satisfy
\begin{equation}
    \left|{{\bf{y}}(t) - {\hat{\bf y}}(t)} \right| \le {L_h}{B_{{\beta _g}}}\sqrt r {\varepsilon _1} + q{\varepsilon _2}
    \label{eq:8}
\end{equation}
for all $t \in [0,T]$, where $L_h$ is the Lipschitz constant of $h(\cdot)$. For ${\it{\Gamma}}(\cdot)$, the widths of its input, hidden, and output layers are $\wGammain = 2r + p$, $\wGamma = O\left(r \varepsilon_{1}^{-2}\right)$, and $\wGammaout = r$, respectively. For ${\it{\Pi}}(\cdot)$, the widths of its input, hidden, and output layers are $\wPiin = r + p + 1$, $\wPi = O\left(q \varepsilon_{2}^{-2}\right)$, and $\wPiout = q$, respectively.}

{\color{black} In the two theorems, the convergence norms are the $L^{\infty}$-norm. The activation functions $\sigmagamma[\cdot]$ of $\itGamma(\cdot)$ and $\sigmapi[\cdot]$ of $\itPi(\cdot)$ are $\sigma_{\mathrm{ReLU}}(\cdot)$. In the numerical studies in Section~\ref{sec4}, $\sigma_{\mathrm{ReLU}}(\cdot)$ is adopted for $\itGamma(\cdot)$ and single-hidden-layer $\itPi(\cdot)$. For multi-hidden-layer $\itPi(\cdot)$, the Parametric Rectified Linear Unit (PReLU) $\sigma_{\text{PReLU}}(x)=\max(0,x) + \alpha\min(0,x)$, where $\alpha$ is a parameter learned during training \citep{he2015delving}, is employed to improve the training process.}

\section{Upper Approximation Bounds of the Neural Oscillator}
\label{sec3}
In this section, an upper approximation bound of the neural oscillator in Eq.~\eqref{eq:1} for approximating causal and uniformly continuous operators between continuous temporal function spaces and that for approximating uniformly asymptotically incrementally stable second-order dynamical systems are established in Subsections~\ref{subsec3.1} and \ref{subsec3.2}, respectively.

\subsection{Causal Continuous Operators}
\label{subsec3.1}
In this subsection, Lemma 1 derives an error bound for approximating all $\mathbf{u}(t) \in K$ using their initial values $\mathbf{u}(0)$ and a finite number of time-varying sine transform coefficients, combined with a set of corresponding deterministic base functions. Lemma 2 proves the existence of an MLP ${\it{\Gamma}}(\cdot)$ with $\sigma_{\mathrm{ReLU}}(\cdot)$, such that the solution $\mathbf{x}(t)$ {\color{black} of the neural} ODE in Eq.~\eqref{eq:1} corresponds to the sine transform coefficient vector of the input function $\mathbf{u}(t)$. Lemma 3 demonstrates that the target causal continuous operator $\Phi:\Czero{p} \to \Czero{q}$ in Setting 1 can be approximated by a continuous function $\Psi\left[\mathbf{x}(t),\mathbf{u}(0),t\right]$, which depends on time $t$, $\mathbf{u}(0)$, and the solution $\mathbf{x}(t)$, and establishes an associated error
bound. Lemma 4 determines the Lipschitz constants, linear moduli of continuity, and a compact argument domain of $\Psi\left[\mathbf{x}(t),\mathbf{u}(0),t\right]$. Finally, by approximating $\Psi\left[\mathbf{x}(t),\mathbf{u}(0),t\right]$ with an MLP ${\it{\Pi}}(\cdot)$ in Eq.~\eqref{eq:3} and combing its approximation error with the error bound from Lemma 3, Theorem 1 establishes an approximation bound for the neural oscillator in Eq.~\eqref{eq:1}. The proofs for all lemmas and Theorem 1 in this subsection are presented in \ref{appA}.

{\bf Lemma 1:} {\it Given a compact subset $K \subset \Czero{p}$ satisfying Assumptions 1 and 2, for an arbitrary $v$ satisfying $0 < v < T$ and every positive integer M {\color{black} larger than two}, there exist \textit{M} frequencies $\omega_1, \omega_2,..., \omega_M \in \mathbb{R}$, \textit{M} phases $\theta_1, \theta_2,..., \theta_M \in \mathbb{R}$, and \textit{M} weights $\alpha_1, \alpha_2,..., \alpha_M \in \mathbb{R}$, such that 
\begin{equation}
   \mathop {\sup }\limits_{\tau  \in [0,t]} \left| {{\bf{u}}(t - \tau ) - \left( {{\bf{u}}(0) + \sum\limits_{n = 1}^M {{\alpha _n}\left\{ {{\mathcal{L}_t}{\bf{u}}({\omega _n}) + \frac{{{\bf{u}}(0)}}{{{\omega _n}}}\left[ {\cos ({\omega _n}t) - 1} \right]} \right\}\sin ({\omega _n}\tau  - {\theta _n})} } \right)} \right| \le {\varepsilon _K}
   \label{eq:9}
\end{equation} 
for all $t \in [0, T]$ and all $\mathbf{u}(t) \in K$, where
\begin{equation}
   {\mathcal{L}_t}{\bf{u}}(\omega ) = \int_0^t {{\bf{u}}(t - \tau )\sin (\omega \tau ){\rm{d}}\tau},
   \label{eq:10}
\end{equation}
{\color{black}\begin{equation}
   {\varepsilon _K} = \left|{\mathbf{\upphi}_K(v)}\right| + \frac{c_K^{r+1}T^r\left|{\mathbf{\upphi}_K(T)}\right|\left(r!\right)^2\ln{M}}{v^rM^r},
   \label{eq:11}
\end{equation}}
{\color{black} $c_K > 1$ is an independent constant, $r$ can be an arbitrary positive integer}, and $\mathbf{\upphi}_K(t)$ is a modulus of continuity for all $\mathbf{u}(t) \in K$ in Assumption 2.}

{\bf Remark 1:} Lemma 1 is an enhanced version of Lemma B.1 in \citet{lanthaler2023neural} by providing a precise error bound and incorporating the influence of the initial value $\mathbf{u}(0)$. The initial value $\mathbf{u}(0)$ in the original Lemma B.1 is assumed to be zero. To address this limitation, a warm-up phase was introduced in \citet{lanthaler2023neural} to linearly extend $\mathbf{u}(t)$ from $t = 0$ to a negative time instant $t = -t_0$, $t_0 > 0$. However, this approach may introduce some practical considerations in selecting an appropriate value for $t_0$. 

{\color{black} {\bf Remark 2:} In Lemma 1, in addition to the positive integer \textit{M}, there exists an additional free parameter \textit{v} arising from the discontinuities of $\tilde{\mathbf{f}}_t(\tau)$ in Eq.~\eqref{eq:a1} in \ref{appA.1} and its derivatives at $\tau = 0$ and $\tau = \pm t$. Since $\mathbf{\upphi}_{K}(t)$ increases monotonically, \textit{v} can be eliminated by minimizing $\varepsilon_K$. For example, if $\mathbf{\upphi}_{K}(t) = [L_Kt, L_Kt,..., L_Kt]^\top$, where $L_K$ is a positive Lipschitz constant of all $\mathbf{u}(t) \in K$, then, $\varepsilon_K$ is calculated as
\begin{equation}
    {\varepsilon _K} = p{L_K}v + \frac{c_K^{r+1}pL_kT^{r+1}\left(r!\right)^2\ln{M}}{v^rM^r}.
    \label{eq:12}
\end{equation}
The minimum of $\varepsilon_K$ in Eq.~\eqref{eq:12} is

\begin{equation}
    \begin{aligned}
        & {\varepsilon _K} = p{L_K}\left[r^{1/(r+1)}+r^{-r/(r+1)} \right]\frac{c_KT\left(r!\right)^{2/(r+1)}\left(\ln{M}\right)^{1/(r+1)}}{M^{r/(r+1)}} \\  
        &\;\;\;\;\le  2.5c_Kp{L_K}T\frac{r^2\left(M\ln{M}\right)^{1/(r+1)}}{M} \mathop \le \limits^{(a)}  4e^2c_Kp{L_K}T\frac{\left(\ln{M}\right)^{2}}{M},
    \end{aligned}
    \label{eq:13}
\end{equation}
which is obtained by
\begin{equation}
    \frac{{{\rm{d}}{\varepsilon _K}}}{{{\rm{d}}v}} = p{L_K} - \frac{rc_K^{r+1}pL_kT^{r+1}\left(r!\right)^2\ln{M}}{v^{r+1}M^r}= 0 \Rightarrow v = \frac{r^{1/(r+1)}c_KT\left(r!\right)^{2/(r+1)}\left(\ln{M}\right)^{1/(r+1)}}{M^{r/(r+1)}},
    \label{eq:14}
\end{equation}
where in Step (\textit{a}) $r=\floor{\ln{M}}$ and $M^{1/(1+r)} \le M^{1/\floor{\ln{M}}}=e^{\ln{M}/\floor{\ln{M}}} < e^2$ for $M>2$ are utilized. The appropriate ranges of $v$ and $M$ are

\begin{equation}
    \left\{ 
    \begin{aligned}
        & \frac{T}{M} \le v \le c_KT\frac{r^{(2r+1)/(r+1)}\left(\ln{M}\right)^{1/(r+1)}}{M^{r/\left(1+r\right)}} \\
        & c_K\le \frac{M^{r/\left(1+r\right)}}{r^{(2r+1)/(r+1)}\left(\ln{M}\right)^{1/(r+1)}}
    \end{aligned}. \right.
    \label{eq:15}
\end{equation}
where $r=\floor{\ln{M}}$. Thus, for each integer \textit{M} satisfying Eq.~\eqref{eq:15},
the approximation error $\varepsilon_K$ in Eq.~\eqref{eq:9} can be expressed by the result in Eq.~\eqref{eq:13}. The condition of $M$ in Eq.~\eqref{eq:15} ensures that the underlying parameter $v$ is smaller than $T$}.

{\bf Lemma 2:} {\it Given a compact subset K $\subset \Czero{p}$ satisfying Assumptions 1 and 2, for every positive integer M {\color{black} larger than two} and arbitrary M frequencies $\omega_1$, $\omega_2$,..., $\omega_M$ $\in \mathbb{R}$, there exists a one-hidden-layer MLP ${\it{\Gamma}}(\cdot)$ with $\sigma_{\mathrm{ReLU}}(\cdot)$, the widths of its input, hidden, and output layers are $\wGammain = p(2M+1)$, $\wGamma = 2pM$, and $\wGammaout = pM$, respectively, such that for all $t \in [0,T]$ and all $\mathbf{u}(t) \in K$, the solution $\mathbf{x}(t) = [\mathbf{x}_1^\top(t),\mathbf{x}_2^\top(t),..., \mathbf{x}_M^\top(t)]^\top$ from the second-order ODE in Eq.~\eqref{eq:1} subject to $\mathbf{u}(t)$ with the initial conditions $\mathbf{x}(0) = \mathbf{x}'(0) = \mathbf{0}$ satisfies $\mathbf{x}(t) = [\mathcal{L}_t\mathbf{u}^\top(\omega_1),\mathcal{L}_t\mathbf{u}^\top(\omega_2),..., \mathcal{L}_t\mathbf{u}^\top(\omega_M)]^\top$, where $\mathcal{L}_t\mathbf{u}(\omega)$ is defined in Eq.~\eqref{eq:10} in Lemma 1.}

{\bf Lemma 3:} {\it Given a compact subset $K \subset \Czero{p}$ and a causal and uniformly continuous operator $\Phi: \Czero{p} \to \Czero{q}$ satisfying Assumptions 1, 2, 3, and 6, for an arbitrary positive $v < T$ and every positive integer M {\color{black} larger than two}, there exists a continuous mapping $\Psi = [\psi_1, \psi_2,..., \psi_q]: {\mathbb{R}}^{p(M+1)}\times{[0, T]} \to {\mathbb{R}}^q$ and a one-hidden-layer MLP ${\it{\Gamma}}(\cdot)$ specified in Lemma 2, such that for all $\mathbf{u}(t) \in K$, the solution $\mathbf{x}(t) = [\mathbf{x}_1^\top(t),\mathbf{x}_2^\top(t),..., \mathbf{x}_M^\top(t)]^\top$ from the second-order ODE in Eq.~\eqref{eq:1} subject to $\mathbf{u}(t)$ with the initial conditions $\mathbf{x}(0) = \mathbf{x}'(0) = \mathbf{0}$ satisfies
\begin{equation}
    \left| {\Phi \left[ {{\bf{u}}(\tau )} \right](t) - \Psi ({\bf{x}},{{\bf{u}}_0},t)} \right| \le {\phi _\Phi }({\varepsilon _K})
    \label{eq:16}
\end{equation}
for all $t \in [0, T]$, where $\mathbf{x} = [\mathbf{x}_1^\top,\mathbf{x}_2^\top,..., \mathbf{x}_M^\top]^\top$, $\mathbf{x}_n$ represents the value of $\mathbf{x}_n(t)$, $\mathbf{u}_0$ represents the value of $\mathbf{u}(0)$, $\phi_{\Phi}(\cdot)$ is a modulus of continuity of $\Phi$ in Assumption 6, and ${\varepsilon _K}$ is defined in Eq.~\eqref{eq:11} in Lemma 1.}

{\bf Remark 3:} Lemma 3 is an enhanced version of the novel fundamental lemma in \citet{lanthaler2023neural}
by providing a precise error bound and incorporating the influence of the initial value $\mathbf{u}(0)$, while adding one constraint that the target operator $\Phi$ is uniformly continuous. This constraint can ensure that the error term $\phi_{\Phi}(\varepsilon_K)$ in Eq.~\eqref{eq:16} is independent of $\mathbf{u}(t)$. Under the additional condition $\mathbf{\upphi}_{K}(t) = [L_Kt, L_Kt,..., L_Kt]^\top$, $\varepsilon_K$ in Eq.~\eqref{eq:16} can be calculated by Eq.~\eqref{eq:13} for $M$ satisfying Eq.~\eqref{eq:15}.

{\bf Remark 4:} Even if two additional conditions that $\mathbf{u}(0)$ = $\mathbf{0}$ for all $\mathbf{u}(t) \in K$ and $\Phi$ being time-autonomous are imposed, the term \textit{t} of $\Psi\left(\mathbf{x},{\mathbf{0}},t\right)$ in Lemma 3 cannot be ignored because $\sin\left[\omega_n\left(t-\tau \right) - \theta_n\right]$ in Eq.~\eqref{eq:a16} in Appendix A.3 contains the time $t$. The detailed explanation is provided in \ref{appC}.

{\bf Lemma 4:} {\it Under the conditions in Lemma 3 and Assumption 5, along with $\mathrm{\upphi}_K(t) = [L_Kt,L_Kt,...,L_Kt]^\top$, $v$ satisfying Eq.~\eqref{eq:14}, and an integer $M$ satisfying Eq.~\eqref{eq:15}, the values of all arguments of $\psi_j\left(\mathbf{x},\mathbf{u}_0,t\right)$ in Lemma 3 are within a compact domain $\Omega_{M}\left(TB_K,B_K,T\right) = \left[-TB_K,TB_K\right]^{pM} \times \left[-B_K,B_K\right]^p \times \left[0,T\right]$ for all $\mathbf{u}(t) \in K$ and $j = 1, 2,..., q$, the {\color{black} Lipschitz} constant $L_{{\psi _j}}\left(M\right)$ of $\psi_j\left(\mathbf{x},\mathbf{u}_0,t\right)$ with respect to its each argument is
{\color{black} \begin{equation}
    {L_{{\psi _j}}\left(M\right)} = \max \left\{ {{D_{{\Phi _j}}} + p{B_K}M^{3}T^{-1}c_{L}{{\phi '}_{{\Phi _j}}}(0),\max \left[ {\left( {1 + 2M} \right),T^{-1}} \right]{{\phi '}_{{\Phi _j}}}(0)} \right\},
    \label{eq:17}
\end{equation}}
and a linear modulus of continuity $\phi_{\psi_j,M}(x)$ of $\psi_j\left(\mathbf{x},\mathbf{u}_0,t\right)$ under {\color{black} the} $L^2$-norm is
\begin{equation}
    {\phi _{{\psi _j,M}}}(x) = \sqrt {p(M + 1) + 1} {L_{{\psi _j}}\left(M\right)}x,
    \label{eq:18}
\end{equation}
where $\phi_{\Phi_j}(\cdot)$ is the monotonic modulus of continuity of $\Phi_j$ in Assumption 6, ${\phi}'_{\Phi_j}(0) $ is the first-order derivative of  $\phi_{\Phi_j}(\cdot)$ at zero, $B_K$ is the bound of all $\mathbf{u}(t) \in K$ in Assumption 2, $D_{\Phi_j}$ is the derivative bound of the image set $\Phi_j\left\{ \Czero{p} \right\}$ in Assumption 5, and $c_L$ is an independent constant.}

{\bf Remark 5:} Lemma 4 establishes a compact domain $\Omega_{M}\left(TB_K,B_K,T\right)$ for the arguments of $\psi_j\left(\mathbf{x},\mathbf{u}_0,t\right)$ that is valid for all $\mathbf{u}(t) \in K$, every integer $M$ satisfying Eq.~\eqref{eq:15}, and $v$ satisfying Eq.~\eqref{eq:14}, along with the linear modulus of continuity $\phi_{\psi_j,M}(x)$ in Eq.~\eqref{eq:18} of $\psi_j\left(\mathbf{x},\mathbf{u}_0,t\right)$. Consequently, $\psi_j\left(\mathbf{x},\mathbf{u}_0,t\right)$ over $\Omega_{M}\left(TB_K,B_K,T\right)$ can be approximated using an MLP, and existing MLP approximation bounds can be applied to quantify the approximation error. By combining a utilized MLP approximation error with the term ${\phi _\Phi }({\varepsilon _K})$ from Eq.~\eqref{eq:16} in Lemma 3, an approximation bound of the neural oscillator in Eq.~\eqref{eq:1}, for approximating the causal and uniformly continuous operators between continuous temporal function spaces, is derived and presented in following Theorem 1.

{\bf Theorem 1:} {\it Given a compact subset $K \subset \Czero{p}$ and a causal and uniformly continuous operator $\Phi: \Czero{p} \to \Czero{q}$ satisfying Assumptions 1 to 6 with $\mathrm{\upphi}_K(t) = [L_Kt,L_Kt,...,L_Kt]^\top$, for every integer $M_{\it{\Gamma}}$ satisfying Eq.~\eqref{eq:15} along with $v$ satisfying Eq.~\eqref{eq:14}, and every positive integer $H_{\it{\Pi}}$, there exist two MLPs $\it{\Gamma}(\cdot)$ and $\it{\Pi}(\cdot)$ employing $\sigma_{\mathrm{ReLU}}(\cdot)$, such that for every $\mathbf{u}(t) \in K$, the corresponding solution $\mathbf{y}(t) = [y_1(t), y_2(t),...,y_q(t)]^{\top}$ from Eq.~\eqref{eq:1} subject to $\mathbf{u}(t)$ with the initial conditions $\mathbf{x}(0) = {\mathbf{x}}'(0) = \mathbf{0}$ satisfies
\begin{equation}
    \left| {\Phi \left[ {{\bf{u}}(\tau )} \right](t) - {\bf{y}}(t)} \right| \le {\varepsilon _{\bf{y}}}
    \label{eq:19}
\end{equation}
for all $t \in [0,T]$, where
{\color{black} \begin{equation}
    {\varepsilon _{\bf{y}}} = {\phi _\Phi }\left[ \frac{32c_KTpL_K\left(\ln{M_{\itGamma}}\right)^2}{M_{\itGamma}} \right] 
    + \frac{{{B_\Omega }\left[{p\left( {{M_{\it{\Gamma}} } + 1} \right) + 1}\right]^{2} }}{{\left(0.5H_{\it{\Pi}}\right)^{{1 \mathord{\left/
    {\vphantom {1 {\left[ {p\left( {{M_{\it{\Gamma}} } + 1} \right) + 1} \right]}}} \right.
    \kern-\nulldelimiterspace} {\left[ {p\left( {{M_{\it{\Gamma}} } + 1} \right) + 1} \right]}}}}}\left[\sum\limits_{j = 1}^q {{L_{{\psi _j}}}\left( {{M_{\it{\Gamma}} }} \right)}\right],
    \label{eq:20}
\end{equation}}
$B_{\Omega} = 2\max\left(TB_K,B_K,T\right)$, $B_K$ is the bound of all $\mathbf{u}(t) \in K$ in Assumption 2, ${L_{{\psi _j}}\left(\cdot\right)}$ is in Eq.~\eqref{eq:17}, ${\phi_{\Phi}}(\cdot)$ is the monotonic modulus of continuity of $\Phi$ in Assumption 6, and {\color{black} $c_K$ is the independent constant in Lemma 1}. $\it{\Gamma}(\cdot)$ has one hidden layer and the widths of its input, hidden, and output layers are $\wGammain$, $\wGamma$, and $\wGammaout$, respectively
\begin{equation}
    \left\{ 
    \begin{aligned}
        &w_{\it{\Gamma} ,{\rm{in}}} = p\left(2M_{\it{\Gamma}} + 1\right)\\
        &w_{\it{\Gamma}} = 2pM_{\it{\Gamma}}\\
        &{w_{\it{\Gamma} ,{\rm{out}}}} = pM_{\it{\Gamma}}
    \end{aligned}. \right.
    \label{eq:21}
\end{equation}
For ${\it{\Pi}}(\cdot)$, the widths of its input, hidden, and output layers are $\wPiin$, $\wPi$, and $\wPiout$, respectively
\begin{equation}
    \left\{
    \begin{aligned}
        &{w_{{\it{\Pi}} ,{\rm{in}}}} = p\left(M_{\it{\Gamma}} + 1\right) + 1\\
        &\wPi = q\left[p\left( M_{\it{\Gamma}} + 1 \right) + 4\right]\\
        &{w_{{\it{\Pi}},{\rm{out}}}} = q
    \end{aligned}, \right.
    \label{eq:22}
\end{equation}
and its depth is $\hPi = H_{\it{\Pi}} + 1$.} 

{\bf Remark 6:} Beyond the theoretical upper approximation bound for the neural oscillator, Lemma 1, Lemma 3, Lemma 4, and Theorem 1 establish a proof method for deriving the upper approximation bounds of the causal operator network architectures consisting of a linear transformation of input functions followed by an MLP. Particularly, Lemma 4 provides a method for characterizing the continuous mapping $\Psi \left( {\bf{x}},{\bf{u}}_0,t \right)$ in Lemma 3. If the parameters of linear time-continuous RNNs are complex \citep{ran2024provable}, the output functions of the linear complex RNNs can be expressed as the convolution of the trigonometric basis functions with the input functions. Consequently, the proof method established in this study can be directly employed to derive the approximation bound for the complex SS models consisting of a linear time-continuous complex RNN followed by an MLP.

{\bf Remark 7:} {\color{black} The approximation error of the neural oscillator in Theorem 1, which can be expressed as ${\varepsilon _{\bf{y}}} = {\phi _\Phi}\left\{O\left[\left(\ln{M_{\itGamma}}\right)^{2}/{M_{\it{\Gamma}}}\right]\right\} + O\left\{M_{\it{\Gamma}}^{5}H_{\it{\Pi}}^{-1/\left[p\left(M_{\itGamma}+1\right) +1\right]}\right\}$ in Eq~\eqref{eq:20}, consists of two parts. The second one arises from approximating the continuous mapping $\Psi\left(\mathbf{x},\mathbf{u}_0,t  \right)$ in Lemma 3 using the ReLU MLP ${\it{\Pi}}(\cdot)$ and it is derived from Theorem 1 by \citet{hanin2019universal}, where $p\left(M_{\itGamma}+1\right) +1$ in ${\varepsilon _{\bf{y}}}$ is the argument dimension of $\Psi ({\bf{x}},{{\bf{u}}_0},t)$. Theorem 4 in \citet{yarotsky2017error} provides a lower bound with a convergence rate of $-2/d$ for approximating Lipschitz continuous functions using ReLU MLPs, where $d$ is the argument dimension. A corresponding upper bound $O\left[\left(\ln{W}/W\right)^{2/d}\right]$, where $W$ represents the total parameter number of the employed ReLU MLP, was established in Theorem 4.1 of \citet{yarotsky2020phase}. Although the upper approximation bound by \citet{yarotsky2020phase} is closer to the lower bound with a convergence rate of $-2/d$ than $-1/d$ of Theorem 1 by \citet{hanin2019universal}, its approximation bound constant depends on the argument dimension of a target function, and this dependence is implicit. This implicit relationship hinders the establishment of the explicit relationship between the final approximation error ${\varepsilon _{\bf{y}}}$ in Eq.~\eqref{eq:19} and $M_{\it{\Gamma}}$ controlling the width of the hidden layer of the MLP ${\it{\Gamma}}(\cdot)$. The convergence rate of an upper bound for neural operators presented in Proposition 1 of \citet{kratsios2024mixture} is also in the form of $-1/d$. The first part of ${\varepsilon _{\bf{y}}}$ comes from approximating the input function $\mathbf{u}(t)$ using the sine transform. The convergence rate of a lower bound for approximating compactly supported continuous functions via the Fourier series is $-1$ \citep{devore1998nonlinear}. Compared to this lower bound, the additional logarithmic term $\left(\ln{M_{\itGamma}} \right)^{2}$ in Theorem 1 is due to the discontinuities of $\tilde{\mathbf{f}}_t(\tau)$ in Eq.~\eqref{eq:a1} of \ref{appA.1}, as discussed in Remark 2. The sine transform encoding method is linear, whereas the neural ODE in the neural oscillator can represent a wide range of nonlinear dynamical systems. Whether there exists an encoding method based on nonlinear neural ODEs that can achieve a faster bound than $O\left[\left(\ln{M_{\itGamma}}\right)^{2}/{M_{\it{\Gamma}}}\right]$ in Theorem 1 remains an interesting direction for future investigation.}

{\color{black} {\bf Remark 8:} Theorem 1 in \citet{zhang2024deep} demonstrates that for a large set of activation functions, including the commonly used tanh function $\sigma_{\text{tanh}}(x) = \left(e^{x}-e^{-x}\right)/\left(e^{x}+e^{-x}\right)$ and sigmoid function $\sigma_{\text{sigmoid}}(x) = 1/\left(1+e^{-x}\right)$, any ReLU MLP with the maximum width $W$ and depth $H$ can be approximated to arbitrary accuracy by an MLP using any activation function from the activation function set, with the maximum width $3W$ and depth $2H$. Furthermore, Lemma 3.7 in \citet{lanthaler2023neural} shows that the sine transform $\mathcal{L}_t\mathbf{u}(\omega)$ in Eq.~{\eqref{eq:10}} can be approximated to arbitrary accuracy by the solution of a second-order neural ODE with an activation function $\sigma(x)$ satisfying $\sigma(0) = 0$ and $\sigma'(0) = 1$. These two conditions are satisfied by $\sigma_{\text{tanh}}(x)$ and the scaled sigmoid function $2\sigma_{\text{sigmoid}}(2x)-1$. Therefore, approximation bounds with slightly larger bound constants, that are similar to the result in Theorem 1, can be achieved when $\sigma_{\text{tanh}}(x)$ or $\sigma_{\text{sigmoid}}(x)$ is utilized in the neural oscillator.}

{\color{black} {\bf Remark 9:} Theorem 1 first decomposes the input $\mathbf{u}(t)$ into $M_{\itGamma}$ components, and subsequently maps these $M_{\itGamma}$ components to the output $\mathbf{y}(t)$. Since the function set $K$ containing $\mathbf{u}(t)$ can be infinite-dimensional, $M_{\itGamma}$ must tend to infinity for the first part of ${\varepsilon _{\bf{y}}}$ in Eq.~\eqref{eq:20} to vanish. However, this will cause the convergence rate $-1/\left[p\left(M_{\it{\Gamma}} + 1\right) + 1 \right]$  in the second part of ${\varepsilon _{\bf{y}}}$ to diminish towards zero.} Thus, the neural oscillator in Eq.~\eqref{eq:1} for approximating the causal continuous operators between continuous temporal function spaces also faces the curse of parametric complexity, which is a common problem in neural operators \citep{lanthaler2023operator,lanthaler2025parametric}.

One method to overcome the curse of parametric complexity is applying other super-expressive activation functions \citep{maiorov1999lower,shen2021neural,yarotsky2021elementary,zhang2022deep}. For instance, Theorem 4 in \citet{maiorov1999lower} demonstrates the existence of an activation function that enables a two-hidden-layer neural network architecture to approximate continuous functions with arbitrary accuracy, even when the widths of the hidden layers are held constant. Theorem 1.1 and Corollary 1.2 in \citet{shen2021neural} provide a three-hidden-layer MLP architecture, called Floor-Exponential-Step (FLES) network, utilizing the floor function $\floor{x}$, the exponential function $2^x$, and the step function as the activation function in each neuron. FLES is capable of approximating arbitrary continuous functions with the number of parameters scaling logarithmically with the reciprocal of the approximation error. Theorem 1 in \citet{zhang2022deep} introduces a neural network architecture that employs an activation function combining a softsign activation function with a periodic triangular-wave function. This neural network architecture with a width $36d(2d + 1)$ and a depth 11 can approximate any continuous function on a $d$-dimensional hypercube within an arbitrarily small error. Super-expressive activation functions have been utilized to overcome the curse of parametric complexity in neural operators \citep{schwab2023deep}. Since the explicit expressions for the Lipschitz constant $L_{\it{\psi}_j}(M)$ in Eq.~\eqref{eq:17} and the linear modulus of continuity $\phi_{\it{\psi}_j,M}(x)$ in Eq.~\eqref{eq:18} of the continuous mapping $\psi_j\left(\mathbf{x},\mathbf{u}_0,t\right)$, along with its valid compact argument domain $\Omega_{M}\left(TB_K,B_K,T\right)$, have been derived, the super-expressive activation functions mentioned above can be directly applied to the MLP ${\it{\Pi}}(\cdot)$ to overcome the curse of parametric complexity in the neural oscillator. On the other hand, \citet{lanthaler2024operator} has demonstrated that decreasing the required number of parameters in neural operators using super-expressive neural networks may come at the expense of requiring larger model parameter values.

\subsection{Second-Order Dynamical Systems}
\label{subsec3.2}
In this subsection, an upper approximation bound of the neural oscillator in Eq.~\eqref{eq:1} for approximating uniformly asymptotically incrementally stable second-order dynamical systems is derived. First, the asymptotically incrementally stable second-order dynamical system is briefly introduced. Subsequently, Lemma 5 and Lemma 6 establish the relationships between the solution error of two ODEs subject to the same input functions and the error of the derivative functions of the two ODEs. Finally, by utilizing the result in Lemma 6 and the error bound for the MLP approximation of Barron class functions {\color{black} \citep{barron1993universal,siegel2023characterization}}, the upper approximation bound of the neural oscillator is demonstrated in Theorem 2. This subsection builds on the methodology of \citet{hanson2020universal} and offers technical contributions through detailed mathematical derivations, as elaborated in the accompanying remarks. The proofs for all lemmas and Theorem 2 in this subsection are presented in \ref{appB}.

Two second-order ODEs ${\mathbf{x}}_i''(t) = g_i\left[{\mathbf{x}}_i(t),{\mathbf{x}}_i'(t),\mathbf{u}(t)\right]$, where ${\mathbf{x}}_i(t) = [x_{i,1}(t)$ $,x_{i,2}(t),\dots,x_{i,r}(t)]^{\top}$, \textit{i} = 1 and 2, can be rewritten in the state function form
\begin{equation}
    \left\{ 
    \begin{aligned}
    &{{{\bf{z}}}_{i,1}'}(t) = {{\bf{z}}_{i,2}}(t)\\
    &{{{\bf{z}}}_{i,2}'}(t) = {g_i}\left[ {{{\bf{z}}_{i,1}}(t),{{\bf{z}}_{i,2}}(t),{\bf{u}}(t)} \right]
    \end{aligned}\right.,
    \label{eq:23}
\end{equation}
where ${\mathbf{z}}_i(t) = \left[{\mathbf{z}}_{i,1}^{\top}(t),{\mathbf{z}}_{i,2}^{\top}(t)\right]^{\top} = \left[{\mathbf{x}}_{i}^{\top}(t),{\mathbf{x}}_{i}'^{\top}(t)\right]^{\top}$.

{\bf Definition 1 \citep{hanson2020universal}:} The dynamical system in the form of Eq.~\eqref{eq:23} is uniformly asymptotically incrementally stable for the inputs in a subset $K \subset \Czero{p}$ on a domain $\Omega_{\mathbf{z}} \subset \mathbb{R}^{2r}$ if there exists a set of non-negative stability functions $\beta_k(h,t)$ over $h \geq 0$ and $t \geq 0$, $k = 1, 2,..., 2r$, where $\beta_k(0,t) = 0$,$\lim\limits_{t\to +\infty}\beta_k(h,t) = 0$, $\beta_k(h,t)$ is continuous and strictly increasing with respect to \textit{h} and is continuous and strictly decreasing with respect to \textit{t}, such that
\begin{equation}
    \left| {{z_k}(t,{{\bf{z}}_\tau }) - {z_k}(t,{{{\hat{\bf z}}}_\tau })} \right| \le {\beta _k}\left( {{{\left| {{{\bf{z}}_\tau } - {{{\hat{\bf z}}}_\tau }} \right|}_2},t - \tau } \right)
    \label{eq:24}
\end{equation}
holds for all $\mathbf{u}(t) \in K$, all ${\mathbf{z}}_\tau$ and ${\hat{\bf z}}_\tau$ $\in \Omega_{\mathbf{z}}$, all $t \in [0,T]$, and all $0 \leq \tau \leq t$, where $z_k\left(t, {\mathbf{z}}_\tau\right)$ is the $k^{\text{th}}$ element of $\mathbf{z}\left(t, {\mathbf{z}}_\tau\right)$ and $\mathbf{z}\left(t, {\mathbf{z}}_\tau\right)$ represents the solution $\mathbf{z}(t)$ at \textit{t} driven by $\mathbf{u}(t)$ with an initial condition ${\mathbf{z}}_{\tau}$ at an initial time instant $\tau$, $0 \leq \tau \leq t$. The corresponding stability bound $B_\beta$ is calculated as
\begin{equation}
    {B_\beta } = \int_0^{ + \infty } {{{\left. {\frac{\partial }{{\partial h}}\sum\limits_{k = 1}^{2r} {{\beta _k}(h,\tau )} } \right|}_{h = {0^ + }}}{\rm{d}}\tau}.
    \label{eq:25}
\end{equation}

{\bf Lemma 5:} {\it Let K be a compact subset of $\Czero{p}$, $g_1(\cdot)$ and $g_2(\cdot)$ in Eq.~\eqref{eq:23} be Lipschitz continuous over $\mathbb{R}^{2r+p}$, and $\Omega_{\mathbf{u}} \subset \mathbb{R}^p$ and $\Omega_{\mathbf{z}} \subset \mathbb{R}^{2r}$ be two compact domains. Assume that the values of all $\mathbf{u}(t) \in K$ are in $\Omega_{\mathbf{u}}$ and the values of all possible solutions to Eq.~\eqref{eq:23}, governed by $g_1(\cdot)$ or $g_2(\cdot)$, driven by all $\mathbf{u}(t) \in K$, with a specified initial condition ${\mathbf{z}}_0 \in \Omega_{\mathbf{z}}$, remain within $ \Omega_{\mathbf{z}}$. Then, for the dynamical systems governed by $g_1(\cdot)$ or $g_2(\cdot)$ in Eq.~\eqref{eq:23}, their corresponding solutions ${\mathbf{z}}_1(t) = [{\mathbf{z}}_{1,1}^{\top}(t),{\mathbf{z}}_{1,2}^{\top}(t)]^{\top}$ and ${\mathbf{z}}_2(t) = [{\mathbf{z}}_{2,1}^{\top}(t),{\mathbf{z}}_{2,2}^{\top}(t)]^{\top}$, driven by the same arbitrary $\mathbf{u}(t) \in K$ with the same initial conditions ${\mathbf{z}_1(0) = {\mathbf{z}}_2(0) = {\mathbf{z}}_0}$, satisfy
\begin{equation}
    {{\bf{z}}_2}(t,{{\bf{z}}_0}) - {{\bf{z}}_1}(t,{{\bf{z}}_0}) = \int_0^t {\frac{{\partial {{\bf{z}}_1}\left[ {t,{{\bf{z}}_2}(\tau ,{{\bf{z}}_0})} \right]}}{{\partial {{\bf{z}}_{2,2}}(\tau ,{{\bf{z}}_0})}}\left\{ {{g_2}\left[ {{{\bf{z}}_{2,1}}(\tau ,{{\bf{z}}_0}),{{\bf{z}}_{2,2}}(\tau ,{{\bf{z}}_0}),{\bf{u}}(\tau )} \right] - {g_1}\left[ {{{\bf{z}}_{2,1}}(\tau ,{{\bf{z}}_0}),{{\bf{z}}_{2,2}}(\tau ,{{\bf{z}}_0}),{\bf{u}}(\tau )} \right]} \right\}{\rm{d}}\tau }
    \label{eq:26}
\end{equation}
for all $t \in [0, T]$, where ${\mathbf{z}}_i\left[t, {\mathbf{z}}_j\left({\hat{\tau}}, {\mathbf{z}}_{\tau} \right) \right]$ represents the solution ${\mathbf{z}}_i(t)$ at t driven by $\mathbf{u}(t)$ with the initial condition ${\mathbf{z}}_j\left({\hat{\tau}}, {\mathbf{z}}_{\tau} \right)$ at ${\hat{\tau}}$ with $0 \leq \tau \leq {\hat{\tau}} \leq t$ and $i,j = $ 1 and 2.}

{\color{black} {\bf Remark 10:}} Being analogous to the fundamental Lemma 3.1.2 in \citet{van2007filtering}, which has been employed by \citet{hanson2020universal} to derive the approximation bound of RNNs for modeling uniformly asymptotically incrementally stable first-order dynamical systems, Lemma 5 in this study provides a more rigorous proof. In the proof of the step involving Eqs.~\eqref{eq:b2} and \eqref{eq:b3} in \ref{appB.1}, the proof of the original lemma in \citet{van2007filtering} utilizes ${\mathbf{z}}_1(\tau) = {\mathbf{z}}_1\left(\tau, {\mathbf{z}}_0\right)$ but not ${\mathbf{z}}_1\left({\hat{\tau}},{\mathbf{z}}_{\tau}\right) = {\mathbf{z}}_{\tau}$ with $\hat{\tau} = \tau$ and replaces an arbitrary ${\mathbf{z}}_1\left(\tau, {\mathbf{z}}_0\right)$ with its corresponding ${\mathbf{z}}_2\left(\tau, {\mathbf{z}}_0\right)$ under the same $\mathbf{u}(t)$ to prove Eq.~\eqref{eq:b3}. However, this proof method is not rigorous. For an arbitrary time instant $\tau$ ($0 \leq \tau \leq T$), the range of all possible values of ${\mathbf{z}}_1\left(\tau, {\mathbf{z}}_0\right)$ depends on the differential equation governed by $g_1(\cdot)$ and all $\mathbf{u}(t) \in K$. It is evident that an arbitrary value within the range of all possible values of ${\mathbf{z}}_1\left(\tau, {\mathbf{z}}_0\right)$ satisfies Eq.~\eqref{eq:b2}. However, it remains uncertain whether all possible values of ${\mathbf{z}}_2\left(\tau, {\mathbf{z}}_0\right)$ at the same time instant $\tau$ driven by $\mathbf{u}(t)\in K$ lie within the {\color{black} same} range of the values of ${\mathbf{z}}_1\left(\tau, {\mathbf{z}}_0\right)$. If not, it is unclear whether ${\mathbf{z}}_2\left(\tau, {\mathbf{z}}_0\right)$ satisfies Eq.~\eqref{eq:b2} for all $0 \leq \tau \leq T$ and all $\mathbf{u}(t) \in K$. In this study, employing an arbitrary $\tau$ and an arbitrary ${\mathbf{z}}_1\left({\hat{\tau}},{\mathbf{z}}_{\tau}\right) = {\mathbf{z}}_{\tau}$ with $\hat{\tau} = \tau$ in Eq.~\eqref{eq:b2} resolves this uncertainty. ${\mathbf{z}}_1\left({\hat{\tau}},{\mathbf{z}}_{\tau}\right)$ is the solution at ${\hat{\tau}}$ caused by the initial condition ${\mathbf{z}}_{\tau}$ at $\tau$. Being an initial condition, the value of ${\mathbf{z}}_{\tau}$ can be an arbitrary vector within $\Omega_{\mathbf{z}}$. ${\mathbf{z}}_1\left({\hat{\tau}},{\mathbf{z}}_{\tau}\right) = {\mathbf{z}}_{\tau}$ with $\hat{\tau} = \tau$ indicates that the value of ${\mathbf{z}}_1\left({\hat{\tau}},{\mathbf{z}}_{\tau}\right)$ can also be an arbitrary vector within $\Omega_{\mathbf{z}}$. This implies that all vectors within $\Omega_{\mathbf{z}}$ satisfy Eq.~\eqref{eq:b2}. From the assumption of Lemma 5, the possible values of ${\mathbf{z}}_2\left(\tau, {\mathbf{z}}_0\right)$ fall within $\Omega_{\mathbf{z}}$ for all $\tau \in [0,T]$ and all $\mathbf{u}(t) \in K$. Consequently, all possible values of ${\mathbf{z}}_2\left(\tau, {\mathbf{z}}_0\right)$ satisfy Eq.~\eqref{eq:b2}.

{\bf Lemma 6:} {\it Under the conditions in Lemma 5, the difference between solutions ${\mathbf{z}}_1(t)$ and ${\mathbf{z}}_2(t)$ can be bounded by
\begin{equation}
    \left| {{{\bf{z}}_2}(t,{{\bf{z}}_0}) - {{\bf{z}}_1}(t,{{\bf{z}}_0})} \right| \leq {e^{T{L_{{g_1}}}}}T\mathop{\max }\limits_{\tau  \in [0,T]} \left| {{g_2}\left[ {{{\bf{z}}_{2,1}}(\tau ,{{\bf{z}}_0}),{{\bf{z}}_{2,2}}(\tau ,{{\bf{z}}_0}),{\bf{u}}(\tau )} \right] - {g_1}\left[ {{{\bf{z}}_{2,1}}(\tau ,{{\bf{z}}_0}),{{\bf{z}}_{2,2}}(\tau ,{{\bf{z}}_0}),{\bf{u}}(\tau )} \right]} \right|
    \label{eq:27}
\end{equation}
for all $t \in [0,T]$ and all $\mathbf{u}(t) \in K$, where $L_{g_1} = \max\left[L_{g_1,1},\left(L_{g_1,2} +1\right)\right]$, $L_{g_1,1}$ and $L_{g_1,2}$ are respectively the two Lipschitz constants of $g_1(\cdot)$ with respect to its first and second arguments. Furthermore, if the dynamical system governed by $g_1(\cdot)$ in Eq.~\eqref{eq:23} is uniformly asymptotically incrementally stable for all $\mathbf{u}(t) \in K$ on the domain $\Omega_{\mathbf{z}}$ in Lemma 5 with a stability bound $B_{\beta}$ in {\color{black} Eq.~\eqref{eq:25}}, then, $\left|{\mathbf{z}}_2\left(\tau, {\mathbf{z}}_0\right) - {\mathbf{z}}_1\left(\tau, {\mathbf{z}}_0\right)  \right|$ can be bounded by
\begin{equation}
    \left| {{{\bf{z}}_2}(t,{{\bf{z}}_0}) - {{\bf{z}}_1}(t,{{\bf{z}}_0})} \right|\leq \mathop {\max }\limits_{\tau  \in [0,T]} {\left| {{g_2}\left[ {{{\bf{z}}_{2,1}}(\tau ,{{\bf{z}}_0}),{{\bf{z}}_{2,2}}(\tau ,{{\bf{z}}_0}),{\bf{u}}(\tau )} \right] - {g_1}\left[ {{{\bf{z}}_{2,1}}(\tau ,{{\bf{z}}_0}),{{\bf{z}}_{2,2}}(\tau ,{{\bf{z}}_0}),{\bf{u}}(\tau )} \right]} \right|_2}{B_\beta }
    \label{eq:28}
\end{equation}
for all $\mathbf{u}(t) \in K$ and all $t \in [0,T]$.}

{\color{black} {\bf Remark 11:}} As shown in \ref{appD}, the result in Eq.~\eqref{eq:27} can also be proven using the method based on the Grönwall's inequality \citep{ames1997inequalities}, which was frequently employed in deriving approximation bounds of RNNs for modeling first-order dynamical systems \citep{funahashi1993approximation,chow2000modeling}. The proof of Lemma 6 shows that the method based on Lemma 5, which enhances the proof of Lemma 3.1.2 in \citet{van2007filtering}, can simultaneously obtain both categories of the bounds in {\color{black} Eqs.~\eqref{eq:27} and \eqref{eq:28}} for the difference between the solutions of two different second-order ODEs subject to the same input functions. This advancement enhances the theoretical understanding of the second-order ODEs. Lemma 6 will be utilized in the subsequent Theorem 2 to establish the approximation bound of the neural oscillator for modeling uniformly asymptotically incrementally stable second-order dynamical systems.

{\bf Theorem 2:} {\it Let K be a compact subset of $\Czero{p}$ and a causal continuous operator $\Phi: \Czero{p} \to \Czero{q}$ be a second-order dynamical system represented by
\begin{equation}
    \left\{
    \begin{aligned}
        &{\hat{\bf x}}''(t) = g\left[ {{\hat{\bf x}}(t),{\hat{\bf x}}'(t),{\bf{u}}(t)} \right]\\
        &{\hat{\bf y}}(t) = h\left[ {{\hat{\bf x}}(t)} \right]
    \end{aligned} 
    \right.
    \label{eq:29}
\end{equation}
with the initial conditions ${\hat{\mathbf{x}}}(0) = {{\hat{\mathbf{x}}}}'(0) = \mathbf{0}$, where ${\hat{\mathbf{x}}}(t) = \left[{\hat{x}}_1(t),{\hat{x}}_2(t),\dots,{\hat{x}}_r(t)\right]^{\top}$ and ${\hat{\mathbf{y}}}(t) = \left[{\hat{y}}_1(t),{\hat{y}}_2(t),\dots,{\hat{y}}_q(t)\right]^{\top}$. {\color{black} $g(\cdot) = \left[g_1(\cdot), g_2(\cdot),\dots,g_r(\cdot)\right]^{\top}$ satisfies that for every compact domain $\Omega \subset $ ${\mathbb{R}}^{2r+p}$, there exists a Barron function extension $\tilde{g}_{\Omega}(\cdot) = g(\cdot)$ on $\Omega$ and the spectral Barron norms of all extensions are uniformly bounded. $h(\cdot) = \left[h_1(\cdot), h_2(\cdot),\dots,h_q(\cdot)\right]^{\top}$ satisfies that for every compact domain $\Omega \subset $ ${\mathbb{R}}^{r}$, there exists a Barron function extension $\tilde{h}_{\Omega}(\cdot) = h(\cdot)$ on $\Omega$ and the spectral Barron norms of all extensions are uniformly bounded.} Assume that the dynamical system in Eq.~\eqref{eq:29} is uniformly asymptotically incrementally stable for all $\mathbf{u}(t) \in K$ on a domain $\left[-B_{\mathbf{x}}, B_{\mathbf{x}}\right]^{r} \times \left[-B_{{\mathbf{x}}'}, B_{{\mathbf{x}}'}\right]^{r} \times \left[-B_K, B_K\right]^{p} \subset \mathbb{R}^{2r+p}$, where $B_{{\mathbf{x}}} = \alpha B_{\beta{g}} + B_{\hat{\mathbf{x}}}$, $B_{{\mathbf{x}
}'} = \alpha B_{\beta{g}} + B_{\hat{\mathbf{x}}'}$, $\alpha$ is a positive coefficient, $B_{\hat{\mathbf{x}}}$ and $B_{\hat{\mathbf{x}}'}$ are the bounds of ${\hat{\mathbf{x}}}(t)$ and ${\hat{\mathbf{x}}}'(t)$, respectively, $B_K$ is the bound of all $\mathbf{u}(t) \in K$ in Assumption 2, and $B_{\beta{g}}$ is the stability bound of the dynamical system in Eq.~\eqref{eq:25}, then, for arbitrary positive errors $\varepsilon_1 \leq \alpha/r$ and $\varepsilon_2$, there exist two one-hidden-layer MLPs ${\it{\Gamma}}(\cdot)$ and ${\it{\Pi}}(\cdot)$ employing $\sigma_{\mathrm{ReLU}}(\cdot)$, such that for all $\mathbf{u}(t) \in K$, the corresponding solutions $\mathbf{y}(t)$ to Eq.~\eqref{eq:1} and ${\hat{\mathbf{y}}}(t)$ to Eq.~\eqref{eq:29} satisfy
\begin{equation}
    \left|{{\bf{y}}(t) - {\hat{\bf y}}(t)} \right| \le {L_h}{B_{{\beta _g}}}\sqrt r {\varepsilon _1} + q{\varepsilon _2}
    \label{eq:30}
\end{equation}
for all $t \in [0,T]$, where $L_h$ is the Lipschitz constant of $h(\cdot)$ in Eq.~\eqref{eq:29}. For ${\it{\Gamma}}(\cdot)$, the widths of its input, hidden, and output layers are $\wGammain$, $\wGamma$, and $\wGammaout$, respectively,
\begin{equation}
    \left\{ 
    \begin{aligned}
        &{w_{{\it{\Gamma}} ,{\rm{in}}}} = 2r + p\\
        &{w_{\it{\Gamma}}} = 8r\left\lceil {{C_{\it{\Gamma}} }\varepsilon _1^{ - 2}} \right\rceil\\
        &{w_{{\it{\Gamma}},{\rm{out}}}} = r
    \end{aligned} 
    \right.,
    \label{eq:31}
\end{equation}
where $C_{\it{\Gamma}}$ is a constant only depending on $g(\cdot)$, $\sigma_{\mathrm{ReLU}}(\cdot)$, $r$, $p$, $B_{\mathbf{x}}$, $B_{{\mathbf{x}}'}$, and $B_K$. For ${\it{\Pi}}(\cdot)$, the widths of its input, hidden, and output layers are $\wPiin$, $\wPi$, and $\wPiout$, respectively,
\begin{equation}
    \left\{ 
    \begin{aligned}
        &{w_{{\it{\Pi}} ,{\rm{in}}}} = r + p + 1\\
        &{w_{\it{\Pi}}} = 8q\left\lceil {{C_{\it{\Pi}}}\varepsilon _2^{ - 2}} \right\rceil\\
        &{w_{{\it{\Pi}} ,{\rm{out}}}} = q
    \end{aligned} 
    \right.,
    \label{eq:32}
\end{equation}
where $C_{\it{\Pi}}$ is a constant only depending on $h(\cdot)$, $\sigma_{\mathrm{ReLU}}(\cdot)$, $r$, and $B_{\mathbf{x}}$.}

{\color{black} {\bf Remark 12:}} Theorem 2 demonstrates that when the target causal continuous operator can be explicitly represented as a uniformly asymptotically incrementally stable second-order dynamical system, the widths of the hidden layers in the two MLPs utilized in the neural oscillator scale polynomially with the reciprocals of the approximation errors {\color{black} under the $\it{L^{\infty}}$-norm. This implies that the curse of parametric complexity of the neural oscillator in this situation is overcome, in contrast to the result established in Theorem 1 that is discussed in Remark 9. The distinction stems from the fact the target operator in Theorem 2 is an explicit second-order ODE and the neural oscillator only needs to use its MLPs $\itGamma(\cdot)$ and $\itPi(\cdot)$ to approximate the functions $g(\cdot)$ and $h(\cdot)$, respectively. Since the argument dimensions of $g(\cdot)$ and $h(\cdot)$ are fixed hyperparameters for a given target ODE, the neural oscillator can achieve polynomial convergence rates independent of the dimension of the function set $K$ containing $\mathbf{u}(t)$.}

{\color{black} {\bf Remark 13:} The dimension-independent convergence rates of $-0.5$ in Eqs.~\eqref{eq:31} and \eqref{eq:32} are achieved because $g(\cdot)$ and $h(\cdot)$ in Theorem 2 admit Barron function extensions over any compact domain of their respective arguments, with the corresponding spectral Barron norms defined in Theorem 4 of \citet{siegel2023characterization} being uniformly bounded. Theorem 2 in \citet{barron1992neural} indicates that similar error bounds, with the convergence rates of $-0.5$ as in Eqs.~\eqref{eq:30}–\eqref{eq:32}, can also be achieved when the tanh or sigmoid activation function is utilized. $g(\cdot)$ and $h(\cdot)$ belong to the variation space generated by a spectral dictionary that was established by \citet{siegel2023characterization}. The Barron spectrum space ${\mathcal{B}_{1,2}}(\mathbb{R}^r)$ in Definition 2.1 of \citet{meng2022new} is a subset of the possible $h(\cdot)$ in the variation space. Theorem 4.3 in \citet{meng2022new} demonstrates that the Sobolev space ${\mathcal{W}^{\alpha,2}}(\mathbb{R}^r)$ satisfying $\alpha > 1 + r/2 >0$ is embedded in ${\mathcal{B}_{1,2}}(\mathbb{R}^r)$. Therefore, ${\mathcal{W}^{\alpha,2}}(\mathbb{R}^r)$ with $\alpha > 1 + r/2 >0$ is also a subset of the possible $h(\cdot)$. The approximation error in Theorem 2 has two parts $O\left( w_{\itGamma}^{-1/2}\right) + O\left( w_{\itPi}^{-1/2}\right)$. For the second one, it arises from approximating $h(\cdot)$ using a one-hidden-layer ReLU MLP $\itPi(\cdot)$. Since ${\mathcal{W}^{\alpha,2}}(\mathbb{R}^r)$ is a subset of the possible $h(\cdot)$, and the one-hidden-layer ReLU MLP $\itPi(\cdot)$ for approximating ${\mathcal{W}^{\alpha,2}}(\mathbb{R}^r)$ has a lower bound in the form of ${\it{\Omega}}\left[\wPi^{-\alpha/(r-1)}\right]$ \citep{maiorov1999best}. Thus, $-\left(2 + 0.5r\right)/\left(r-1\right)$ is a lower-bound convergence rate which approaches the dimension-independent rate of $-0.5$ in Theorem 2 as the argument dimension $r$ of $h(\cdot)$ increases. For the first part of the approximation error in Theorem 2, since two ODEs with two different derivative term functions, e.g., $g_1\left[ {{\hat{\bf x}}(t),{\hat{\bf x}}'(t),{\bf{u}}(t)} \right]$ and $g_2\left[ {{\hat{\bf x}}(t),{\hat{\bf x}}'(t),{\bf{u}}(t)} \right]$, can have the same solution $\hat{\mathbf{x}}(t)$ under the same input $\mathbf{u}(t)$, as long as $g_1(\cdot)$ and $g_2(\cdot)$ agree on the trajectories of ${\hat{\bf x}}(t),{\hat{\bf x}}'(t)$, and ${\bf{u}}(t)$, a lower bound of the MLP $\itGamma(\cdot)$ to approximate the function $g(\cdot)$ in Eq.~\eqref{eq:29} is not sufficient  to establish a lower bound of the neural ODE in the neural oscillator to approximate the target ODE. Currently, research on the lower bounds for the neural ODEs viewed as mappings from input functions to solution functions remains limited. \citet{bittner2023vapnik} established the Vapnik-Chervonenkis (VC) dimension bounds for discrete-time RNNs and liquid time constant networks. Since a neural oscillator with a fixed-step ODE solver is structurally similar to these two discrete-time neural networks, extending their framework to derive the VC dimension of the neural oscillator and subsequently establishing its lower bound represents a natural direction for future research.}

{\color{black}{\bf Remark 14:}} Approximating an ODE system by a neural ODE involves approximating the derivative function of the target ODE using an MLP. The approximation bound for an MLP approximating the derivative function is established over a compact domain of the derivative function's arguments and the MLP's arguments. The derivative function's arguments are {\color{black}the} input functions and solutions of the target ODE. The MLP's arguments are {\color{black}the} input functions and solutions of the neural ODE. Therefore, in establishing an approximation bound for a neural ODE approximating an ODE system, where the derivative function of the target ODE is approximated by the MLP of the neural ODE, one critical issue is ensuring that the values of all possible solutions from the neural ODE, driven by a set of considered input functions, can lie within the compact domain, where the error between the MLP of the neural ODE and the target derivative function can be quantified. To meet this requirement, Lemma 7 in \citet{hanson2020universal} proposes to approximate the derivative function of a target ODE using a compactly supported function, such that the values of the solutions from the resulting neural ODE, whose derivative function is modeled by this compactly supported function, can be bounded in a compact domain. In this study, the relationships between the solution error of two ODEs subject to the same input functions and the error of the derivative functions of the two ODEs are established in Eqs.~\eqref{eq:27} and \eqref{eq:28} of Lemma 6. Based on the result of Eq.~\eqref{eq:28}, the error between the MLP ${\it{\Gamma}}(\cdot)$ and the target derivative function $g(\cdot)$ in Eq.~\eqref{eq:29} is quantified over the compact domain $\Omega\left(B_{\mathbf{x}}^{r}, B_{{\mathbf{x}}'}^{r}, B_K^p\right) = \left[-B_{\mathbf{x}}, B_{\mathbf{x}}\right]^{r} \times \left[-B_{{\mathbf{x}}'}, B_{{\mathbf{x}}'}\right]^{r} \times \left[-B_K, B_K\right]^{p}$. And all possible values of the solutions $\mathbf{x}(t)$ and ${\mathbf{x}}'(t)$ to the second-order ODE in Eq.~\eqref{eq:1} with ${\it{\Gamma}}(\cdot)$ driven by all $\mathbf{u} \in K$ are within the domain $\Omega\left(B_{\mathbf{x}}^{r}, B_{{\mathbf{x}}'}^{r}, B_K^p\right)$, as highlighted in {\color{black} Eqs.~\eqref{eq:b20}, \eqref{eq:b21}, and \eqref{eq:b27}} in the proof of Theorem 2 in \ref{appB.3}.

{\color{black} {\bf Remark 15: } The approximation error bounds established in Theorems 1 and 2 are existence results. These results show that there exist weight configurations for the two MLPs $\itGamma(\cdot)$ and $\itPi(\cdot)$ such that the neural oscillator in Eq.~\eqref{eq:1} can approximate the target causal operators or dynamical systems within the established error bounds. Furthermore, upper bounds on the generalization errors of the neural oscillator, accounting for the error induced by the finite number of training samples, have been established in \citet{huang2026upper}. However, these theoretical results do not address the training problem. Specifically, they do not specify which training algorithms can efficiently identify models that can achieve the theoretically predicted approximation and generalization error bounds. This gap between theory and practice in neural networks is a frontier research challenge. Notably, \citet{kratsios2025beyond} proposed a specific neural network architecture that combines the attention mechanism with a random neural network, for which the theoretically predicted generalization performance can be achieved by trained models with high probability using a finite number of samples. Developing specialized neural oscillator architectures, together with efficient training strategies, that enable trained models to realize the theoretically predicted error bounds, is a promising direction for future research.}

{\color{black} {\bf Remark 16:} Since the error bounds established in Theorems 1 and 2 contain several constants that are difficult to evaluate in practice, they are not directly suitable for determining appropriate neural network sizes for specific problems under prescribed error tolerances. Nevertheless, these theoretical results still provide useful guidance. In practice, initial network sizes can be selected based on prior experience. Then, by combining the convergence rates derived in Theorems 1 and 2 with the approximation errors obtained from the chosen network configurations, the network sizes can be further estimated and adjusted to achieve the desired accuracy. Furthermore, the proof process of the two theorems provides the mathematical explanations for the possible behaviors of the neural oscillator in approximating various operator mappings. This theoretical understanding of the neural oscillator can promote its more effective application compared with other models. For example, the time-continuous RNN, consisting of a neural ODE with a linear readout function, is a universal approximator for dynamical ODE mappings \citep{hanson2021learning}. However, for causal mappings between temporal function spaces that cannot be expressed as dynamical ODEs, the approximation capability of time-continuous RNNs remains unclear. In contrast, Theorem 1 in this study demonstrates that the neural oscillator can universally approximate causal and continuous operators, including those that cannot be represented in the form of dynamical ODEs. While attention-based transformers can serve as universal approximators for continuous operators between continuous function spaces \citep{kovachki2023neural}, they may require large network sizes when applied to target mappings that admit explicit ODE forms due to architectural mismatch. In contrast, Theorem 2 in this study demonstrates that the neural oscillator can directly approximate the derivative terms of the target ODEs, achieving the desired approximation results with more effective network sizes.}

\section{{\color{black} Numerical Studies}}
\label{sec4}
{\color{black} In this section, four numerical cases are presented to validate the convergence rates of the derived approximation error bounds and to illustrate the convergence properties of the neural oscillator. The code for all the numerical cases is available at: \url{https://github.com/ZifengH22/Upper-Approximation-Bounds-for-Neural-Oscillators}.}

\subsection{{\color{black} Case 1}}
\label{subsec4.1}
In {\color{black} Cases 1 and 2}, a Bouc-Wen nonlinear system driven by stochastic excitations is considered. A numerical study in \cite{zifeng2025} has shown that the neural oscillator can effectively learn the mapping relationship between the input excitation and the output response of a Bouc-Wen system exhibiting plastic deformation. The differential equation of a {\color{black} 5-degree-of-freedom} Bouc-Wen system is
\begin{equation}
    \left\{
    \begin{aligned}
        &\mathbf{M}{\color{black}\mathbf{X}''}(t) + \mathbf{C}{\color{black}\mathbf{X}'}(t) + \lambda{\mathbf{K}\mathbf{X}}(t) + (1 - \lambda ){\tilde{\mathbf{K}}\mathbf{Z}}(t) =  - {\mathbf{M}}{U_{\text{e}}}(t)\\
        &{{\color{black} {Z}'_i}}(t) = {\color{black} \tilde{X}'_i}(t) - \beta \left|{\color{black} \tilde{X}'_i}(t)\right|{\left|{Z_i}(t)\right|^{s - 1}}{Z_i}(t) - \gamma {\color{black} \tilde{X}'_i}(t){\left| {{Z_i}(t)}\right|^s},\;\;i = 1,2, \ldots,5
    \end{aligned}
    \right.,
    \label{eq:33}
\end{equation}
where ${\mathbf{X}} = \left[X_1(t), X_2(t),\dots,X_5(t)\right]^\top$, {\color{black} $\tilde{\bf{X}}(t) = [X_1(t), X_2(t)-X_1(t),\dots,X_5(t)-X_4(t)]^\top$}, ${\mathbf{Z}} = \left[Z_1(t), Z_2(t),\dots,Z_5(t)\right]^\top$, ${\mathbf{X}}(0)={\mathbf{X}}'(0)={\mathbf{Z}}(0)={\mathbf{0}}$,
\begin{equation}
    {\mathbf{M}} = 
    \begin{bmatrix}
        m & 0 & 0 & 0 & 0 \\
        0 & m & 0 & 0 & 0 \\
        0 & 0 & m & 0 & 0 \\
        0 & 0 & 0 & m & 0 \\
        0 & 0 & 0 & 0 & m
    \end{bmatrix},\;\;\;
    {\mathbf{K}} = 
    \begin{bmatrix}
        2k & -k & 0 & 0 & 0 \\
        -k & 2k & -k & 0 & 0 \\
        0 & -k & 2k & -k & 0 \\
        0 & 0 & -k & 2k & -k \\
        0 & 0 & 0 & -k & k
    \end{bmatrix},\;\;\;
    {\tilde{{\mathbf{K}}}} = 
    \begin{bmatrix}
        k & -k & 0 & 0 & 0 \\
        0 & k & -k & 0 & 0 \\
        0 & 0 & k & -k & 0 \\
        0 & 0 & 0 & k & -k \\
        0 & 0 & 0 & 0 & k
    \end{bmatrix},\;\;\;
    \label{eq:34}
\end{equation}
%%%%%%%%%%
% \begin{equation}
%     \begin{aligned}
%         \mathbf{M} &= 
%         \begin{bmatrix}
%         m & 0 & \cdots & 0 \\
%         0 & m & \cdots & 0 \\
%         \vdots & \vdots & \ddots & \vdots \\
%         0 & 0 & \cdots & m
%         \end{bmatrix}_{5\times5}, \\[0.8em]
%         \mathbf{K} &= 
%         \begin{bmatrix}
%         2k & -k & \cdots & 0 \\
%         -k & 2k & \cdots & 0 \\
%         \vdots & \vdots & \ddots & -k \\
%         0 & 0 & \cdots & 2k
%         \end{bmatrix}_{5\times5}, \\[0.8em]
%         \tilde{\mathbf{K}} &=
%         \begin{bmatrix}
%         k & -k & \cdots & 0 \\
%         0 & k & \cdots & 0 \\
%         \vdots & \vdots & \ddots & -k \\
%         0 & 0 & \cdots & k
%         \end{bmatrix}_{5\times5},
%     \end{aligned}
%     \label{eq:34}
% \end{equation}
%%%%%%%%%%%%%%%
$m=1382.4$ kilogram, $k=1.7 \times 10^6$ Newton/meter, $\mathbf{C}$ is determined using $\mathbf{M}$ and $\mathbf{K}$ such that the damping ratios of all response modes are 0.05, $\lambda = 0.01$, $\beta = 2$, $\gamma = 2$, and $s = 3$. $U_{\text{e}}(t)$ in Eq.~\eqref{eq:33} represents different stochastic excitations in {\color{black} Cases 1 and 2}. The ODE system in Eq.~\eqref{eq:33} is solved numerically using a fourth-order Runge-Kutta method \citep{griffiths2010numerical} with a time discretization of $\Delta t = 0.01$ seconds over the time interval $[0, 10]$. For the numerical implementation of training, the neural oscillator is discretized using a second-order Runge–Kutta scheme \citep{zifeng2025}. The approach for training the neural oscillator with input-output pair samples is briefly described in \ref{appE}.

In this case, $U_{\text{e}}(t)$ in Eq.~\eqref{eq:33} is a stochastic excitation
\begin{equation}
    U_{\text{e}}(t) = c_1\sin(0.4{\pi}t)+c_2\cos(0.8{\pi}t)+c_3\sin(1.2{\pi}t)+c_4\cos(1.6{\pi}t)+c_5\sin(2{\pi}t),
    \label{eq:35}
\end{equation}
where $c_j, j = 1, 2,\ldots, 5$ are are independent random variables uniformly distributed on the interval $\left[-35,35\right]$. $5 \times 10^4$ independent discrete-time samples $U_{\text{e},l}(t_i)$ of $U_{\text{e}}(t)$, where $i = 0, 1, 2, \dots, 999$ and $l = 1, 2, \dots, 5 \times 10^4$, are simulated. Their induced $5\times10^4$ samples $E_{X_5,l}(t_i)$ of the extreme value process $E_{X_5}(t) = \max_{\tau \in \left[0, t\right]}\left|X_5(\tau)\right|$ are computed by numerically solving Eq.~\eqref{eq:33}. The mapping from $U_{\text{e}}(t)$ to $E_{X_5}(t)$ is learned using the neural oscillator with the simulated samples $U_{\text{e},l}(t_i)$ and $E_{X_5,l}(t_i)$. 

{\color{black} In this case, five neural oscillators ${\it{\Pi}}_i\circ{\Phi}_{{\it{\Gamma}}_i}\left(\cdot\right)$, $i = 1, 2, \ldots, 5$, are utilized to validate the convergence rate in Theorem 1}. The sizes of the five one-hidden-layer MLPs ${\itGamma}_i(\cdot)$ are the same. Their {\color{black} input, hidden, and output layer widths} are 11, 10, and 5, respectively. The activation functions of the five MLPs $\itGamma_i(\cdot)$ are $\sigma_{\mathrm{ReLU}}(\cdot)$. The five MLPs ${\itPi}_i(\cdot)$ corresponding to the five neural oscillators share the same input, hidden, and output layer widths of 7, 10, and 1, respectively. The numbers of hidden layers in ${\itPi}_1(\cdot)$, ${\itPi}_2(\cdot)$, ${\itPi}_3(\cdot)$,
${\itPi}_4(\cdot)$, and ${\itPi}_5(\cdot)$ are 1, 2, 3, 4, and 5, respectively. The activation functions of the five MLPs ${\itPi}_i(\cdot)$ are the {\color{black} PReLU function} $\sigma_{\text{PReLU}}(x)=\max(0,x) + \alpha\min(0,x)$, where $\alpha$ is a parameter learned during training \citep{he2015delving}. Since $\sigma_{\mathrm{ReLU}}(\cdot)$ is a special case of $\sigma_{\text{PReLU}}(\cdot)$, the result in Theorem 1 also holds for the neural oscillators ${\it{\Pi}}_i\circ{\Phi}_{{\it{\Gamma}}_i}\left(\cdot\right)$ when ${\itPi}_i(\cdot)$ employs $\sigma_{\text{PReLU}}(\cdot)$. 

The $5 \times 10^4$ samples $E_{X_5,l}(t_i)$ of output $E_{X_5}(t)$ are arranged in a descending order with respect to their maximum values $\hat{E}_{X_5,l}=\max_{t_i\in\left[0,10\right]}E_{X_5,l}(t_i)$. Then, 10000 samples $E_{X_5,l}(t_i)$, whose maximum values $\hat{E}_{X_5,l}$ are approximately equally spaced from the largest one to the smallest one of the $5 \times 10^4$ $\hat{E}_{X_5,l}$, along with their corresponding input samples $U_{\text{e},l}(t_i)$, are selected. Among these 10000 sample pairs, the $5^\text{th}$, $10^\text{th}$, $15^\text{th}$,$\dots$, $10000^\text{th}$ pairs (total 2000 pairs) are selected as a validation dataset, and the remaining 8000 sample pairs constitute a training dataset. As this study is concerned solely with validating the approximation error bound rather than investigating the generalization performance, this sampling strategy is appropriate to build the training data. Each of the five neural oscillators is trained using the training approach outlined in \ref{appE} with the 8000 sample pairs. The {\color{black} $8^{\text{th}}$} power loss function $\ell_8$ in Eq.~\eqref{eq:e3} is employed to ensure that the neural oscillator adequately learns the target mapping with respect to the $L^\infty$-norm. The training phase comprises 10000 epochs, with a batch size of 2000 for each epoch. The initial learn rate is 0.005. The learning rate drop period is 100 epochs. The learning rate drop factor is 0.9772. To mitigate the risk of exploding gradients during training, the gradient norm clipping strategy \citep{pascanu2013difficulty} with a threshold of 1 is employed for the five MLPs $\itGamma_i(\cdot)$. For the five MLPs ${\itPi}_i(\cdot)$, their norm clipping thresholds are $1, \sqrt{203/92},\sqrt{314/92},\sqrt{425/92},\sqrt{536/92}$, respectively, where $92,203,314,425,$ and $536$ are the numbers of learnable parameters in ${\itPi}_1(\cdot)$, ${\itPi}_2(\cdot)$, ${\itPi}_3(\cdot)$, ${\itPi}_4(\cdot)$, and ${\itPi}_5(\cdot)$. Batch normalization \citep{ioffe2015batch} is applied to the five MLPs ${\itPi}_i(\cdot)$ to facilitate training due to the depth of their hidden layers. Since batch normalization only adds an affine transformation to the input of each hidden layer of the five MLPs ${\itPi}_i(\cdot)$, the result of Theorem 1 still holds for ${\itPi}_i(\cdot)$ with batch normalization. For the first neural oscillator with $\itGamma_1(\cdot)$ and ${\itPi}_1(\cdot)$, the initial values of the learnable network parameters are randomly selected. For the second neural oscillator with $\itGamma_2(\cdot)$ and ${\itPi}_2(\cdot)$, the initial parameter values of $\itGamma_2(\cdot)$ are taken directly from the parameters of $\itGamma_1(\cdot)$ after the last update. The initial parameter values for the first hidden layer and output layer of ${\itPi}_2(\cdot)$ are taken directly from the parameters of the corresponding layers in ${\itPi}_1(\cdot)$ after the last update. The second hidden layer of ${\itPi}_2(\cdot)$ is initialized as an identity mapping. The MLPs in the remaining three neural oscillators are initialized in the same manner. For each neural oscillator, the neural network achieving the smallest validation loss under the $L^\infty$-norm is selected as the final model.

\begin{figure}[t]
    \centering
    \begin{tikzpicture}
        \node (a) {\includegraphics[width=0.45\linewidth]{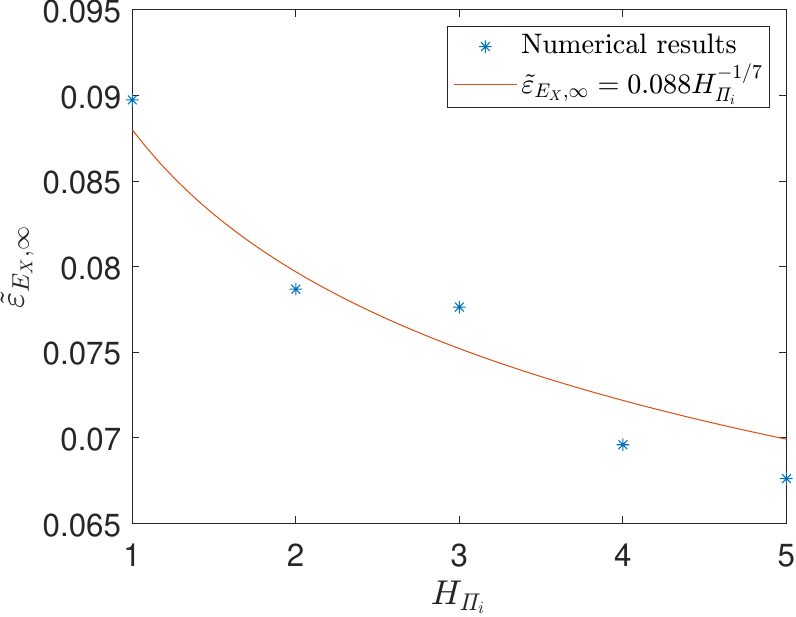}};
        \node[fill=white, inner sep=1.8pt, font=\bfseries\sffamily\small] at (a.north west) {(a)};
    \end{tikzpicture}\hfill
    \begin{tikzpicture}
        \node (b) {\includegraphics[width=0.45\linewidth]{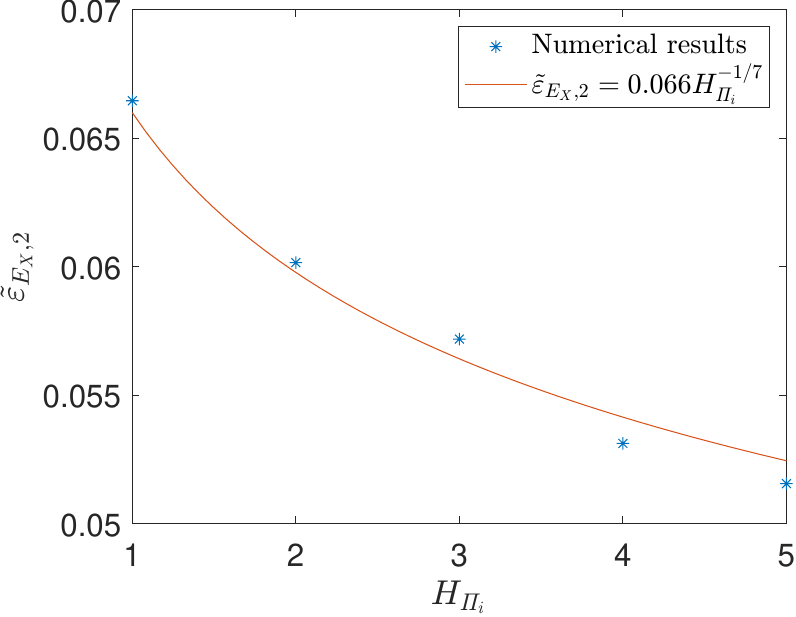}};
        \node[fill=white, inner sep=1.8pt, font=\bfseries\sffamily\small] at (b.north west) {(b)};
    \end{tikzpicture}
    
    \caption{{\color{black} Approximation errors $\tilde{\varepsilon}_{E_X,\infty}$ and $\tilde{\varepsilon}_{E_X,2}$ versus the number of hidden layers $H_{\itPi_i}$ of $\itPi_i(\cdot)$.} (a) $\tilde{\varepsilon}_{E_X,\infty}$ versus $H_{\itPi_i}$. (b) $\tilde{\varepsilon}_{E_X,2}$ versus $H_{\itPi_i}$.}
    \label{fig:2}
\end{figure}

For each trained neural oscillator, its $5 \times 10^4$ prediction samples $\tilde{E}_{X_5,l}(t_i)$ driven by the input samples $U_{\text{e},l}(t_i)$ are computed. The relative approximation errors under {\color{black} the} $L^\infty$-norm and $L^2$-norm, which are respectively denoted as  $\tilde{\varepsilon}_{E_X,\infty}$ and $\tilde{\varepsilon}_{E_X,2}$, are computed as
\begin{equation}
    \tilde{\varepsilon}_{E_X,\infty} = \frac{\max_{t_i \in \left[0,10\right],1 \leq l \leq 50000}\left|\tilde{E}_{X_5,l}(t_i)- E_{X_5,l}(t_i)\right|}{\max_{t_i \in \left[0,10\right],1 \leq l \leq 50000}\left|{\color{black}E_{X_5,l}}(t_i)\right|}
    \label{eq:36}
\end{equation}
and
\begin{equation}
    \tilde{\varepsilon}_{E_X,2} = \frac{\sqrt{\sum\limits_{l = 1}^{50000}{\sum\limits_{i = 0}^{999} \left[{{E}_{X_5,l}}(t_i) - \tilde{E}_{X_5,l}(t_i)\right]^2}}}{\sqrt{\sum\limits_{l = 1}^{50000}\sum\limits_{i = 0}^{999} E_{X_5,l}^2(t_i)}}.
    \label{eq:37}
\end{equation}
For the five trained neural oscillators, the decays of their $\tilde{\varepsilon}_{E_X,\infty}$ and $\tilde{\varepsilon}_{E_X,2}$ with respect to the increasing {\color{black} number of hidden layers $H_{{\itPi}_i}$ of $\itPi_i(\cdot)$} are shown in Figure {\color{black}\ref{fig:2}}. In this numerical case, each neural oscillator uses its second-order ODE to encode the input $U_{\text{e}}(t)$ and employs its MLP $\itPi_i(\cdot)$ to map the encoded features to the output function $\mathbf{y}(t)$. The results in Figure {\color{black}\ref{fig:2}} illustrate that the decays of $\tilde{\varepsilon}_{E_X,\infty}$ and {\color{black}$\tilde{\varepsilon}_{E_X,2}$} with {\color{black} the increasing number of hidden layers} $H_{{\itPi}_i}$ of $\itPi_i(\cdot)$ can be approximated by two curves with the same {\color{black} convergence rate} of $-1/7$. This value coincides with the theoretical {\color{black} convergence rate} $-1/\left[p(M_{\itGamma}+1+1)\right] = -1/\left[1\times(5+1)+1\right] = -1/7$ in Eq.~\eqref{eq:20} of Theorem 1. {\color{black} Such agreement validates the convergence rate established in Theorem 1.}

\subsection{Case 2}
\label{subsec4.2}
In this case, $U_{\text{e}}(t)$ in Eq.~\eqref{eq:33} is a stochastic Gaussian harmonizable earthquake ground motion acceleration whose Wigner-Ville spectrum is \citep{huang2024probability}
\begin{equation}
    {W_{\text{e}}}(t,f) = 2500{t^2}{f^2}\exp \left[ { - 0.3(1 + {f^2})t} \right].
    \label{eq:38}
\end{equation}
$5 \times 10^4$ independent discrete-time samples $U_{\text{e},l}(t_i)$ of $U_{\text{e}}(t)$ are simulated. Their induced $5\times10^4$ samples $X_{5,l}(t_i)$ of the response $X_5(t)$ are computed. In this case, the mapping from $U_{\text{e}}(t)$ to $X_5(t)$ is learned using the neural oscillator with the simulated samples $U_{\text{e},l}(t_i)$ and $X_{5,l}(t_i)$, and the {\color{black} convergence rate} in Theorem 2 is validated.

{\color{black} Six neural oscillators ${\it{\Pi}}_i\circ{\Phi}_{{\it{\Gamma}}_i}\left(\cdot\right)$, $i = 1, 2, \dots, 6$, are considered in this subsection}. The activation functions of all MLPs in the six neural oscillators are $\sigma_{\mathrm{ReLU}}(\cdot)$. The sizes of the six one-hidden-layer MLPs ${\itPi}_i(\cdot)$ of the six neural oscillators are the same. Their {\color{black} input, hidden, and output layer widths} are 12, 20, and 1, respectively. The six one-hidden-layer MLPs ${\itGamma}_i(\cdot)$ of the six neural oscillators share the same input and output layer widths of 21 and 10, respectively. The {\color{black} hidden-layer} widths of ${\itGamma}_1(\cdot)$, ${\itGamma}_2(\cdot)$, ${\itGamma}_3(\cdot)$,
${\itGamma}_4(\cdot)$, ${\itGamma}_5(\cdot)$, and ${\itGamma}_6(\cdot)$ are 2, 5, 10, 20, 30, and 40, respectively.

The $5 \times 10^4$ samples $U_{\text{e},l}(t_i)$ of input $U_{\text{e}}(t)$ are arranged in a descending order with respect to their standard deviations. Then the $1^{\text{st}}$, $26^{\text{th}}$, $51^\text{st}$,$\dots$, $49976^\text{th}$ input samples and their induced samples $X_{5,l}(t_i)$ of $X_5(t)$ (total 2000 input and response sample pairs) are selected for training. Among these 2000 sample pairs, the $5^\text{th}$, $10^\text{th}$, $15^\text{th}$,$\dots$, $2000^\text{th}$ pairs (total 400 pairs) are selected as a validation dataset, and the remaining 1600 sample pairs constitute a training dataset. Each of the six neural oscillators is trained using the training approach outlined in \ref{appE} with these 2000 sample pairs. The mean squared loss function $\ell_2$ in Eq.~\eqref{eq:e3} is used. The training phase comprises 5000 epochs, with a batch size of 1600 for each epoch. The initial learn rate is 0.02. The learning rate drop period is 100 epochs. The learning rate drop factor is set to 0.965. A gradient norm clipping strategy \citep{pascanu2013difficulty} with a threshold of 1 is applied to all MLPs. The parameters of all MLPs are randomly initialized. For each neural oscillator, the neural network after the last update is selected as the final model.

\begin{figure}[t]
    \centering
    \begin{tikzpicture}
        \node (a) {\includegraphics[width=0.45\linewidth]{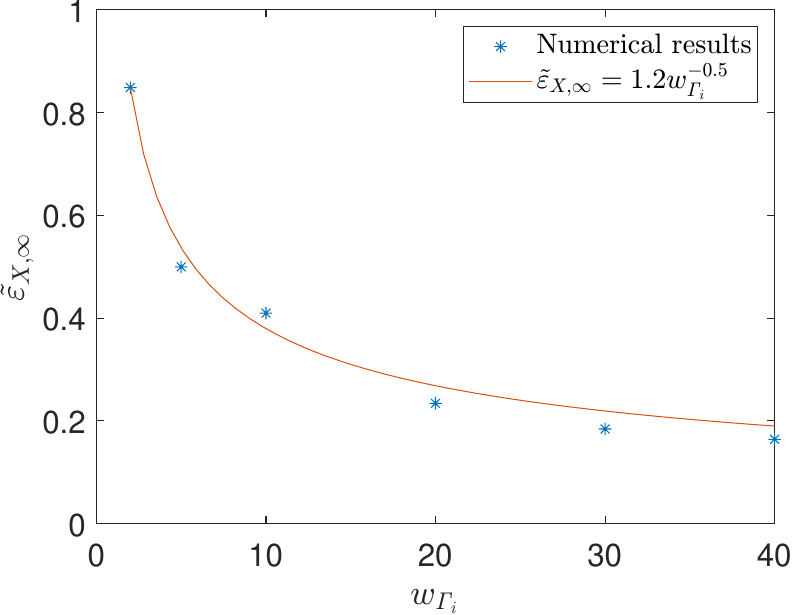}};
        \node[fill=white, inner sep=1.8pt, font=\bfseries\sffamily\small] at (a.north west) {(a)};
    \end{tikzpicture}\hfill
    \begin{tikzpicture}
        \node (b) {\includegraphics[width=0.45\linewidth]{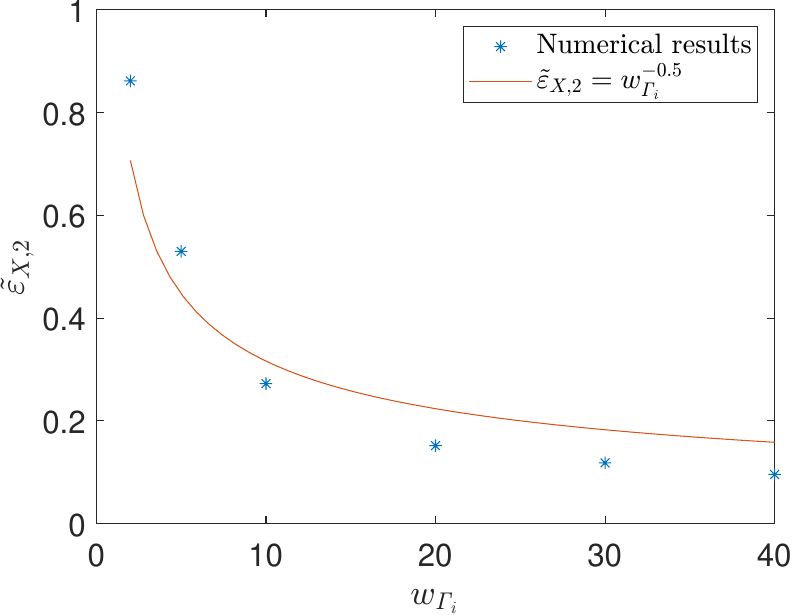}};
        \node[fill=white, inner sep=1.8pt, font=\bfseries\sffamily\small] at (b.north west) {(b)};
    \end{tikzpicture}
    
    \caption{{\color{black} Approximation errors $\tilde{\varepsilon}_{X,\infty}$ and $\tilde{\varepsilon}_{X,2}$ versus the hidden-layer width $w_{{\itGamma}_i}$ of $\itGamma_i(\cdot)$.} (a) $\tilde{\varepsilon}_{X,\infty}$ versus $w_{\itGamma_i}$. (b) $\tilde{\varepsilon}_{X,2}$ versus $w_{\itGamma_i}$.}
    \label{fig:3}
\end{figure}

For each trained neural oscillator, its $5 \times 10^4$ prediction samples $\tilde{X}_{5,l}(t_i)$ driven by the input samples $U_{\text{e},l}(t_i)$ are computed. The relative approximation errors $\tilde{\varepsilon}_{X,\infty}$ and $\tilde{\varepsilon}_{X,2}$ are computed by replacing $\tilde{E}_{X_5,l}(t_i)$ and $E_{X_5,l}(t_i)$ in {\color{black} Eqs.~\eqref{eq:36} and \eqref{eq:37}} with $\tilde{X}_{5,l}(t_i)$ and $X_{5,l}(t_i)$, respectively. For the six trained neural oscillators, the decays of their $\tilde{\varepsilon}_{X,\infty}$ and $\tilde{\varepsilon}_{X,2}$ with respect to the increase of {\color{black} the hidden-layer width} $w_{\itGamma_i}$ of $\itGamma_i(\cdot)$ are shown in Figure {\color{black}\ref{fig:3}}. The ODE system in Eq.~\eqref{eq:33} consists of five second-order ODEs with respect to $\mathbf{X}(t)$ and five first-order ODEs with respect to $\mathbf{Z}(t)$. In this case, the each neural oscillator uses its second-order ODE to approximate the ODE system in Eq.~\eqref{eq:33} and employs its $\itPi_i(\cdot)$ as a readout function. {\color{black} The results in Figure \ref{fig:3}(a) illustrate} that the decay of $\tilde{\varepsilon}_{X,\infty}$ with {\color{black} the increasing hidden-layer width} $w_{\itGamma_i}$ of $\itGamma_i(\cdot)$ can be well represented by a curve with a {\color{black} convergence rate} of $-0.5$. This value is the same as the {\color{black} convergence rate} in Eq.~\eqref{eq:31} of Theorem 2, where the approximation error is under {\color{black} the} $L^{\infty}$-norm. The {\color{black} convergence rate} of $\tilde{\varepsilon}_{X,2}$ under {\color{black} the} $L^2$-norm is slightly smaller than $-0.5$. These results demonstrate that the {\color{black} convergence rate} in Theorem 2 is reasonable.

\begin{figure}[t]
    \centering
    \begin{tikzpicture}
        \node (a) {\includegraphics[width=0.45\linewidth]{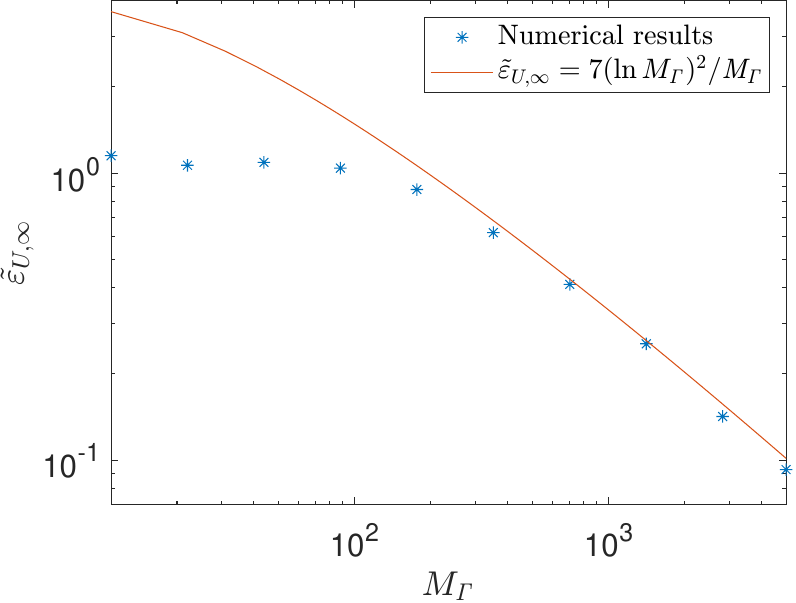}};
        \node[fill=white, inner sep=1.8pt, font=\bfseries\sffamily\small] at (a.north west) {(a)};
    \end{tikzpicture}\hfill
    \begin{tikzpicture}
        \node (b) {\includegraphics[width=0.45\linewidth]{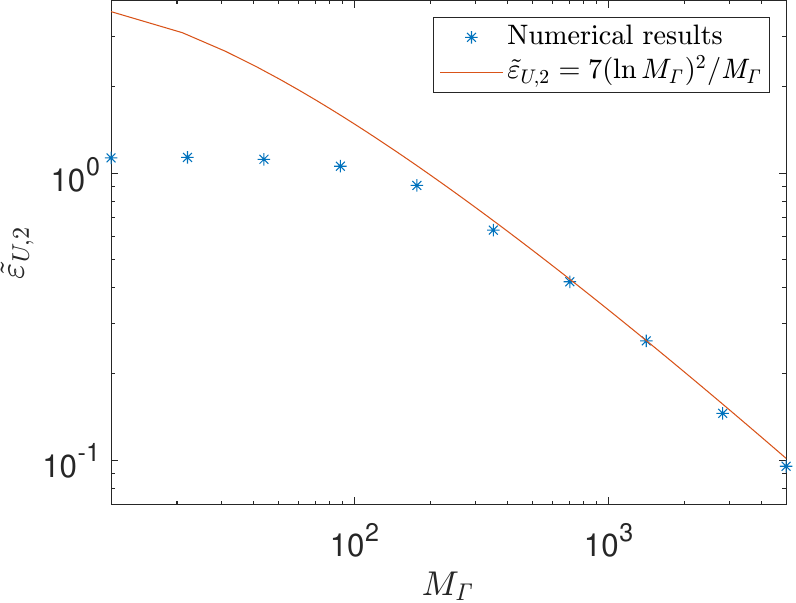}};
        \node[fill=white, inner sep=1.8pt, font=\bfseries\sffamily\small] at (b.north west) {(b)};
    \end{tikzpicture}
    
    \caption{{\color{black} Approximation errors $\tilde{\varepsilon}_{U,\infty}$ and $\tilde{\varepsilon}_{U,2}$ versus $M_{\itGamma}$. (a) $\tilde{\varepsilon}_{U,\infty}$ versus $M_{\itGamma}$. (b) $\tilde{\varepsilon}_{U,2}$ versus $M_{\itGamma}$.}}
    \label{fig:4}
\end{figure}

{\color{black}
\subsection{Case 3}
\label{subsec4.3}
In this subsection, a zero-mean Gaussian stochastic process $U_{\text{e}}(t)$ defined over $\left[0, 10\right]$ seconds is considered. Its correlation function is
\begin{equation}
    {C_{\text{e}}}(t_1,t_2) = \exp \left\{{-200\sin^2{\left[0.1\left(t_1 - t_2\right) \right]}}\right\}.
    \label{eq:39}
\end{equation}
The error bound  $O\left[\left(\ln{M_{\itGamma}}\right)^{2}/{M_{\it{\Gamma}}}\right]$ in Eq.~\eqref{eq:20} of Theorem 1 is numerically validated using this Gaussian process. Specifically, 1000 samples $U_{\text{e},l}(t_i)$ of $U_{\text{e}}(t)$ are simulated with $\Delta t = 0.001$ seconds, $l = 1, 2,\dots,1000$. Then, a set of layer widths $M_{\itGamma} = \left[10, 20, 40, 80, 160, 320, 640,1280, 2560, 5000\right]^{\top}$ is considered. For each value of $M_{\itGamma}$, the corresponding $v\left(M_{\itGamma}\right)$ is calculated using Eq.~\eqref{eq:14} with $r = \floor{\ln{M_{\itGamma}}}$, $c_K = 1.1$, and $T = 10$, and the corresponding 1000 approximations $\tilde{U}_{\text{e},l}(t_i)$ are calculated by
\begin{equation}
    {{\tilde{U}}_{\text{e},l}}(t_i) = {U_{\text{e},l}}(0) + \sum\limits_{n = 1}^{M_{\itGamma}} {{\alpha _n}\left\{ {{\mathcal{L}_T}{U_{\text{e},l}}(\omega_n) + \frac{{U_{\text{e},l}}(0)}{{{\omega _n}}}\left[ {\cos ({\omega _n}T) - 1} \right]} \right\}\sin \left[ {{\omega _n}(T - t_i) - {\theta _n}} \right]},
    \label{eq:40}
\end{equation}
where $\omega_n = n\pi/\left(2T\right)$, ${\mathcal{L}_T}{U_{\text{e},l}}(\omega_n)$ is the sine transform of $U_{\text{e},l}(t_i)$ calculated by Eq.~\eqref{eq:10}, $\alpha_n$ and $\theta_n$ are respectively the scaled Fourier modulus and phase of $\rho_{v\left(M_{\itGamma}\right)}(\tau)$ in Eq.~\eqref{eq:a3} that are calculated according to Eqs.~\eqref{eq:a12} and \eqref{eq:a13}. The relative approximation errors $\tilde{\varepsilon}_{U,\infty}$ under the $L^{\infty}$-norm and $\tilde{\varepsilon}_{U,2}$ under the $L^{2}$-norm are computed by replacing $\tilde{E}_{X_5,l}(t_i)$ and $E_{X_5,l}(t_i)$ in Eqs.~\eqref{eq:36} and \eqref{eq:37} with $\tilde{U}_{\text{e},l}(t_i)$ and $U_{\text{e},l}(t_i)$, respectively. The numerical results in Figure~\ref{fig:4} show that the observed convergence behaviors of both 
$\tilde{\varepsilon}_{U,\infty}$ and $\tilde{\varepsilon}_{U,2}$ are asymptotically consistent 
with the theoretical error bound 
$O\!\left[(\ln M_{\itGamma})^{2}/M_{\itGamma}\right]$ in Theorem~1.}

{\color{black}
\subsection{Case 4}
\label{subsec4.4}
In this case, a 200-degree-of-freedom Duffing system subjected to the same earthquake excitation in Case 2 is considered to illustrate the convergence properties of the neural oscillator for high-dimensional systems. Its governing differential equation is
\begin{equation}
    \mathbf{M}\mathbf{X}''(t) + \mathbf{C}\mathbf{X}'(t) + \mathbf{F}\left[\mathbf{X}(t)\right] =  - {\mathbf{M}}{U_{\text{e}}}(t),
    \label{eq:41}
\end{equation}
where ${\mathbf{X}} = \left[X_1(t), X_2(t),\dots,X_{200}(t)\right]^\top$, $\mathbf{X}(0)=\mathbf{X}'(0)=\mathbf{0}$, $\mathbf{M} \in \mathbb{R}^{200 \times 200}$ is a diagonal matrix $\mathbf{M} = \text{diag}(1000,1000,$ $\dots,1000)$, $\mathbf{C} \in \mathbb{R}^{200 \times 200}$ is a tridiagonal matrix with $40020$ on the main diagonal and $-20000$ on the sub- and super-diagonals, and $\mathbf{F}\left[\mathbf{X} (t)\right] = \left\{F_{1}\left[\mathbf{X} (t)\right],F_{2}\left[\mathbf{X} (t)\right],\dots,F_{200}\left[\mathbf{X} (t)\right] \right\}^{\top}$ is the restoring force vector with $F_{i}\left[\mathbf{X} (t)\right]$ being
\begin{equation}
    F_{i}\left[\mathbf{X} (t)\right] = \left\{ 
    \begin{aligned}
        & f_{i}\left[\mathbf{X} (t)\right]- f_{i+1}\left[\mathbf{X} (t)\right],\;\;\;\; i < 200\\
        & f_{i}\left[\mathbf{X} (t)\right],\;\;\;\;\;\;\;\;\;\;\;\;\;\;\;\;\;\;\;\;\;\; i = 200,
    \end{aligned}
    \right.
    \label{eq:42}
\end{equation}
and
\begin{equation}
    f_{i}\left[\mathbf{X} (t)\right] = \left\{ 
    \begin{aligned}
        &10^7\left[X_{i}(t) - X_{i-1}(t)\right]+10^7\left[X_{i}(t) - X_{i-1}(t)\right]^3,\;\;\;\; i > 1\\
        &10^7X_{i}(t)+10^7X_{i}^3(t),\;\;\;\;\;\;\;\;\;\;\;\;\;\;\;\;\;\;\;\;\;\;\;\;\;\;\;\;\;\;\;\;\;\;\;\;\;\; i = 1.
    \end{aligned}
    \right.
    \label{eq:43}
\end{equation}
The ODE system in Eq.~\eqref{eq:41} is solved numerically using a fourth-order Runge-Kutta method \citep{griffiths2010numerical} with a time discretization of $\Delta t = 0.005$ seconds over the time interval $[0, 20]$.

In this case, seven neural oscillators ${\it{\Pi}}_i\circ{\Phi}_{{\it{\Gamma}}_i}\left(\cdot\right)$, $i = 1, 2, \dots, 7$, are utilized to learn the mapping from $U_{\text{e}}(t)$ to $X_{200}(t)$. The activation functions of all MLPs in the seven neural oscillators are $\sigma_{\mathrm{ReLU}}(\cdot)$. The seven one-hidden-layer MLPs ${\itGamma}_i(\cdot)$ of the seven neural oscillators share the same input and output layer widths of 41 and 20, respectively. The seven one-hidden-layer MLPs ${\itPi}_i(\cdot)$ share the same input and output layer widths of 22 and 1, respectively. For each neural oscillator ${\it{\Pi}}_i\circ{\Phi}_{{\it{\Gamma}}_i}\left(\cdot\right)$, ${\itGamma}_i(\cdot)$ and ${\itPi}_i(\cdot)$ share the same hidden-layer width $w_i$. The values of $w_i$ corresponding to $i = 1, 2, \dots, 7$ are 2, 5, 10, 20, 30, 40, and 60, respectively.

The $10^4$ samples $U_{\text{e},l}(t_i)$ of input $U_{\text{e}}(t)$ are simulated and arranged in a descending order with respect to their standard deviations. Then the $1^{\text{st}}$, $6^{\text{th}}$, $11^\text{st}$,$\dots$, $9996^\text{th}$ input samples and their induced samples $X_{200,l}(t_i)$ of $X_{200}(t)$ (total 2000 input and response sample pairs) are selected for training. Among these 2000 sample pairs, the $5^\text{th}$, $10^\text{th}$, $15^\text{th}$,$\dots$, $2000^\text{th}$ pairs (total 400 pairs) are selected as a validation dataset, and the remaining 1600 sample pairs constitute a training dataset. Each of the seven neural oscillators is trained using the training approach outlined in \ref{appE} with these 2000 sample pairs. The mean squared loss function $\ell_2$ in Eq.~\eqref{eq:e3} is used. The training phase comprises 10000 epochs, with a batch size of 1600 for each epoch. The initial learn rate is 0.02. The learning rate drop period is 100 epochs. The learning rate drop factor is set to 0.96. A gradient norm clipping strategy \citep{pascanu2013difficulty} with a threshold of 1 is applied to all MLPs. The parameters of all MLPs are randomly initialized. For each neural oscillator, the neural network after the last update is selected as the final model.

\begin{figure}[t]
    \centering
    \begin{tikzpicture}
        \node (a) {\includegraphics[width=0.45\linewidth]{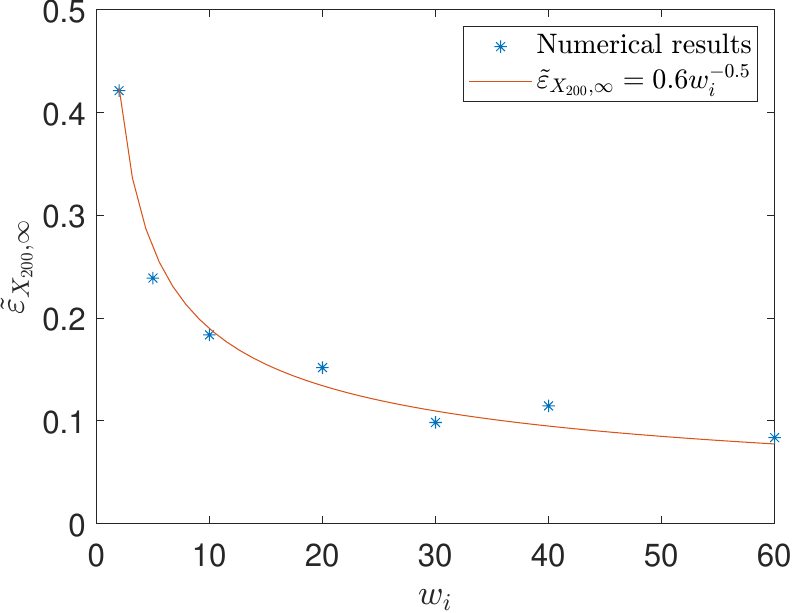}};
        \node[fill=white, inner sep=1.8pt, font=\bfseries\sffamily\small] at (a.north west) {(a)};
    \end{tikzpicture}\hfill
    \begin{tikzpicture}
        \node (b) {\includegraphics[width=0.45\linewidth]{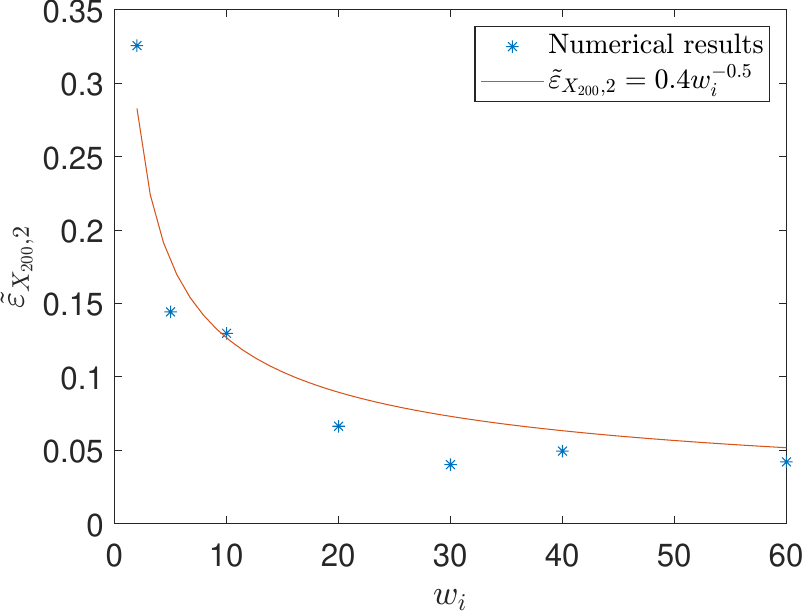}};
        \node[fill=white, inner sep=1.8pt, font=\bfseries\sffamily\small] at (b.north west) {(b)};
    \end{tikzpicture}
    
    \caption{{\color{black} Approximation errors $\tilde{\varepsilon}_{X_{200},\infty}$ and $\tilde{\varepsilon}_{X_{200},2}$ versus the hidden-layer widths $w_{i}$ of $\itGamma_i(\cdot)$ and $\itPi_i(\cdot)$. (a) $\tilde{\varepsilon}_{X_{200},\infty}$ versus $w_{i}$. (b) $\tilde{\varepsilon}_{X_{200},2}$ versus $w_{i}$.}}
    \label{fig:5}
\end{figure}

For each trained neural oscillator, its $10^4$ prediction samples $\tilde{X}_{200,l}(t_i)$ driven by the input samples $U_{\text{e},l}(t_i)$ are computed. The relative approximation errors $\tilde{\varepsilon}_{X_{200},\infty}$ and $\tilde{\varepsilon}_{X_{200},2}$ are computed by replacing $\tilde{E}_{X_5,l}(t_i)$ and $E_{X_5,l}(t_i)$ in Eqs.~\eqref{eq:36} and \eqref{eq:37} with $\tilde{X}_{200,l}(t_i)$ and $X_{200,l}(t_i)$, respectively. For the seven trained neural oscillators, the decays of their $\tilde{\varepsilon}_{X_{200},\infty}$ and $\tilde{\varepsilon}_{X_{200},2}$ with respect to the increase of the hidden-layer widths $w_{i}$ of $\itGamma_i(\cdot)$ and $\itPi_i(\cdot)$ are shown in Figure \ref{fig:5}. The results illustrate that the decays of $\tilde{\varepsilon}_{X_{200},\infty}$ and $\tilde{\varepsilon}_{X_{200},2}$ with respect to $w_{i}$ can be approximately represented by two curves with the same convergence rate of $-0.5$, that is consistent with the result in Theorem 2. Furthermore, the relative errors $\tilde{\varepsilon}_{X_{200},\infty} < 0.1$ and $\tilde{\varepsilon}_{X_{200},2} < 0.05$ are achieved by the model with $w_{i} = 30$. The probability distribution function (PDF) and cumulative distribution function (CDF) of the extreme value $E_{\text{max}} = \max_{t \in \left[0,20\right]}\left|X_{200}(t) \right|$ estimated from the $10^4$ samples $\tilde{X}_{200,l}(t_i)$ by the the model with $w_{i} = 30$ are consistent with the results from the samples $X_{200,l}(t_i)$, as shown in Figure \ref{fig:6}. These results suggest that, since the number of the dominant modes governing the response of the high-dimensional system in Eq.~\eqref{eq:41} is limited, a neural oscillator with a moderate network size can effectively approximate the system. From Figure \ref{fig:5}, the comparison between the errors from the model with $w_{i} = 30$ and those from the model with $w_{i} = 60$ indicates that, given the convergence rate of $-0.5$, the expressive power of the neural oscillator improves sublinearly with respect to the hidden-layer width $w_i$. As a result, the marginal benefit of increasing model size gradually diminishes. Therefore, excessively large networks may not yield proportional performance improvements, and their practical advantage may be limited relative to their computational cost.}

\begin{figure}[t]
    \centering
    \begin{tikzpicture}
        \node (a) {\includegraphics[width=0.45\linewidth]{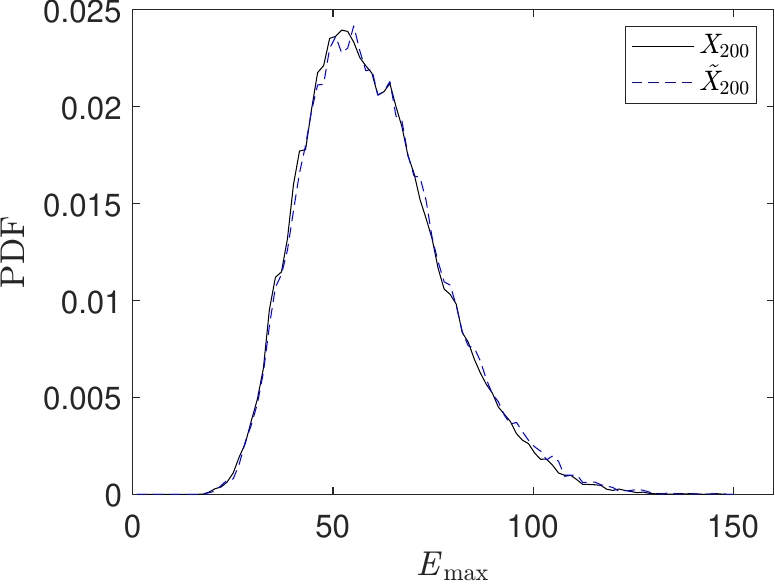}};
        \node[fill=white, inner sep=1.8pt, font=\bfseries\sffamily\small] at (a.north west) {(a)};
    \end{tikzpicture}\hfill
    \begin{tikzpicture}
        \node (b) {\includegraphics[width=0.45\linewidth]{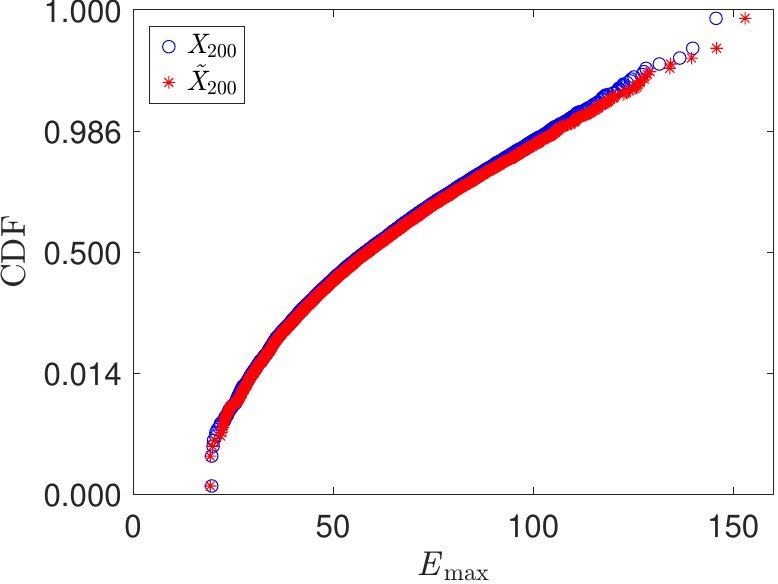}};
        \node[fill=white, inner sep=1.8pt, font=\bfseries\sffamily\small] at (b.north west) {(b)};
    \end{tikzpicture}
    
    \caption{{\color{black} PDF and CDF of $E_{\text{max}}$. (a) PDF of $E_{\text{max}}$. (b) CDF of $E_{\text{max}}$.}}
    \label{fig:6}
\end{figure}

\section{Conclusions}
In this study, two upper approximation bounds of the neural oscillator consisting of a second-order ODE followed by an MLP are derived for approximating the causal and uniformly continuous operators between continuous temporal function spaces and for approximating the uniformly asymptotically incrementally stable second-order dynamical systems. The established proof method of the upper approximation bound for approximating the causal continuous operators can also be directly applied to the SS models consisting of a linear time-continuous complex RNN followed by an MLP. Theorem 1 indicates that for the causal continuous operators, the {\color{black} convergence rate} in the approximation error polynomially diminishes as the {\color{black} hidden-layer width} of the MLP ${\it{\Gamma}}(\cdot)$ increases. Thus, the neural oscillator for approximating the causal continuous operators between continuous temporal function spaces also faces the curse of parametric complexity. Theorem 2 demonstrates that for approximating the uniformly asymptotically incrementally stable second-order dynamical systems, the neural oscillator can provide an approximation error scaling polynomially with the reciprocals of the widths of the utilized MLPs. This implies that the curse of parametric complexity can be overcome in this situation. {\color{black} Four numerical cases are presented to validate the convergence rates of the derived approximation error bounds and to illustrate the convergence properties of the neural oscillator. The results show that the convergence rate with respect to the number of hidden layers of the MLP $\itPi(\cdot)$ in Theorem 1 and those with respect to the hidden-layer widths of the MLPs $\itGamma(\cdot)$ and $\itPi(\cdot)$ in Theorems 1 and 2 are consistent with the numerical results. A neural oscillator with a moderate network size can effectively approximate high-dimensional systems characterized by a small number of dominant modes.}

\section*{CRediT authorship contribution statement}
{\bf Zifeng Huang:} Conceptualization, Methodology, Software, Writing-reviewing \& editing, Project administration. {\bf Konstantin M. Zuev:} Methodology, Writing-reviewing \& editing. {\bf Yong Xia:} Writing-reviewing \& editing, Supervision, Project administration. {\bf Michael Beer:} Supervision, Project administration.

\section*{Declaration of competing interest}
The authors declare that they have no known competing financial interests or personal relationships that could have appeared to influence the work reported in this paper.

\section*{Data and Code Availability}
{\color{black} All code used for the numerical experiments in this study is available at: 
\url{https://github.com/ZifengH22/Upper-Approximation-Bounds-for-Neural-Oscillators}. 
All data presented in this study were generated using this code.}

\section*{Acknowledgments}
This work is supported by the Joint Research Centre for Marine Infrastructure at The Hong Kong Polytechnic University and the German Research Foundation (Project No. 554325683 and HU 3418/2-1). The opinions and conclusions presented in this paper are entirely those of the authors.

\appendix
\section{Proofs for Lemmas and Theorem 1 in Section \ref{subsec3.1}}
\label{appA}

\subsection{Proof of Lemma 1}
\label{appA.1}
{\bf Proof:} Given an arbitrary $\mathbf{u}(t) = [u_1(t), u_2(t),..., u_p(t)]^{\top} \in K$, since $K$ is compact, based on Assumption 2, there exists a continuous monotonic modulus of continuity $\mathbf{\upphi}_K(t)$ such that for arbitrary $0 \leq t_1 \leq t_2 \leq T$ and all $\mathbf{u}(t) \in K$, it satisfies $|\mathbf{u}(t_1) - \mathbf{u}(t_2)| \leq |{\mathbf{\upphi}}_K(t_2 - t_1)|$.
Define $\mathbf{f}_t(\tau) = \mathbf{u}(t-\tau) - \mathbf{u}(0)$ for $\tau \in [0, t]$ and
\begin{equation}
    {\tilde{\mathbf{f}}}_t(\tau ) =
    \left\{
    \begin{aligned}
    &{\bf{f}}_t(\tau ),\;\;\;\;\;\;\;\;\;\;\;0 \le \tau  \le t\\
    &-{{\bf{f}}_t}( - \tau ),\;\;\;\; - t \le \tau  < 0\\
    &{\bf{0}},\;\;\;\;\;\;\;\;\;\;\;\;\;\;\;\;\;{\rm{otherwise}}
    \end{aligned},
    \right.
    \label{eq:a1}
\end{equation}
${\tilde{\mathbf{f}}}_t(\tau)$ is odd and compactly supported, and its Fourier transform $\mathcal{H}{{{\tilde{\mathbf{f}}}}_t}(\omega )$ is
\begin{equation}
    \begin{aligned}
    &\mathcal{H}{{{\tilde{\mathbf{f}}}}_t}(\omega ) = \int_{ - \infty }^{ + \infty } {{e^{ - {\rm{i}}\omega \tau }}{{{\tilde{\mathbf{f}}}}_t}(\tau ){\rm{d}}\tau }  =  - {\rm{i}}\int_{ - t}^t {\sin (\omega \tau ){{{\tilde{\mathbf{f}}}}_t}(\tau ){\rm{d}}\tau } \\
    &\;\;\;\;\;\;\;\;\;\;\; =  - 2{\rm{i}}\int_0^t {\sin (\omega \tau ){{\bf{f}}_t}(\tau ){\rm{d}}\tau }  =  - 2{\rm{i}}\int_0^t {\sin (\omega \tau )\left[ {{\bf{u}}(t - \tau ) - {\bf{u}}(0)} \right]{\rm{d}}\tau } \\
    &\;\;\;\;\;\;\;\;\;\;\; =  - 2{\rm{i}}\left\{ {{\mathcal{L}_t}{\bf{u}}(\omega ) + \frac{{{\bf{u}}(0)}}{\omega }\left[ {\cos (\omega t) - 1} \right]} \right\}.
    \end{aligned}
    \label{eq:a2}
\end{equation}

Consider a smooth bump mollifier function $\rho_v(\tau)$
\begin{equation}
    {\rho _v}(\tau ) = 
    \left\{
    \begin{aligned}
    &c_v^{ - 1}{e^{{{ - {v^2}} \mathord{\left/
    {\vphantom {{ - {v^2}} {\left[ {{v^2} - {{(2\tau  + v)}^2}} \right]}}} \right.
    \kern-\nulldelimiterspace} {\left[ {{v^2} - {{(2\tau  + v)}^2}} \right]}}}},\;\;\; - v \le \tau  \le 0\\ &0,\;\;\;\;\;\;\;\;\;\;\;\;\;\;\;\;\;\;\;\;\;\;\;\;\;\;\;{\rm{otherwise}}
    \end{aligned} 
    \right.,
    \label{eq:a3}
\end{equation}
where $v$ is an arbitrary coefficient satisfying $0 < v < T$ and $c_v$ is a normalization constant depending on $v$ such that $\int_{-v}^{0}\rho_v(\tau)\mathrm{d}\tau = 1$. The mollification $\tilde{\mathbf{f}}_v(\tau)$ of $\tilde{\mathbf{f}}_t(\tau)$ is
\begin{equation}
    {{\tilde{\mathbf{f}}}_v}(\tau ) = \int_{ - \infty ( - v)}^{ + \infty (0)} {{{{\tilde{\mathbf{f}}}}_t}(\tau  - \lambda ){\rho _v}(\lambda ){\rm{d}}\lambda }.
    \label{eq:a4}
\end{equation}
$\tilde{\mathbf{f}}_v(\tau)$ is employed to approximate $\tilde{\mathbf{f}}_t(\tau) = \mathbf{u}(t-\tau) - \mathbf{u}(0)$ over $\tau \in [0,t]$ and their difference is bounded by
\begin{equation}
    \begin{aligned}
    &\left| {{{{\tilde{\mathbf{f}}}}_v}(\tau ) - \left[ {{\bf{u}}(t - \tau ) - {\bf{u}}(0)} \right]} \right| = \left| {{{{\tilde{\mathbf{f}}}}_v}(\tau ) - {{{\tilde{\mathbf{f}}}}_t}(\tau )} \right| = \left| {\int_{ - v}^0 {\left[ {{{{\tilde{\mathbf{f}}}}_t}(\tau  - \lambda ) - {{{\tilde{\mathbf{f}}}}_t}(\tau )} \right]{\rho _v}(\lambda ){\rm{d}}\lambda } } \right|\\
    &\;\;\;\;\;\;\;\;\;\;\;\;\;\;\;\;\;\;\;\;\;\;\;\;\;\;\;\;\;\;\;\;\;\;\;\;\; \le \int_{ - v}^0 {\left| {{{{\tilde{\mathbf{f}}}}_t}(\tau  - \lambda ) - {{{\tilde{\mathbf{f}}}}_t}(\tau )} \right|{\rho _v}(\lambda ){\rm{d}}\lambda } \mathop  \le \limits^{(a)} \int_{ - v}^0 {\left|{\mathbf{\upphi}_K(v)}\right|{\rho _v}(\lambda ){\rm{d}}\lambda }  = \left|{\mathbf{\upphi}_K(v)}\right|,
    \end{aligned}
    \label{eq:a5}
\end{equation}
where Step (\textit{a}) holds from $|\mathbf{u}(t_1) - \mathbf{u}(t_2)| \leq \left|{\mathbf{\upphi}}_K(t_2 - t_1)\right|$ and $\left|{\mathbf{\upphi}}_K(t)\right|$ being non-negative and monotonically increasing.

{\color{black} Since the compactly supported $\rho_v(\tau)$ belongs to a Gevrey class of order 2 \citep{rodino1993linear} and it has arbitrary finite-order continuous derivatives, the $r^{\text{th}}$-order derivative ${\rm{d}}^r{\tilde{\mathbf{f}}}_v(\tau)/{\rm{d}}\tau^r$ of ${\tilde{\mathbf{f}}}_v(\tau)$ over $\tau \in [-2T,2T]$ for all $t \in [0,T]$ and all positive $r \in \mathbb{Z}$ is bounded by
\begin{equation}
    \begin{aligned}
    &\left|\frac{{\rm{d}}^r{\tilde{\mathbf{f}}}_v(\tau)}{{\rm{d}}\tau^r}\right| = \left| {\frac{{\rm{d}}^r}{{\rm{d}}\tau^r}\int_{ - \infty }^{ + \infty } {{{{ \tilde{\mathbf{f}}}}_t}(\lambda ){\rho _v}(\tau  - \lambda ){\rm{d}}\lambda } } \right| \le \int_{ - \infty }^{ + \infty } {\left| {{{{ \tilde{\mathbf{f}} }}_t}(\lambda )} \right|\left| {\frac{{\rm{d}}^r}{{\rm{d}}\tau^r}{\rho _v}(\tau  - \lambda )} \right|{\rm{d}}\lambda } \\ 
    &\;\;\;\;\;\;\;\;\;\;\;\; \le \int_\tau ^{\tau  + v} {\left| {{{{ \tilde{\mathbf{f}} }}_t}(\lambda )} \right|\mathop {\max }\limits_{t \in [ - v,0]} \left| {\frac{{\rm{d}}^r}{{\rm{d}}\tau^r}{{\rho}_v}(t)} \right|{\rm{d}}\lambda }  \le \int_0^v {\mathop {\max }\limits_{t \in [0,T]} \left| {{\bf{u}}(t) - {\bf{u}}(0)} \right|\mathop {\max }\limits_{t \in [ - v,0]} \left| {\frac{{\rm{d}}^r}{{\rm{d}}\tau^r}{{\rho}_v}(t)} \right|{\rm{d}}\lambda } \\
    &\;\;\;\;\;\;\;\;\;\;\;\; \mathop  \le \limits^{(a)} v\left|{\mathbf{\upphi}_K(T)}\right|\mathop {\max }\limits_{t \in [ - v,0]} \left| {\frac{{\rm{d}}^r}{{\rm{d}}\tau^r}{{\rho}_v}(t)} \right|\mathop  \le \limits^{(b)} \left|{\mathbf{\upphi}_K(T)}\right|\frac{c_{\rho}^{r+1}\left(r!\right)^2}{v^r},
    \end{aligned}
    \label{eq:a6}
\end{equation}
where Step (\textit{a}) employs $|\mathbf{u}(t)-\mathbf{u}(0)| \leq \left|{\mathbf{\upphi}}_K(t)\right| \leq \left|{\mathbf{\upphi}}_K(T)\right|$, Step (\textit{b}) holds becasue $\rho_v(\tau)$ belongs to a Gevrey class of order 2 \citep{rodino1993linear}, and $c_{\rho} > 1$ is an independent constant. The bound of ${\rm{d}}^r{\tilde{\mathbf{f}}}_v(\tau)/{\rm{d}}\tau^r$ in Eq.~\eqref{eq:a6} is independent of \textit{t} and $\tau$.

$\tilde{\mathbf{f}}_v(\tau)$, $\tilde{\mathbf{f}}_t(\tau)$, and $\rho_v(\tau)$ are compactly supported over $\tau \in [-2T,2T]$ for all $t \in [0,T]$ and all $v \in \left(0, T\right)$. $\tilde{\mathbf{f}}_v(\tau)$ is odd with respect to $\tau = -0.5v$ because $\tilde{\mathbf{f}}_t(\tau)$ is an odd function and $\rho_v(\tau)$ is symmetric with respect to $\tau = -0.5v$, thereby $\int_{-\infty}^{+\infty}\tilde{\mathbf{f}}_v(\tau)\mathrm{d}\tau = \int_{-2T}^{2T}\tilde{\mathbf{f}}_v(\tau)\mathrm{d}\tau = \mathbf{0}$. Then, $\tilde{\mathbf{f}}_v(\tau)$ over $\tau \in [-2T,2T]$ can be expressed as
\begin{equation}
    {{\tilde{\mathbf{f}}}_v}(\tau ) = \frac{1}{2T}\sum\limits_{n = 1}^{+\infty} {{\mathop{\rm Re}\nolimits} \left[ {{e^{{\rm{i}}{\omega _n}\tau }}\mathcal{H}{{{\tilde{\mathbf{f}}}}_v}({\omega _n})} \right]} \mathop  = \limits^{(a)} \frac{1}{2T}\sum\limits_{n = 1}^{+\infty} {{\mathop{\rm Re}\nolimits} \left[ {{e^{{\rm{i}}{\omega _n}\tau }}\mathcal{H}{{{\tilde{\mathbf{f}}}}_t}({\omega _n})\mathcal{H}{\rho _v}({\omega _n})} \right]},
    \label{eq:a7}
\end{equation}
where $\omega_n = \pi n/{(2T)}$, $\mathcal{H}{{\tilde{\mathbf{f}}}}_v(\omega _n)$ is the Fourier transform of $\tilde{\mathbf{f}}_v(\tau)$
\begin{equation}
    \mathcal{H}{{\tilde{\mathbf{f}}}}_v(\omega _n)=\int_{-2T}^{2T}e^{-\mathrm{i}\omega_n\tau}\tilde{\mathbf{f}}_v(\tau){\mathrm{d}\tau},
    \label{eq:a8}
\end{equation}
$\mathcal{H}\rho_v(\omega_n)$ is the Fourier transform of $\rho_v(\tau)$
\begin{equation}
   \mathcal{H}\rho_v(\omega_n)=\int_{-2T}^{2T}e^{-\mathrm{i}\omega_n\tau}\rho_v(\tau){\mathrm{d}\tau},
   \label{eq:a9}
\end{equation}
and Step (\textit{a}) is valid because $\tilde{\mathbf{f}}_v(\tau)$ is the convolution between $\tilde{\mathbf{f}}_t(\tau)$ and $\rho_v(\tau)$.

Eq.~\eqref{eq:a6} indicates that $\left|{\mathbf{\upphi}_K(T)}\right|c_{\rho}^{r+1}\left(r!\right)^2v^{-r}$ is a Lipschitz constant of ${\rm{d}}^{r-1}{\tilde{\mathbf{f}}}_v(\tau)/{\rm{d}}\tau^{r-1}$. Then, according to Corollary III and Theorem X of \citet{jackson1930theory}, for every positive integer \textit{M} larger than two and every positive integer $r$, it can be obtained
\begin{equation}
    \left| {{{ \tilde{\bf f} }}_v}(\tau ) - \frac{1}{2T}\sum\limits_{n = 1}^{M} {{\mathop{\rm Re}\nolimits} \left[ {{e^{{\rm{i}}{\omega _n}\tau }}\mathcal{H}{{{\tilde{\mathbf{f}}}}_t}({\omega _n})\mathcal{H}{\rho _v}({\omega _n})} \right]} \right| \le \frac{2^rc_F^rT^r\left|{\mathbf{\upphi}_K(T)}\right|c_{\rho}^{r+1}\left(r!\right)^2\ln{M}}{\pi^rv^rM^r},
    \label{eq:a10}
\end{equation}
where $c_F > 1$ is an independent constant.

Finally, based on Eqs.~\eqref{eq:a5} and \eqref{eq:a10}, for every positive integer \textit{M} larger than two and every positive integer $r$, $\mathbf{u}(t-\tau)$ over $\tau \in [0,t]$ satisfies
\begin{equation}
    \begin{aligned}
        &\left| {{\bf{u}}(t - \tau ) - \left\{ {\frac{1}{2T}\sum\limits_{n = 1}^{M} {{\mathop{\rm Re}\nolimits} \left[ {{e^{{\rm{i}}{\omega _n}\tau }}\mathcal{H}{{{\tilde{\mathbf{f}}}}_t}({\omega _n})\mathcal{H}{\rho _v}({\omega _n})} \right]}  + {\bf{u}}(0)} \right\}} \right|\\
        &\;\;\;\; = \left| {{\bf{u}}(t - \tau ) - {\bf{u}}(0) - \frac{1}{2T}\sum\limits_{n = 1}^{M} {{\mathop{\rm Re}\nolimits} \left[ {{e^{{\rm{i}}{\omega _n}\tau }}\mathcal{H}{{{\tilde{\mathbf{f}}}}_t}({\omega _n})\mathcal{H}{\rho _v}({\omega _n})} \right]} } \right|\\
        &\;\;\;\; \le \left| {{\bf{u}}(t - \tau ) - {\bf{u}}(0) - {{{ \tilde{\mathbf{f}} }}_v}(\tau )} \right| + \left| {{{{ \tilde{\mathbf{f}} }}_v}(\tau ) - \frac{1}{2T}\sum\limits_{n = 1}^{M} {{\mathop{\rm Re}\nolimits} \left[ {{e^{{\rm{i}}{\omega _n}\tau }}\mathcal{H}{{{\tilde{\mathbf{f}}}}_t}({\omega _n})\mathcal{H}{\rho _v}({\omega _n})} \right]} } \right|\\
        &\;\;\;\; \le \left|{\mathbf{\upphi}_K(v)}\right| + \frac{c_K^{r+1}T^r\left|{\mathbf{\upphi}_K(T)}\right|\left(r!\right)^2\ln{M}}{v^rM^r}
    \end{aligned}
    \label{eq:a11}
\end{equation}
for all $t \in [0,T]$ and all $\mathbf{u}(t) \in K$, where $c_K = c_Fc_{\rho} > 1$ is an independent constant. Based on Eq.~\eqref{eq:a2}, $\left(2T\right)^{-1}{{\mathop{\rm Re}\nolimits} \left[ {{e^{{\rm{i}}{\omega _n}\tau }}\mathcal{H}{{{\tilde{\mathbf{f}}}}_t}({\omega _n})\mathcal{H}{\rho _v}({\omega _n})} \right]}$ in Eq.~\eqref{eq:a11} can be expressed as
\begin{equation}
    \begin{aligned}
        &\frac{1}{2T}{{\mathop{\rm Re}\nolimits} \left[ {{e^{{\rm{i}}{\omega _n}\tau }}\mathcal{H}{{{\tilde{\mathbf{f}}}}_t}({\omega _n})\mathcal{H}{\rho _v}({\omega _n})} \right]}\\
        &\;\;\;\;\; = \frac{1}{T}\left\{ {{\mathcal{L}_t}{\bf{u}}({\omega _n}) + \frac{{{\bf{u}}(0)}}{{{\omega _n}}}\left[ {\cos ({\omega _n}t) - 1} \right]} \right\}\left\{ {{\mathop{\rm Re}\nolimits} \left[ {\mathcal{H}{\rho _v}({\omega _n})} \right]\sin ({\omega _n}\tau ) + {\mathop{\rm Im}\nolimits} \left[ {\mathcal{H}{\rho _v}({\omega _n})} \right]\cos ({\omega _n}\tau )} \right\}\\
        &\;\;\;\;\; = {\alpha _n}\left\{ {{\mathcal{L}_t}{\bf{u}}({\omega _n}) + \frac{{{\bf{u}}(0)}}{{{\omega _n}}}\left[ {\cos ({\omega _n}t) - 1} \right]} \right\}\sin ({\omega _n}\tau  - {\theta _n}),
    \end{aligned}
    \label{eq:a12}
\end{equation}
where $\alpha_n$ and $\theta_n$ are the scaled Fourier modulus and phase of $\rho_v(\tau)$, respectively, and they are independent of the input function $\mathbf{u}(t)$. $\alpha_n$ can be bounded by
\begin{equation}
    \alpha_n=\frac{1}{T}\left|\int_{-2T}^{2T}e^{-\text{i}\omega_n\tau}\rho_v(\tau)\text{d}\tau\right| \leq \frac{1}{T}\int_{-2T}^{2T}\rho_v(\tau)\text{d}\tau = \frac{1}{T}.
    \label{eq:a13}
\end{equation}
Eq.~\eqref{eq:9} can be proven by substituting Eq.~\eqref{eq:a12} into Eq.~\eqref{eq:a11}.
\hfill$\blacksquare$}

\subsection{Proof of Lemma 2}
\label{appA.2}
{\bf Proof:} For every positive integer $M$ {\color{black} larger than two} and arbitrary $M$ real frequencies $\omega_1$, $\omega_2$,..., $\omega_M$, with $\sigma_{\mathrm{ReLU}}(x) - \sigma_{\mathrm{ReLU}}(-x) = x$, there exists a set of equations with the initial conditions $\mathbf{x}(0) = \mathbf{x}'(0) = \mathbf{0}$
\begin{equation}
    \left\{
    \begin{aligned}
    &{{{\bf{x}}}_1''}(t) = {\sigma _{{\rm{ReLU}}}}\left[ { - \omega _1^2{{\bf{x}}_1}(t) - 0{{{\bf{x}}}_1'}(t) + {\omega _1}{\bf{u}}(t)} \right] - {\sigma _{{\rm{ReLU}}}}\left[ {\omega _1^2{{\bf{x}}_1}(t) + 0{{{\bf{x}}}_1'}(t) - {\omega _1}{\bf{u}}(t)} \right]\\
    &{{{\bf{x}}}_2''}(t) = {\sigma _{{\rm{ReLU}}}}\left[ { - \omega _2^2{{\bf{x}}_2}(t) - 0{{{\bf{x}}}_2'}(t) + {\omega _2}{\bf{u}}(t)} \right] - {\sigma _{{\rm{ReLU}}}}\left[ {\omega _2^2{{\bf{x}}_2}(t) + 0{{{\bf{x}}}_2'}(t) - {\omega _2}{\bf{u}}(t)} \right]\\  &\;\;\;\;\;\;\;\;\;\;\;\;\;\;\;\;\;\;\;\;\;\;\;\;\;\;\;\;\;\;\;\;\;\;\;\;\;\;\;\;\;\;\;\;\;\;\;\;\;\;\;\;\;\;\;\;\;\;\;\;\;\;\;\;\;\;\;\;\ \vdots \\
    &{{{\bf{x}}}_M''}(t) = {\sigma _{{\rm{ReLU}}}}\left[ { - \omega _M^2{{\bf{x}}_M}(t) - 0{{{\bf{x}}}_M'}(t) + {\omega _M}{\bf{u}}(t)} \right] - {\sigma _{{\rm{ReLU}}}}\left[ {\omega _M^2{{\bf{x}}_M}(t) + 0{{{\bf{x}}}_M'}(t) - {\omega _M}{\bf{u}}(t)} \right]\\
    \end{aligned},
    \right.
    \label{eq:a14}
\end{equation}
such that its solution satisfies $\mathbf{x}(t) = [\mathbf{x}_1^\top(t),\mathbf{x}_2^\top(t),..., \mathbf{x}_M^\top(t)]^\top = [\mathcal{L}_t\mathbf{u}^\top(\omega_1),\mathcal{L}_t\mathbf{u}^\top(\omega_2),...,\mathcal{L}_t\mathbf{u}^\top(\omega_M)]^\top$ for all $t \in [0, T]$ and all $\mathbf{u}(t) \in K$. Eq.~\eqref{eq:a14} can be expressed in the form of the second-order ODE in Eq.~\eqref{eq:1} with a one-hidden-layer MLP ${\it{\Gamma}}(\cdot)$ utilizing $\sigma_{\mathrm{ReLU}}(\cdot)$. The widths of the input, hidden, and output layers of ${\it{\Gamma}}(\cdot)$ are $\wGammain = p(2M+1)$, $\wGamma = 2pM$, and $\wGammaout = pM$, respectively. 
\hfill$\blacksquare$

\subsection{Proof of Lemma 3}
\label{appA.3}
{\bf Proof:} From Lemma 1 and Lemma 2, for an arbitrary positive $v < T$, and every positive integer $M$ {\color{black} larger than two}, there exists an MLP ${\it{\Gamma}}(\cdot)$ described in Lemma 2, such that for every $\mathbf{u}(t) \in K$, its corresponding solution $\mathbf{x}(t) = [\mathbf{x}_1^\top(t),\mathbf{x}_2^\top(t),..., \mathbf{x}_M^\top(t)]^\top$ from Eq.~\eqref{eq:1} described in Lemma 2 satisfies
\begin{equation}
    \sup_{\tau \in [0,t]}\left| {{\bf{u}}(\tau ) - {{{\hat{\mathbf u}}}_t}(\tau )} \right| \le {\varepsilon _K}
    \label{eq:a15}
\end{equation}
for all $t \in [0, T]$, where $\varepsilon_K$ is in Eq.~\eqref{eq:11}, ${\hat{\mathbf{u}}}_t(\tau)$ over $\tau \in [0, t]$ is expressed as
\begin{equation}
    {{\hat{\mathbf u}}_t}(\tau ) = {\bf{u}}(0) + \sum\limits_{n = 1}^M {{\alpha _n}\left\{ {{{\bf{x}}_n}(t) + \frac{{{\bf{u}}(0)}}{{{\omega _n}}}\left[ {\cos ({\omega _n}t) - 1} \right]} \right\}\sin \left[ {{\omega _n}(t - \tau ) - {\theta _n}} \right]},
    \label{eq:a16}
\end{equation}
$\mathbf{x}_n(t)$ is the $n^{\text{th}}$ vector element of $\mathbf{x}(t)$ that can be calculated as
\begin{equation}
    {{\bf{x}}_n}(t) = {\mathcal{L}_t}{\bf{u}}({\omega _n}) = \int_0^t {{\bf{u}}(t - \tau )\sin ({\omega _n}\tau ){\rm{d}}\tau },
    \label{eq:a17}
\end{equation}
$\omega_1$, $\omega_2$,..., $\omega_M$ are related to the network parameters of ${\it{\Gamma}}(\cdot)$, and $\alpha_n$ and $\theta_n$, $n = 1, 2,..., M$, are specified in Lemma 1.

Since $\Phi$ is a causal and uniformly continuous operator from $\Czero{p}$ to $\Czero{q}$, and the parameters $\omega_n$, $\alpha_n$, and $\theta_n$ in Eq.~\eqref{eq:a16} are independent of the input function $\mathbf{u}(t)$ according to the proof of Lemma 1, ${\Phi}[{\hat{\mathbf{u}}}_t(\tau)](t)$ can be regarded as a continuous function with respect to the values of $\mathbf{x}(t)$, $\mathbf{u}(0)$, and time $t$. The continuity of ${\Phi}[{\hat{\mathbf{u}}}_t(\tau)](t)$ with respect to the values of $\mathbf{x}(t)$ and $\mathbf{u}(0)$ is evident because $\Phi$ is a continuous operator. The continuity with respect to the value of $t$ is analytically established through Eq.~\eqref{eq:a36} in the proof of Lemma 4 in \ref{appA.4}. Thus, for every positive integer \textit{M} {\color{black} larger than two}, a related continuous mapping $\Psi: {\mathbb{R}}^{p(M+1)}\times{[0, T]} \to {\mathbb{R}}^q$ can be defined as $\Psi(\mathbf{x}, \mathbf{u}_0, t) = \Psi[\mathbf{x}(t), \mathbf{u}(0), t] = {\Phi}[{\hat{\mathbf{u}}}_t(\tau)](t)$, where $\mathbf{x} = [\mathbf{x}_1^\top,\mathbf{x}_2^\top,..., \mathbf{x}_M^\top]^\top$, $\mathbf{x}_n$ and $\mathbf{u}_0$ represent the values of $\mathbf{x}_n(t)$ and $\mathbf{u}(0)$ in Eq.~\eqref{eq:a16}, respectively.

The value of ${\Phi}[{\hat{\mathbf{u}}}_t(\tau)](t)$ at $t$ is independent of the values of ${\hat{\mathbf{u}}}_t(\tau)$ over $\tau > t$. Given a time instant $t_0$, $0 \leq t_0 \leq T$, it can be obtained 
\begin{equation}
    \begin{aligned}
    &|{\Phi}[\mathbf{u}(\tau)](t_0) - \Psi(\mathbf{x}, \mathbf{u}_0, t_0)| = |{\Phi}[\mathbf{u}(\tau)](t_0) - {\Phi}[{\hat{\mathbf{u}}}_{t_0}(\tau)](t_0)| \leq \norminft{t_0}{{\Phi}[\mathbf{u}(\tau)](t) - {\Phi}[{\hat{\mathbf{u}}}_{t_0}(\tau)](t)}\\
    &\;\;\;\;\;\;\;\;\;\;\;\;\;\;\;\;\;\;\;\;\;\;\;\;\;\;\;\;\;\;\;\;\;\;\;\;\; \mathop \le \limits^{(a)}{\phi_{\Phi}}\left[\norminft{t_0}{\mathbf{u}(t) - \hat{\mathbf{u}}_{t_0}(t)}\right] \mathop  \le \limits^{(b)} {\phi_{\Phi}}(\varepsilon_K),
    \end{aligned}
    \label{eq:a18}
\end{equation}
where ${\phi_{\Phi}}(\cdot) = \sum_{j=1}^{q}{\phi}_{\Phi_j}(\cdot)$ is a monotonic modulus of continuity of $\Phi$ in Assumption 6, $\phi_{\Phi_j}(\cdot)$ is a monotonic modulus of continuity of the $j^{\mathrm{th}}$ element $\Phi_j$ of $\Phi$, $\varepsilon_K$ is in Eq.~\eqref{eq:11} of Lemma 1, Step (\textit{b}) follows from Eq.~\eqref{eq:a15}, and  Step (\textit{a}) holds from the causality of $\Phi$ and it will be proven subsequently. Since $\varepsilon_K$ is independent of \textit{t}, $|{\Phi}[\mathbf{u}(\tau)](t_0) - \Psi(\mathbf{x}, \mathbf{u}_0, t_0)| \leq {\phi_{\Phi}}(\varepsilon_K)$ is valid for all $t_0 \in [0,T]$, which proves Eq.~\eqref{eq:16}.

$\norminft{t_0}{{\Phi}[\mathbf{u}_1(\tau)](t) - {\Phi}[\mathbf{u}_2(\tau)](t)} \leq {\phi_{\Phi}}\left[\norminft{t_0}{\mathbf{u}_1(t) - \mathbf{u}_2(t)}\right]$ is satisfied for arbitrary $\mathbf{u}_1(t)$ and $\mathbf{u}_2(t)$ $\in \Czero{p}$. This can be proven by setting
\begin{equation}
    {{\bf{u}}_i}(t) = 
    \left\{
    \begin{aligned}
        &{{\bf{u}}_i}(t),\;\;\;0 \le t \le {t_0}\\
        &{{\bf{u}}_i}({t_0}),\;\;{t_0} \le t \le T
    \end{aligned}, 
    \right. \;\;i = 1\;{\rm{and}}\;2,
    \label{eq:a19}
\end{equation}
then, it is satisfied $\norminft{t_0}{{\Phi}[\mathbf{u}_1(\tau)](t) - {\Phi}[\mathbf{u}_2(\tau)](t)} \leq \norminf{{\Phi}[\mathbf{u}_1(\tau)](t) - {\Phi}[\mathbf{u}_2(\tau)](t)} \leq {\phi_{\Phi}}\left[ \norminf{\mathbf{u}_1(t) - \mathbf{u}_2(t)} \right] = {\phi_{\Phi}}\left[\norminft{t_0}{\mathbf{u}_1(t) - \mathbf{u}_2(t)}\right]$. Since the values of $\lVert {\Phi}[\mathbf{u}_1(\tau)](t) - $ ${\Phi}[\mathbf{u}_2(\tau)](t)\rVert_{L^{\infty}[0,t_0]}$ and ${\phi_{\Phi}}\left[\norminft{t_0}{\mathbf{u}_1(t) - \mathbf{u}_2(t)}\right]$ are independent of the values of $\mathbf{u}_1(t)$ and $\mathbf{u}_2(t)$ over $t_0 \leq t \leq T$, $\norminft{t_0}{{\Phi}[\mathbf{u}_1(\tau)](t) - {\Phi}[\mathbf{u}_2(\tau)](t)} \leq {\phi_{\Phi}}\left[\norminft{t_0}{\mathbf{u}_1(t) - \mathbf{u}_2(t)}\right]$ is valid for arbitrary $\mathbf{u}_1(t)$ and $\mathbf{u}_2(t)$ $\in \Czero{p}$. Since $\mathbf{u}(t)$ and $\hat{\mathbf{u}}_{t_0}(t)$ in Eq.~\eqref{eq:a18} belong to $\Czero{p}$, Step (\textit{a}) in Eq.~\eqref{eq:a18} is valid.
\hfill$\blacksquare$

\subsection{Proof of Lemma 4}
\label{appA.4}
{\bf Proof:} Since \textit{K} is compact, by Assumption 2, it is closed and there exists a constant $B_K$ such that $\norminf{u_i(t)} \leq B_K$ for all $\mathbf{u}(t) \in K$ and $1 \leq i \leq p$, where $u_i(t)$ is the $i^{\text{th}}$ element of $\mathbf{u}(t)$. For every $M$  satisfying Eq.~\eqref{eq:15}, $\Psi \left( {\bf{x}},{\bf{u}}_0,t \right) = \Psi \left[ {\bf{x}}(t),{\bf{u}}(0),t \right] = \Phi[{\hat{\mathbf{u}}}_t(\tau)](t)$ is a continuous function over a compact domain $\Omega_{M}\left(TB_K,B_K,T\right) = \left[-TB_K,TB_K\right]^{pM} \times \left[-B_K,B_K\right]^p \times \left[0,T\right]$, where $TB_K$ is the bound for the elements of $\mathbf{x}_n$, $n = 1, 2,...,M$, because $\mathbf{x}_n = \mathbf{x}_n(t)$ is calculated from Eq.~\eqref{eq:a17}. In $\Omega_{M}\left(TB_K,B_K,T\right)$, $\left[-TB_K,TB_K\right]^{pM}$ is the value range of $\mathbf{x}$, $\left[-B_K,B_K\right]^p$ is the value range of $\mathbf{u}_0$, and $[0, T]$ is the value range of \textit{t}.

The $i^{\text{th}}$ element ${\hat{u}_{t,i}}(\tau)$ of ${\hat{\mathbf{u}}_t}(\tau) = [{\hat{u}_{t,1}}(\tau),{\hat{u}_{t,2}}(\tau),..., {\hat{u}_{t,p}}(\tau)]^{\top}$ in Eq.~\eqref{eq:a16} can be expressed as
\begin{equation}
    {\hat u_{t,i}}(\tau ) = {u_i}(0) + \sum\limits_{n = 1}^{{M}} {{\alpha _n}\left\{ {{x_{n,i}}(t) + \frac{{{u_i}(0)}}{{{\omega _n}}}\left[ {\cos ({\omega _n}t) - 1} \right]} \right\}\sin \left[ {{\omega _n}(t - \tau ) - {\theta _n}} \right]},
    \label{eq:a20}
\end{equation}
where $x_{n,i}(t)$ and $u_i(0)$ are respectively the $i^{\text{th}}$ elements of $\mathbf{x}_n(t)$ and $\mathbf{u}(0)$. Consider two {\color{black} approximations} ${\hat{\mathbf{u}}_{t,1}(\tau)}$ and ${\hat{\mathbf{u}}_{t,2}(\tau)}$ of  ${\mathbf{u}}_1(t)$ and ${\mathbf{u}}_2(t)$ $\in K$, respectively, with $\mathbf{x}_{1,n}(t) = \mathcal{L}_t\mathbf{u}_1(\omega_n)$ and $\mathbf{x}_{2,n}(t) = \mathcal{L}_t\mathbf{u}_2(\omega_n)$, from Eq.~\eqref{eq:a20}, the difference between the $i^{\text{th}}$ elements ${\hat{u}_{t,1,i}}(\tau)$ of ${\hat{\mathbf{u}}_{t,1}(\tau)}$ and ${\hat{u}_{t,2,i}}(\tau)$ of ${\hat{\mathbf{u}}_{t,2}(\tau)}$ can be bounded by
\begin{equation}
    \begin{aligned}
    &\mathop {\sup }\limits_{\tau  \in [0,t]} \left| {{{\hat u}_{t,1,i}}(\tau ) - {{\hat u}_{t,2,i}}(\tau )} \right|\\
    &\;\;\;\; \le \left| {{u_{1,i}}(0) - {u_{2,i}}(0)} \right| + \mathop {\sup }\limits_{\tau  \in [0,t]} \left( {\sum\limits_{n = 1}^M {\left| {{\alpha _n}\left\{ {\left[ {{x_{1,n,i}}(t) - {x_{2,n,i}}(t)} \right] + \frac{{{u_{1,i}}(0) - {u_{2,i}}(0)}}{{{\omega _n}}}\left[ {\cos ({\omega _n}t) - 1} \right]} \right\}\sin \left[ {{\omega _n}(t - \tau ) - {\theta _n}} \right]} \right|} } \right)\\
    &\;\;\;\; \le \left| {{u_{1,i}}(0) - {u_{2,i}}(0)} \right| + \sum\limits_{n = 1}^M {\left| {{\alpha _n}} \right|\left| {\left[ {{x_{1,n,i}}(t) - {x_{2,n,i}}(t)} \right] + \frac{{{u_{1,i}}(0) - {u_{2,i}}(0)}}{{{\omega _n}}}\left[ {\cos ({\omega _n}t) - 1} \right]} \right|} \\
    &\;\;\;\; \mathop \le \limits^{(a)} \left| {{u_{1,i}}(0) - {u_{2,i}}(0)} \right| + \frac{1}{T}\sum\limits_{n = 1}^M {\left[ {\left| {{x_{1,n,i}}(t) - {x_{2,n,i}}(t)} \right| + 2\left| {\frac{{{u_{1,i}}(0) - {u_{2,i}}(0)}}{{{\omega _n}}}} \right|} \right]} \\
    &\;\;\;\; \le \left( {1 + \frac{{2M}}{{T{\omega _1}}}} \right)\left| {{u_{1,i}}(0) - {u_{2,i}}(0)} \right| + \frac{1}{T}\sum\limits_{n = 1}^M {\left| {{x_{1,n,i}}(t) - {x_{2,n,i}}(t)} \right|} \\
    &\;\;\;\; \mathop \le \limits^{(b)} \max \left[ {\left( {1 + 2M} \right),{T^{ - 1}}} \right]\left[ {\left| {{u_{1,i}}(0) - {u_{2,i}}(0)} \right| + \sum\limits_{n = 1}^M {\left| {{x_{1,n,i}}(t) - {x_{2,n,i}}(t)} \right|} } \right],
    \end{aligned}
    \label{eq:a21}
\end{equation}
where $u_{1,i}(0)$ and $u_{2,i}(0)$ are respectively the $i^{\text{th}}$ elements of $\mathbf{u}_1(0)$ and $\mathbf{u}_2(0)$, $x_{1,n,i}(t)$ and $x_{2,n,i}(t)$ are respectively the $i^{\text{th}}$ elements of $\mathbf{x}_{1,n}(t)$ and $\mathbf{x}_{2,n}(t)$, Step ($a$) holds because $\alpha_n$ is bounded by $T^{-1}$ in Eq.~\eqref{eq:a13} and Step ($b$) follows because $\omega_n = \pi n/(2T)$.

Subsequently, the first-order derivatives of the $j^{\text{th}}$ element $\psi_j\left(\mathbf{x},\mathbf{u}_0,t\right)$ of $\Psi\left(\mathbf{x},\mathbf{u}_0,t\right)$, $j = 1,2,...,q$, are analyzed. With the modulus of continuity $\phi_{\Phi_j}(\cdot)$ of $\Phi_j$, which is the $j^{\text{th}}$ element the casual operator $\Phi$, and Eq.~\eqref{eq:a21}, the difference between $\psi_j\left(\mathbf{x},\mathbf{u}_0,t\right)$ and $\psi_j\left(\mathbf{x}+\Delta\mathbf{x},\mathbf{u}_0+\Delta\mathbf{u}_0,t\right)$ can be bounded by
\begin{equation}
    \begin{aligned}
    &\left| {{\psi _j}({\bf{x}} + \Delta {\bf{x}},{{\bf{u}}_0} + \Delta {{\bf{u}}_0},t) - {\psi _j}({\bf{x}},{{\bf{u}}_0},t)} \right|\\
    &\;\;\;\;\; \mathop  \le \limits^{(a)}{\phi _{{\Phi _j}}}\left\{ \norminft{t}{
    {\hat{\mathbf{u}}_{t,1}}(\tau) - {\hat{\mathbf{u}}_{t,2}}(\tau)} \right\}
    = {\phi _{{\Phi _j}}}\left({\mathop {\sup }\limits_{\tau  \in [0,t]} \left[ {\sum\limits_{i = 1}^p {\left| {{{\hat u}_{t,1,i}}(\tau ) - {{\hat u}_{t,2,i}}(\tau )} \right|} } \right]}\right)\\
    &\;\;\;\;\; \le {\phi _{{\Phi _j}}}\left[ {\sum\limits_{i = 1}^p {\mathop {\sup }\limits_{\tau  \in [0,t]} \left| {{{\hat u}_{t,1,i}}(\tau ) - {{\hat u}_{t,2,i}}(\tau )} \right|} } \right] \le {\phi _{{\Phi _j}}}\left\{ {\max \left[ {\left( {1 + 2M} \right),{T^{ - 1}}} \right]\left( {\left| {\Delta {\bf{x}}} \right| + \left| {\Delta {{\bf{u}}_0}} \right|} \right)} \right\},
    \end{aligned}
    \label{eq:a22}
\end{equation}
where Step (a) holds from $\psi_j \left( {\bf{x}},{\bf{u}}_0,t \right) = \psi_j \left[ {\bf{x}}(t),{\bf{u}}(0),t \right] = \Phi_j[{\hat{\mathbf{u}}}_t(\tau)](t)$. Thus, the first-order derivative of $\psi_j\left(\mathbf{x},\mathbf{u}_0,t\right)$ with respect to each element of $\mathbf{x}$ or $\mathbf{u}_0$ is bounded by ${\max \left[ {\left( {1 + 2M} \right),{T^{ - 1}}} \right]}{{\phi}'_{{\Phi _j}}}\left(0\right)$, where ${\phi}'_{\Phi_j}(0) $ is the first-order derivative of  $\phi_{\Phi_j}(\cdot)$ at zero.

Next, the first-order derivative of $\psi_j\left(\mathbf{x},\mathbf{u}_0,t\right)$ with respect to $t$ is derived. The difference between $\psi_j\left(\mathbf{x},\mathbf{u}_0,t\right)$ and $\psi_j\left(\mathbf{x},\mathbf{u}_0,t + \Delta t\right)$ can be calculated as
\begin{equation}
    {\psi _j}({\bf{x}},{{\bf{u}}_0},t + \Delta t) - {\psi _j}({\bf{x}},{{\bf{u}}_0},t) = {\Phi _j}\left[ {{{{\hat{\mathbf u}}}_{t + \Delta t}}(\tau )} \right](t + \Delta t) - {\Phi _j}\left[ {{{{\hat{\mathbf u}}}_t}(\tau )} \right](t),
    \label{eq:a23}
\end{equation}
where
\begin{equation}
    {{\hat{\mathbf u}}_t}(\tau ) = {{\bf{u}}_0} + \sum\limits_{n = 1}^M {{\alpha _n}\left\{ {{{\bf{x}}_n} + \frac{{{{\bf{u}}_0}}}{{{\omega _n}}}\left[ {\cos ({\omega _n}t) - 1} \right]} \right\}\sin \left[ {{\omega _n}(t - \tau ) - {\theta _n}} \right]}
    \label{eq:a24}
\end{equation}
and
\begin{equation}
    {{\hat{\mathbf u}}_{t + \Delta t}}(\tau ) = {{\bf{u}}_0} + \sum\limits_{n = 1}^M {{\alpha _n}\left( {{{\bf{x}}_n} + \frac{{{{\bf{u}}_0}}}{{{\omega _n}}}\left\{ {\cos \left[ {{\omega _n}(t + \Delta t)} \right] - 1} \right\}} \right)\sin \left[ {{\omega _n}(t + \Delta t - \tau ) - {\theta _n}} \right]}.
    \label{eq:a25}
\end{equation}
$\left|\psi_j\left(\mathbf{x},\mathbf{u}_0,t + \Delta t\right) - \psi_j\left(\mathbf{x},\mathbf{u}_0,t\right)\right|$ can be bounded by
\begin{equation}
    \begin{aligned}
    &\left| {{\psi _j}({\bf{x}},{{\bf{u}}_0},t + \Delta t) - {\psi _j}({\bf{x}},{{\bf{u}}_0},t)} \right|\\
    &\;\;\; \le \left| {{\Phi _j}\left[ {{{{\hat{\bf u}}}_{t + \Delta t}}(\tau )} \right](t + \Delta t) - {\Phi _j}\left[ {{{{\hat{\mathbf u}}}_{t + \Delta t}}(\tau )} \right](t)} \right| + \left| {{\Phi _j}\left[ {{{{\hat{\mathbf u}}}_{t + \Delta t}}(\tau )} \right](t) - {\Phi _j}\left[ {{{{\hat{\mathbf u}}}_t}(\tau )} \right](t)} \right|\\
    &\;\;\; \le \left| {\Delta t} \right|{D_{\Phi_j}} + \left| {{\Phi _j}\left[ {{{{\hat{\mathbf u}}}_{t + \Delta t}}(\tau )} \right](t) - {\Phi _j}\left[ {{{{\hat{\mathbf u}}}_t}(\tau )} \right](t)} \right|\\
    &\;\;\;\le \left| {\Delta t} \right|{D_{\Phi_j}} + {\phi _{{\Phi _j}}}\left[ {{{\left\| {{{{\hat{\mathbf u}}}_{t + \Delta t}}(\tau ) - {{{\hat{\mathbf u}}}_t}(\tau )} \right\|}_{{L^\infty }\left[0,t\right]}}} \right],
    \end{aligned}
    \label{eq:a26}
\end{equation}
where ${D_{\Phi_j}}$ is the bound of the first-order derivatives of all functions in $\Phi_j\left\{ \Czero{p} \right\}$ in Assumption 5 and it is independent of $M$.

{\color{black} From} Eqs.~\eqref{eq:a24} and \eqref{eq:a25}, the difference between ${\hat{\mathbf u}}_t(\tau)$ and ${{\hat{\mathbf u}}_{t + \Delta t}}(\tau)$, which is in the last term of Eq.~\eqref{eq:a26}, can be calculated as
\begin{equation}
    \begin{aligned}
    &{{{\hat{\mathbf u}}}_{t + \Delta t}}(\tau ) - {{{\hat{\mathbf u}}}_t}(\tau ) = \sum\limits_{n = 1}^M {{\alpha _n}\left( {{{\bf{x}}_n} + \frac{{{{\bf{u}}_0}}}{{{\omega _n}}}\left\{ {\cos \left[ {{\omega _n}(t + \Delta t)} \right] - 1} \right\}} \right)\sin \left[ {{\omega _n}(t + \Delta t - \tau ) - {\theta _n}} \right]} \\
    &\;\;\;\;\;\;\;\;\;\;\;\;\;\;\;\;\;\;\;\;\;\;\;\;\;\;\; - \sum\limits_{n = 1}^M {{\alpha _n}\left\{ {{{\bf{x}}_n} + \frac{{{{\bf{u}}_0}}}{{{\omega _n}}}\left[ {\cos ({\omega _n}t) - 1} \right]} \right\}\sin \left[ {{\omega _n}(t - \tau ) - {\theta _n}} \right]} \\
    &\;\;\;\;\;\;\;\;\;\;\;\;\;\;\;\;\;\;\;\;\;\;\; = \sum\limits_{n = 1}^M {{\alpha _n}\left( {{{\bf{x}}_n} - \frac{{{{\bf{u}}_0}}}{{{\omega _n}}} + \frac{{{{\bf{u}}_0}}}{{{\omega _n}}}\cos \left[ {{\omega _n}(t + \Delta t)} \right]} \right)\sin \left[ {{\omega _n}(t + \Delta t - \tau ) - {\theta _n}} \right]} \\
    &\;\;\;\;\;\;\;\;\;\;\;\;\;\;\;\;\;\;\;\;\;\;\;\;\;\;\; - \sum\limits_{n = 1}^M {{\alpha _n}\left[ {{{\bf{x}}_n} - \frac{{{{\bf{u}}_0}}}{{{\omega _n}}} + \frac{{{{\bf{u}}_0}}}{{{\omega _n}}}\cos ({\omega _n}t)} \right]\sin \left[ {{\omega _n}(t - \tau ) - {\theta _n}} \right]} \\
    &\;\;\;\;\;\;\;\;\;\;\;\;\;\;\;\;\;\;\;\;\;\;\; = \sum\limits_{n = 1}^M {{\alpha _n}\left( {{{\bf{x}}_n} - \frac{{{{\bf{u}}_0}}}{{{\omega _n}}}} \right){A_n}(\Delta t)}  + \sum\limits_{n = 1}^M {{\alpha _n}\frac{{{{\bf{u}}_0}}}{{{\omega _n}}}{B_n}(\Delta t)},
    \end{aligned}
    \label{eq:a27}
\end{equation}
where
\begin{equation}
    {A_n}(\Delta t) = \sin \left[ {{\omega _n}(t + \Delta t - \tau ) - {\theta _n}} \right] - \sin \left[ {{\omega _n}(t - \tau ) - {\theta _n}} \right]
    \label{eq:a28}
\end{equation}
and
\begin{equation}
    {B_n}(\Delta t) = \cos \left[ {{\omega _n}(t + \Delta t)} \right]\sin \left[ {{\omega _n}(t + \Delta t - \tau ) - {\theta _n}} \right] - \cos ({\omega _n}t)\sin \left[ {{\omega _n}(t - \tau ) - {\theta _n}} \right].
    \label{eq:a29}
\end{equation}
Then, $\left| {\hat{\mathbf u}}_{t+\Delta t}(\tau) - {{\hat{\mathbf u}}_{t}}(\tau) \right|$ is bounded by
\begin{equation}
    \left| {{{{\hat{\mathbf u}}}_{t + \Delta t}}(\tau ) - {{{\hat{\mathbf u}}}_t}(\tau )} \right| \le \sum\limits_{n = 1}^M {\left| {{\alpha _n}} \right|\left| {\left( {{{\bf{x}}_n} - \frac{{{{\bf{u}}_0}}}{{{\omega _n}}}} \right)} \right|\left| {{A_n}(\Delta t)} \right|}  + \sum\limits_{n = 1}^M {\left| {{\alpha _n}} \right|\left| {\frac{{{{\bf{u}}_0}}}{{{\omega _n}}}} \right|\left| {{B_n}(\Delta t)} \right|}.
    \label{eq:a30}
\end{equation}

Based on $|\sin(x) - \sin(y)| \leq |x - y|$, $A_n(\Delta t)$ in Eq.~\eqref{eq:a28} is bounded by $\left| A_n(\Delta t) \right| \leq \left| \omega_n\Delta t\right|$ and the first sum on the right side of Eq.~\eqref{eq:a30} can be bounded by
{\color{black} \begin{equation}
    \begin{aligned}
    &\sum\limits_{n = 1}^M {\left| {{\alpha _n}} \right|\left| {\left( {{{\bf{x}}_n} - \frac{{{{\bf{u}}_0}}}{{{\omega _n}}}} \right)} \right|\left| {{A_n}(\Delta t)} \right|}\\
    &\;\;\; \le \sum\limits_{n = 1}^M {\left| {{\alpha _n}} \right|\left| {\left( {{{\bf{x}}_n} - \frac{{{{\bf{u}}_0}}}{{{\omega _n}}}} \right)} \right|\left| {{\omega _n}\Delta t} \right|} = \left| {\Delta t} \right|\sum\limits_{n = 1}^M {\left| {{\alpha _n}{\omega _n}{{\bf{x}}_n} - {\alpha _n}{{\bf{u}}_0}} \right|} \le \left| {\Delta t} \right|\sum\limits_{n = 1}^M {\left| {{\alpha _n}{\omega _n}{{\bf{x}}_n}} \right|}  + \left| {\Delta t} \right|\sum\limits_{n = 1}^M {\left| {{\alpha _n}{{\bf{u}}_0}} \right|} \\
    &\;\;\; \le \left| {\Delta t} \right|\sum\limits_{n = 1}^M {\left| {{\alpha _n}} \right|\left| {{\omega _n}} \right|\left| {{{\bf{x}}_n}} \right|}  + \left| {\Delta t} \right|p{B_K}M{T^{ - 1}} \le \left| {\Delta t} \right|pT{B_K}\sum\limits_{n = 1}^{ + \infty } {\left| {{\alpha _n}} \right|\left| {{\omega _n}} \right|}  + \left| {\Delta t} \right|p{B_K}M{T^{ - 1}}\\
    &\;\;\;\mathop  \le \limits^{(a)} \left| {\Delta t} \right|pT{B_K}\frac{{2c_{\rho ,3}M^3}}{3T^2} + \left| {\Delta t} \right|p{B_K}M{T^{ - 1}} \le \left| {\Delta t} \right|p{B_K}M^{3}T^{-1}\left( {c_{\rho ,3} + 1} \right),
    \end{aligned}
    \label{eq:a31}
\end{equation}
where Step (\textit{a}) is from
% \begin{equation}
%     \begin{array}{l}
%     \sum\limits_{n = 1}^{ +\infty } {\left| {{\omega _n}} \right|\left| {{\alpha _n}} \right|} = \sum\limits_{n = 1}^{ +\infty } \frac{1}{\left| {\omega _n} \right|}\left| {\omega_n^2}{\alpha _n} \right| \mathop \le \limits^{(b)}\frac{1}{T}\sqrt{\sum\limits_{n = 1}^{ +\infty } \frac{1}{\left| {\omega _n} \right|^2}} \sqrt{\sum\limits_{n = 1}^{ +\infty }\left|{\omega_n^2} {\int_{- 2T}^{2T} {{e^{ - {\rm{i}}{\omega _n}\tau }}{\rho _v}(\tau ){\rm{d}}\tau } } \right|^2} \\
    
%     \;\;\;\;\;\;\;\;\;\;\;\;\;\;\;\;\;\le \sqrt{\frac{2}{3}}\sqrt{\int_{ - v}^0 {\left| {{{\rho}''_v}(\tau)} \right|^2{\rm{d}}\tau } } \le \sqrt{v}\mathop {\max }\limits_{t \in [ - v,0]} \left| {{{\rho}''_v}(t)} \right| \le \frac{c_{\rho,2}}{v^{2.5}}
%     \end{array}
%     \label{eq:a32}
% \end{equation}

\begin{equation}
    \begin{aligned}
    &\sum\limits_{n = 1}^{ +\infty } {\left| {{\omega _n}} \right|\left| {{\alpha _n}} \right|} \mathop = \limits^{(b)} \frac{1}{T}\sum\limits_{n = 1}^{ +\infty }\left|{\omega_n} {\int_{- 2T}^{2T} {{e^{ - {\rm{i}}{\omega _n}\tau }}{\rho _v}(\tau ){\rm{d}}\tau } } \right| \le \frac{1}{T}\sum\limits_{n = 1}^{ + \infty } {\frac{1}{{\omega _n^2}}\int_{ - v}^0 {\left| {{{\rho}'''_v}(\tau)} \right|{\rm{d}}\tau } }\\
    &\;\;\;\;\;\;\;\;\;\;\;\;\;\;\;\;\; \le \frac{1}{T}\sum\limits_{n = 1}^{ + \infty } {\frac{4{T^2}v}{{\pi ^2}{n^2}}}\mathop {\max }\limits_{t \in [ - v,0]} \left| {{{\rho}'''_v}(t)} \right| \le \frac{{4T{c_{\rho ,3}}}}{{{\pi ^2}{v^3}}}\sum\limits_{n = 1}^{ + \infty } {\frac{1}{{{n^2}}}} \mathop  = \limits^{(c)} \frac{{2c_{\rho ,3}M^3}}{3T^2},
    \end{aligned}
    \label{eq:a32}
\end{equation}
Step (\textit{b}) is from Eq.~\eqref{eq:a13}, Step (\textit{c}) holds from Eq.~\eqref{eq:15} and $\sum_{n = 1}^{ + \infty }{n^{-2}} = \pi^2/6$, and $c_{\rho ,3} = \max_{\tau \in [-1,0]}\left|{\rho}_1'''(\tau)\right|$ is an independent constant with ${\rho}_1(\tau)$ in Eq.~\eqref{eq:a3}}.

$\left|B_N(\Delta t) \right|$ in Eq.~\eqref{eq:a29} can be bounded by
\begin{equation}
    \begin{aligned}
    &\left| {{B_n}(\Delta t)} \right|\le \left| {\cos \left[ {{\omega _n}(t + \Delta t)} \right]\sin \left[ {{\omega _n}(t + \Delta t - \tau ) - {\theta _n}} \right] - \cos \left[ {{\omega _n}(t + \Delta t)} \right]\sin \left[ {{\omega _n}(t - \tau ) - {\theta _n}} \right]} \right|\\
    &\;\;\;\;\;\;\;\;\;\;\;\;\;\;\;+ \left| {\cos \left[ {{\omega _n}(t + \Delta t)} \right]\sin \left[ {{\omega _n}(t - \tau ) - {\theta _n}} \right] - \cos ({\omega _n}t)\sin \left[ {{\omega _n}(t - \tau ) - {\theta _n}} \right]} \right| \le 2\left| {{\omega _n}} \right|\left| {\Delta t} \right|.
    \end{aligned}
    \label{eq:a33}
\end{equation}
Thus, the second sum on the right side of Eq.~\eqref{eq:a30} can be simplified as
\begin{equation}
    \sum\limits_{n = 1}^M {\left| {{\alpha _n}} \right|\left| {\frac{{{{\bf{u}}_0}}}{{{\omega _n}}}} \right|\left| {{B_n}(\Delta t)} \right|}  \le 2\sum\limits_{n = 1}^M {\left| {{\alpha _n}} \right|\left| {\frac{{{{\bf{u}}_0}}}{{{\omega _n}}}} \right|\left| {{\omega _n}} \right|\left| {\Delta t} \right|}  \le \frac{{2p{B_K}M}}{T}\left| {\Delta t} \right|.
    \label{eq:a34}
\end{equation}
By substituting Eqs.~\eqref{eq:a31} and \eqref{eq:a34} into Eq.~\eqref{eq:a30}, $\left| {\hat{\mathbf u}}_{t + \Delta t}(\tau) - {{\hat{\mathbf u}}_{t}}(\tau) \right|$ is bounded by
{\color{black} \begin{equation}
    \left| {{{{\bf{\hat u}}}_{t + \Delta t}}(\tau ) - {{{\bf{\hat u}}}_t}(\tau )} \right| \le \left| {\Delta t} \right|p{B_K}M^{3}T^{-1}\left( {c_{\rho ,3} + 3}\right).
    \label{eq:a35}
\end{equation}}
By substituting Eq.~\eqref{eq:a35} into Eq.~\eqref{eq:a26}, $\left|\psi_j\left(\mathbf{x},\mathbf{u}_0,t + \Delta t\right) - \psi_j\left(\mathbf{x},\mathbf{u}_0,t\right)\right|$ can be bounded by 
{\color{black}\begin{equation}
    \left| {{\psi _j}({\bf{x}},{{\bf{u}}_0},t + \Delta t) - {\psi _j}({\bf{x}},{{\bf{u}}_0},t)} \right| \le \left| {\Delta t} \right|{D_{{\Phi _j}}} + {\phi _{{\Phi _j}}}\left({\left| {\Delta t} \right|p{B_K}M^{3}T^{-1}c_{L}} \right),
    \label{eq:a36}
\end{equation}}
where {\color{black} $c_{L} = c_{\rho ,3} + 3$} is an independent constant. The first-order derivative of $\psi_j\left(\mathbf{x},\mathbf{u}_0,t\right)$ with respect to \textit{t} is bounded by
{\color{black} \begin{equation}
    \left| {\frac{{{\partial}{\psi _j}({\mathbf{x}},{{\bf{u}}_0},t)}}{{{\partial}t}}} \right| \le {D_{{\Phi _j}}} + \mathop {\lim }\limits_{\left| {\Delta t} \right| \to 0} {\left| {\Delta t} \right|^{ - 1}}{\phi _{{\Phi _j}}}\left( {\left| {\Delta t} \right|p{B_K}{M^{3}}T^{-1}c_{L}} \right) = {D_{{\Phi _j}}} + p{B_K}M^{3}T^{-1}c_{L}{{\phi '}_{{\Phi _j}}}(0).
    \label{eq:a37}
\end{equation}}

Eqs.~\eqref{eq:a22} and \eqref{eq:a37} indicate that for every $M$ satisfying Eq.~\eqref{eq:15} along with $v$ satisfying Eq.~\eqref{eq:14}, over the compact domain $\Omega_{M}\left(TB_K,B_K,T\right) = \left[-TB_K,TB_K\right]^{pM} \times \left[-B_K,B_K\right]^p \times \left[0,T\right]$, the Lipschitz constant of $\psi_j\left(\mathbf{x},\mathbf{u}_0,t\right)$ with respect to its each argument is $L_{\psi_j}\left(M\right)$ in Eq.~\eqref{eq:17}. The linear modulus of continuity $\phi_{\psi_j,M}(x)$ in Eq.~\eqref{eq:18} of $\psi_j\left(\mathbf{x},\mathbf{u}_0,t\right)$ under $L^2$-norm is proven by
\begin{equation}
    \begin{aligned}
    &\left| {{\psi _j}({\bf{x}} + \Delta {\bf{x}},{{\bf{u}}_0} + \Delta {{\bf{u}}_0},t + \Delta t) - {\psi _j}({\bf{x}},{{\bf{u}}_0},t)} \right|\\
    &\;\;\;\;\;\; \le \left| {{\psi _j}({\bf{x}} + \Delta {\bf{x}},{{\bf{u}}_0} + \Delta {{\bf{u}}_0},t + \Delta t) - {\psi _j}({\bf{x}},{{\bf{u}}_0} + \Delta {{\bf{u}}_0},t + \Delta t)} \right|\\
    &\;\;\;\;\;\;\;\;\;\; + \left| {{\psi _j}({\bf{x}},{{\bf{u}}_0} + \Delta {{\bf{u}}_0},t + \Delta t) - {\psi _j}({\bf{x}},{{\bf{u}}_0},t + \Delta t)} \right| + \left| {{\psi _j}({\bf{x}},{{\bf{u}}_0},t + \Delta t) - {\psi _j}({\bf{x}},{{\bf{u}}_0},t)} \right|\\
    &\;\;\;\;\;\; \le {L_{{\psi _j}}\left(M\right)}\left( {\left| {\Delta {\bf{x}}} \right| + \left| {\Delta {{\bf{u}}_0}} \right| + \left| {\Delta t} \right|} \right) \mathop  \le \limits^{(a)} \sqrt {p(M + 1) + 1} {L_{{\psi _j}}\left(M\right)}\sqrt {\left| {\Delta {\bf{x}}} \right|_2^2 + \left| {\Delta {{\bf{u}}_0}} \right|_2^2 + \left| {\Delta t} \right|_2^2},
    \end{aligned}
    \label{eq:a38}
\end{equation}
where Step (\textit{a}) holds from the Cauchy-Schwarz inequality.
\hfill$\blacksquare$

\subsection{Proof of Theorem 1}
\label{appA.5}
{\bf Proof:} From Remark 2 and Lemma 2, for an arbitrary integer $M_{\it{\Gamma}}$ satisfying Eq.~\eqref{eq:15} along with \textit{v} satisfying Eq.~\eqref{eq:14}, there exists a one-hidden-layer MLP ${\it{\Gamma}}(\cdot)$ specified in Lemma 2, for every $\mathbf{u}(t) \in K$, its corresponding solution $\mathbf{x}(t) = [\mathbf{x}_{1}^{\top}(t), \mathbf{x}_{2}^{\top}(t),..., \mathbf{x}_{M_{\it{\Gamma}}}^{\top}(t)]^{\top}$ calculated from the second-order ODE in Eq.~\eqref{eq:1} with the initial conditions $\mathbf{x}(0) = {\mathbf{x}}'(0) = \mathbf{0}$ {\color{black} equals} the sine transform coefficients of $\mathbf{u}(t)$, as shown in Eq.~\eqref{eq:a17}. From Lemma 3 and Remark 3, there exists a continuous mapping $\Psi = [\psi_1, \psi_2,..., \psi_q]^{\top}: \mathbb{R}^{p\left(M_{\it{\Gamma}}+1\right)} \times \left[0,T\right] \to \mathbb{R}^q$ for all $\mathbf{u}(t) \in K$, the difference between $\Phi[\mathbf{u}(\tau)](t)$ and $\Psi[\mathbf{x}(t),\mathbf{u}(0),t]$ can be bounded by
{\color{black} \begin{equation}
    \left| {\Phi \left[ {{\bf{u}}(\tau )} \right](t) - \Psi \left[ {{\bf{x}}(t),{\bf{u}}(0),t} \right]} \right| \le {\phi _\Phi }\left[\frac{32c_KTpL_K\left(\ln{M}\right)^{2}}{M}\right].
    \label{eq:a39}
\end{equation}}

From Lemma 4, the continuous function $\Psi \left({\bf{x}},{\bf{u}}_0,t \right)$ in Eq.~\eqref{eq:a39} is compactly supported within $\Omega_{M_{\it{\Gamma}}}\left(TB_K,B_K,T\right) = \left[-TB_K,TB_K\right]^{pM_{\it{\Gamma}}} \times \left[-B_K,B_K\right]^p \times \left[0,T\right]$ for every integer $M_{\it{\Gamma}}$ satisfying Eq.~\eqref{eq:15} and all $\mathbf{u}(t) \in K$, where $\mathbf{x}$ and $\mathbf{u}_0$ represent the values of $\mathbf{x}(t)$ and $\mathbf{u}(0)$, respectively. The {\color{black} Lipschitz} constant $L_{{\psi _j}}\left(M_{\it{\Gamma}}\right)$ of $\psi_j\left(\mathbf{x},\mathbf{u}_0,t\right)$ of $\Psi \left({\bf{x}},{\bf{u}}_0,t \right)$ with respect to its each argument is in Eq.~\eqref{eq:17} and a linear modulus of continuity $\phi_{\psi_j,M_{\itGamma}}(x)$ of $\psi_j\left(\mathbf{x},\mathbf{u}_0,t\right)$ under {\color{black} the} $L^2$-norm is in Eq.~\eqref{eq:18}.

Theorem 1 by \citet{hanin2019universal} demonstrates that given an arbitrary continuous function $f\left(\mathbf{z}\right)$, $\mathbf{z} = \left[z_1,z_2,...,z_d\right]^{\top}$, mapping from $[0,1]^d$ to $[0,1]$, with a modulus of continuity $w_f(z)$ of $f\left(\mathbf{z}\right)$, for an arbitrary error $\varepsilon > 0$, there exists an MLP $f_N\left(\mathbf{z} \right)$ with $\sigma_{\mathrm{ReLU}}(\cdot)$, input dimension $d$, {\color{black} hidden-layer width} $d+3$, output dimension 1, and depth $2d!\left[\sqrt{d}/w_f^{-1} (\varepsilon)\right]^d$, where $w_f^{-1} (\varepsilon)$ is the inverse function of $w_f\left(z\right)$, such that $\sup_{\mathbf{z}\in\left[0, 1\right]^{d}}\left|f\left(\mathbf{z}\right) -f_N\left(\mathbf{z} \right)\right|\leq \varepsilon$. Theorem 1 by \citet{hanin2017approximating} and Proposition 53 by \citet{kratsios2022universal} provide similar results.

Using Theorem 1 by \citet{hanin2019universal} and the linear modulus of continuity $\phi_{\psi_{j},M_{\itGamma}}(x)$ in Eq.~\eqref{eq:18}, for every positive integer $H_{\it{\Pi}}$, there exists an MLP ${\it{\Pi}}_j(\cdot)$ with $\sigma_{\mathrm{ReLU}}(\cdot)$, the width and depth of ${\it{\Pi}}_j(\cdot)$ are respectively $p\left(M_{\it{\Gamma}}+1\right) + 4$ and $H_{\it{\Pi}} + 1$, such that the difference between $\psi_j \left[ {\bf{x}}(t),{\bf{u}}(0),t \right]$ and ${\it{\Pi}}_j \left[ {\bf{x}}(t),{\bf{u}}(0),t \right]$ over $\Omega_{M_{\it{\Gamma}}}\left(TB_K,B_K,T\right)$ is bounded by
\begin{equation}
    \left| {{\psi _j}\left[ {{\bf{x}}\left( t \right),{\bf{u}}\left( 0 \right),t} \right] - {{\it{\Pi}}_j}\left[ {{\bf{x}}\left( t \right),{\bf{u}}\left( 0 \right),t} \right]} \right| \le \frac{{{B_\Omega }\left[{p\left( {{M_{\it{\Gamma}} } + 1} \right) + 1}\right]^{2}{L_{{\psi _j}}}\left( {{M_{\it{\Gamma}} }} \right)}}{{\left(0.5H_{\it{\Pi}}\right)^{{1 \mathord{\left/
    {\vphantom {1 {\left[ {p\left( {{M_{\it{\Gamma}} } + 1} \right) + 1} \right]}}} \right.
    \kern-\nulldelimiterspace} {\left[ {p\left( {{M_{\it{\Gamma}} } + 1} \right) + 1} \right]}}}}}
    \label{eq:a40}
\end{equation}
for all $\mathbf{u}(t) \in K$, where $B_{\Omega} = 2\max\left(TB_K,B_K,T\right)$. By applying Eq.~\eqref{eq:a40} to all $\psi_j\left(\mathbf{x},\mathbf{u}_0,t\right)$, $j = 1, 2,..., q$, there exists an MLP ${\it{\Pi}}(\cdot)$ consisting of ${\it{\Pi}}_1(\cdot), {\it{\Pi}}_2(\cdot),..., {\it{\Pi}}_q(\cdot)$, the width of ${\it{\Pi}}(\cdot)$ is in Eq.~\eqref{eq:22} and its depth is $H_{\it{\Pi}}+1$. The difference between $\Psi \left[ {\bf{x}}(t),{\bf{u}}(0),t \right]$ and ${\it{\Pi}}\left[ {\bf{x}}(t),{\bf{u}}(0),t \right]$ over $\Omega_{M_{\it{\Gamma}}}\left(TB_K,B_K,T\right)$ is bounded by
\begin{equation}
    \left| {\Psi \left[ {{\bf{x}}\left( t \right),{\bf{u}}\left( 0 \right),t} \right] - {\it{\Pi}} \left[ {{\bf{x}}\left( t \right),{\bf{u}}\left( 0 \right),t} \right]} \right| \le \frac{{{B_\Omega }\left[{p\left( {{M_{\it{\Gamma}} } + 1} \right) + 1}\right]^{2} }}{{\left(0.5H_{\it{\Pi}}\right)^{{1 \mathord{\left/
    {\vphantom {1 {\left[ {p\left( {{M_{\it{\Gamma}} } + 1} \right) + 1} \right]}}} \right.
    \kern-\nulldelimiterspace} {\left[ {p\left( {{M_{\it{\Gamma}} } + 1} \right) + 1} \right]}}}}}\left[\sum\limits_{j = 1}^q {{L_{{\psi _j}}}\left( {{M_{\it{\Gamma}} }} \right)}\right]
    \label{eq:a41}
\end{equation}
for all $\mathbf{u}(t) \in K$. {\color{black} Eq.~\eqref{eq:20}} can be proven by combining Eq.~\eqref{eq:a39} and Eq.~\eqref{eq:a41} and noting that $\mathbf{y}(t) = {\it{\Pi}}\left[ {\bf{x}}(t),{\bf{u}}(0),t \right]$ in Eq.~\eqref{eq:1}. This completes the proof.
\hfill$\blacksquare$

\section{Proofs for Lemmas and Theorem 2 in Section \ref{subsec3.2}}
\label{appB}
\subsection{Proof of Lemma 5}
\label{appB.1}
{\bf Proof:} Since $g_1(\cdot)$ and $g_2(\cdot)$ in Eq.~\eqref{eq:23} are Lipschitz continuous over $\mathbb{R}^{2r+p}$ and the values of all $\mathbf{u}(t) \in K$ are within $\Omega_{\mathbf{u}}$, e.g., $\Omega_{\mathbf{u}} = \left[-B_K,B_K\right]^{p}$, the solution ${\mathbf{z}}_i\left(t,{\mathbf{z}}_\tau\right)$ uniquely exists and its value is bounded for all $\mathbf{u}(t) \in K$, all ${\mathbf{z}}_{\tau} \in \Omega_{\mathbf{z}}$, all $t \in [0,T]$, all $0 \leq \tau \leq t$, and \textit{i} = 1 and 2. Then, the ODEs in Eq.~\eqref{eq:23} governed by $g_1(\cdot)$ or $g_2(\cdot)$, driven by the same arbitrary $\mathbf{u}(t) \in K$ with the same initial condition ${\mathbf{z}}_0$, have unique solutions ${\mathbf{z}}_1\left(t,{\mathbf{z}}_0\right)$ and ${\mathbf{z}}_2\left(t,{\mathbf{z}}_0\right)$, respectively. ${\mathbf{z}}_2\left(t,{\mathbf{z}}_0\right) - {\mathbf{z}}_1\left(t,{\mathbf{z}}_0\right)$ can be calculated as
\begin{equation}
    \begin{aligned}
        &{{\bf{z}}_2}(t,{{\bf{z}}_0}) - {{\bf{z}}_1}(t,{{\bf{z}}_0})\\
        &\;\;\;\;\;= \int_0^t {\frac{{{\rm{d}}{{\bf{z}}_1}\left[ {t,{{\bf{z}}_2}(\tau ,{{\bf{z}}_0})} \right]}}{{{\rm{d}}\tau }}{\rm{d}}\tau } = \int_0^t {\frac{{\rm{d}}}{{{\rm{d}}\tau }}\left\{ {\int_\tau ^t {{{{\bf{z}}}_1'}\left[ {\lambda ,{{\bf{z}}_2}(\tau ,{{\bf{z}}_0})} \right]{\rm{d}}\lambda }  + {{\bf{z}}_2}(\tau ,{{\bf{z}}_0})} \right\}{\rm{d}}\tau } \\
        &\;\;\;\;\;= \int_0^t {\left\{ {\int_\tau ^t {\frac{{\partial {{{\bf{z}}}_1'}\left[ {\lambda ,{{\bf{z}}_2}(\tau ,{{\bf{z}}_0})} \right]}}{{\partial \tau }}{\rm{d}}\lambda }  + \frac{{\partial {{\bf{z}}_2}(\tau ,{{\bf{z}}_0})}}{{\partial \tau }} - {{\left. {\frac{{\partial {{\bf{z}}_1}\left[ {\lambda ,{{\bf{z}}_2}(\tau ,{{\bf{z}}_0})} \right]}}{{\partial \lambda }}} \right|}_{\lambda  = \tau }}} \right\}{\rm{d}}\tau } \\
        &\;\;\;\;\;= \int_0^t {\left( {\left\{ {\int_\tau ^t {\frac{{\partial {{{\bf{z}}}_1'}\left[ {\lambda ,{{\bf{z}}_2}(\tau ,{{\bf{z}}_0})} \right]}}{{\partial {{\bf{z}}_2}(\tau ,{{\bf{z}}_0})}}{\rm{d}}\lambda }  + {\bf{I}}} \right\}\frac{{\partial {{\bf{z}}_2}(\tau ,{{\bf{z}}_0})}}{{\partial \tau }} - {{\left. {\frac{{\partial {{\bf{z}}_1}\left[ {\lambda ,{{\bf{z}}_2}(\tau ,{{\bf{z}}_0})} \right]}}{{\partial \lambda }}} \right|}_{\lambda  = \tau }}} \right){\rm{d}}\tau } \\
        &\;\;\;\;\;= \int_0^t {{\left\{ {\int_\tau ^t {\frac{{\partial {{{\bf{z}}}_1'}\left[ {\lambda ,{{\bf{z}}_2}(\tau ,{{\bf{z}}_0})} \right]}}{{\partial {{\bf{z}}_2}(\tau ,{{\bf{z}}_0})}}{\rm{d}}\lambda }  + {\bf{I}}} \right\}
        {\begin{Bmatrix}
        {{{\bf{z}}_{2,2}}(\tau ,{{\bf{z}}_0})}\\
        {{g_2}\left[ {{{\bf{z}}_{2,1}}(\tau ,{{\bf{z}}_0}),{{\bf{z}}_{2,2}}(\tau ,{{\bf{z}}_0}),{\bf{u}}(\tau )} \right]}
        \end{Bmatrix}}}{\rm{d}}\tau } - \int_0^t {{{{\left. {\frac{{\partial {{\bf{z}}_1}\left[ {\lambda ,{{\bf{z}}_2}(\tau ,{{\bf{z}}_0})} \right]}}{{\partial \lambda }}} \right|}_{\lambda  = \tau }}}{\rm{d}}\tau },
    \end{aligned}
    \label{eq:b1}
\end{equation}
where $\mathbf{I}$ is an identity matrix.

Given an arbitrary initial condition ${\mathbf{z}}_{\tau} \in \Omega_{\mathbf{z}}$ at an arbitrary initial time instant $\tau$ ($0 \leq \tau \leq t$), it can be obtained
\begin{equation}
    \begin{aligned}
        &0 = {\left. {\frac{{{\rm{d}}{{\bf{z}}_1}\left[ {t,{{\bf{z}}_1}(\hat \tau ,{{\bf{z}}_\tau })} \right]}}{{{\rm{d}}\hat \tau }}} \right|_{\hat \tau  = \tau }} = {\left. {\frac{{\rm{d}}}{{{\rm{d}}\hat \tau }}\left\{ {\int_{\hat \tau }^t {{{{\bf{z}}}_1'}\left[ {\lambda ,{{\bf{z}}_1}(\hat \tau ,{{\bf{z}}_\tau })} \right]{\rm{d}}\lambda }  + {{\bf{z}}_1}(\hat \tau ,{{\bf{z}}_\tau })} \right\}} \right|_{\hat \tau  = \tau }}\\
        &\;\;\; = {\left. {\int_{\hat \tau }^t {\frac{{\partial {{{\bf{z}}}_1'}\left[ {\lambda ,{{\bf{z}}_1}(\hat \tau ,{{\bf{z}}_0})} \right]}}{{\partial \hat \tau }}{\rm{d}}\lambda } } \right|_{\hat \tau  = \tau }} + {\left. {\frac{{\partial {{\bf{z}}_1}(\hat \tau ,{{\bf{z}}_\tau })}}{{\partial \hat \tau }}} \right|_{\hat \tau  = \tau }} - {\left. {\frac{{\partial {{\bf{z}}_1}\left[ {\lambda ,{{\bf{z}}_1}(\hat \tau ,{{\bf{z}}_\tau })} \right]}}{{\partial \lambda }}} \right|_{\lambda  = \hat \tau ,\hat \tau  = \tau }}\\
        &\;\;\; = {\left. {\left\{ {\int_{\hat \tau }^t {\frac{{\partial {{{\bf{z}}}_1'}\left[ {\lambda ,{{\bf{z}}_1}(\hat \tau ,{{\bf{z}}_\tau })} \right]}}{{\partial {{\bf{z}}_1}(\hat \tau ,{{\bf{z}}_\tau })}}{\rm{d}}\lambda }  + {\bf{I}}} \right\}\frac{{\partial {{\bf{z}}_1}(\hat \tau ,{{\bf{z}}_\tau })}}{{\partial \hat \tau }}} \right|_{\hat \tau  = \tau }} - {\left. {\frac{{\partial {{\bf{z}}_1}\left[ {\lambda ,{{\bf{z}}_1}(\hat \tau ,{{\bf{z}}_\tau })} \right]}}{{\partial \lambda }}} \right|_{\lambda  = \hat \tau ,\hat \tau  = \tau }}\\
        &\;\;\; = {\left. {\left\{ {\int_{\hat \tau }^t {\frac{{\partial {{{\bf{z}}}_1'}\left[ {\lambda ,{{\bf{z}}_1}(\hat \tau ,{{\bf{z}}_\tau })} \right]}}{{\partial {{\bf{z}}_1}(\hat \tau ,{{\bf{z}}_\tau })}}{\rm{d}}\lambda }  + {\bf{I}}} \right\}
        {\begin{Bmatrix}
        {{{\bf{z}}_{1,2}}(\hat \tau ,{{\bf{z}}_\tau })}\\{{g_1}\left[ {{{\bf{z}}_{1,1}}(\hat \tau ,{{\bf{z}}_\tau }),{{\bf{z}}_{1,2}}(\hat \tau ,{{\bf{z}}_\tau }),{\bf{u}}(\hat \tau )} \right]}
        \end{Bmatrix}}}\right|_{\hat \tau  = \tau }} - {\left. {\frac{{\partial {{\bf{z}}_1}\left[ {\lambda ,{{\bf{z}}_1}(\hat \tau ,{{\bf{z}}_\tau })} \right]}}{{\partial \lambda }}} \right|_{\lambda  = \hat \tau ,\hat \tau  = \tau }}.
    \end{aligned}
    \label{eq:b2}
\end{equation}
Since the values of $\tau$ and ${\mathbf{z}}_{\tau} \in {\Omega}_{\mathbf{z}}$ in Eq.~\eqref{eq:b2} are arbitrary, the value of ${\mathbf{z}}_1\left({\hat{\tau}},{\mathbf{z}}_{\tau}\right) = {\mathbf{z}}_{\tau}$ at the arbitrary $\hat{\tau} = \tau$ is also arbitrary within ${\Omega}_{\mathbf{z}}$. In addition, ${\mathbf{z}}_2\left({\hat{\tau}},{\mathbf{z}}_0\right) \in {\Omega}_{\mathbf{z}}$ for all $\hat{\tau} \in [0, T]$. It is able to replace ${\mathbf{z}}_1\left({\hat{\tau}},{\mathbf{z}}_{\tau}\right)$ in Eq.~\eqref{eq:b2} with ${\mathbf{z}}_2\left({\hat{\tau}},{\mathbf{z}}_0\right)$ to obtain
\begin{equation}
    {\left. {\frac{{\partial {{\bf{z}}_1}\left[ {\lambda ,{{\bf{z}}_2}(\tau ,{{\bf{z}}_0})} \right]}}{{\partial \lambda }}} \right|_{\lambda  = \tau }} = \left\{ {\int_\tau ^t {\frac{{\partial {{{\bf{z'}}}_1}\left[ {\lambda ,{{\bf{z}}_2}(\tau ,{{\bf{z}}_0})} \right]}}{{\partial {{\bf{z}}_2}(\tau ,{{\bf{z}}_0})}}{\rm{d}}\lambda }  + {\bf{I}}} \right\}
    {\begin{Bmatrix}
    {{{\bf{z}}_{2,2}}(\tau ,{{\bf{z}}_0})}\\{{g_1}\left[ {{{\bf{z}}_{2,1}}(\tau ,{{\bf{z}}_0}),{{\bf{z}}_{2,2}}(\tau ,{{\bf{z}}_0}),{\bf{u}}(\tau )} \right]}
    \end{Bmatrix}}.
    \label{eq:b3}
\end{equation}

Substituting Eq.~\eqref{eq:b3} in Eq.~\eqref{eq:b1}, ${\mathbf{z}}_2\left(t,{\mathbf{z}}_0\right) - {\mathbf{z}}_1\left(t,{\mathbf{z}}_0\right)$ can be calculated as
\begin{equation}
    \begin{aligned}
        &{{\bf{z}}_2}(t,{{\bf{z}}_0}) - {{\bf{z}}_1}(t,{{\bf{z}}_0})\\
        &\;\; = \int_0^t{\left\{ {\int_\tau ^t {\frac{{\partial {{{\bf{z}}}_1'}\left[ {\lambda ,{{\bf{z}}_2}(\tau ,{{\bf{z}}_0})} \right]}}{{\partial {{\bf{z}}_2}(\tau ,{{\bf{z}}_0})}}{\rm{d}}\lambda }  + {\bf{I}}} \right\}}{{\begin{Bmatrix}
        {\bf{0}}\\{{g_2}\left[ {{{\bf{z}}_{2,1}}(\tau ,{{\bf{z}}_0}),{{\bf{z}}_{2,2}}(\tau ,{{\bf{z}}_0}),{\bf{u}}(\tau )} \right] - {g_1}\left[ {{{\bf{z}}_{2,1}}(\tau ,{{\bf{z}}_0}),{{\bf{z}}_{2,2}}(\tau ,{{\bf{z}}_0}),{\bf{u}}(\tau )} \right]}\end{Bmatrix}}}\rm{d}\tau \\
        &\;\;\mathop  = \limits^{(a)} \int_0^t {\frac{{\partial {{\bf{z}}_1}\left[ {t,{{\bf{z}}_2}(\tau ,{{\bf{z}}_0})} \right]}}{{\partial {{\bf{z}}_2}(\tau ,{{\bf{z}}_0})}}
        {\begin{Bmatrix}
        {\bf{0}}\\{{g_2}\left[ {{{\bf{z}}_{2,1}}(\tau ,{{\bf{z}}_0}),{{\bf{z}}_{2,2}}(\tau ,{{\bf{z}}_0}),{\bf{u}}(\tau )} \right] - {g_1}\left[ {{{\bf{z}}_{2,1}}(\tau ,{{\bf{z}}_0}),{{\bf{z}}_{2,2}}(\tau ,{{\bf{z}}_0}),{\bf{u}}(\tau )} \right]}
        \end{Bmatrix}}{\rm{d}}\tau } \\
        &\;\; = \int_0^t {\frac{{\partial {{\bf{z}}_1}\left[ {t,{{\bf{z}}_2}(\tau ,{{\bf{z}}_0})} \right]}}{{\partial {{\bf{z}}_{2,2}}(\tau ,{{\bf{z}}_0})}}\left\{ {{g_2}\left[ {{{\bf{z}}_{2,1}}(\tau ,{{\bf{z}}_0}),{{\bf{z}}_{2,2}}(\tau ,{{\bf{z}}_0}),{\bf{u}}(\tau )} \right] - {g_1}\left[ {{{\bf{z}}_{2,1}}(\tau ,{{\bf{z}}_0}),{{\bf{z}}_{2,2}}(\tau ,{{\bf{z}}_0}),{\bf{u}}(\tau )} \right]} \right\}{\rm{d}}\tau },
    \end{aligned}
    \label{eq:b4}
\end{equation}
where Step (a) holds because ${\mathbf{z}}_1\left[\tau,{\mathbf{z}}_2\left(\tau, {\mathbf{z}}_0\right)\right] = {\mathbf{z}}_2\left(\tau, {\mathbf{z}}_0\right)$. This completes the proof.
\hfill$\blacksquare$

\subsection{Proof of Lemma 6}
\label{appB.2}
{\bf Proof:} Considering Eq.~\eqref{eq:26} in Lemma 5 and let $\mathbf{Y}(t) = \partial{\mathbf{z}}_1\left[t,{\mathbf{z}}_2\left(\tau,{\mathbf{z}}_0\right)\right]/\partial{\mathbf{z}}_{2,2}\left(\tau, {\mathbf{z}}_0\right)$, $0 \leq \tau \leq t$, it is satisfied that
\begin{equation}
    \frac{{{\rm{d}}{\bf{Y}}(t)}}{{{\rm{d}}t}} = \frac{{\partial {{{\bf{z}}}_1'}\left[ {t,{{\bf{z}}_2}(\tau ,{{\bf{z}}_0})} \right]}}{{\partial {{\bf{z}}_{2,2}}(\tau ,{{\bf{z}}_0})}} = \frac{{\partial {{{\bf{z}}}_1'}\left[ {t,{{\bf{z}}_2}(\tau ,{{\bf{z}}_0})} \right]}}{{\partial {{\bf{z}}_1}\left[ {t,{{\bf{z}}_2}(\tau ,{{\bf{z}}_0})} \right]}}\frac{{\partial {{\bf{z}}_1}\left[ {t,{{\bf{z}}_2}(\tau ,{{\bf{z}}_0})} \right]}}{{\partial {{\bf{z}}_{2,2}}(\tau ,{{\bf{z}}_0})}} = \frac{{\partial {{{\bf{z}}}_1'}\left[ {t,{{\bf{z}}_2}(\tau ,{{\bf{z}}_0})} \right]}}{{\partial {{\bf{z}}_1}\left[ {t,{{\bf{z}}_2}(\tau ,{{\bf{z}}_0})} \right]}}{\bf{Y}}(t),
    \label{eq:b5}
\end{equation}
with the initial condition $\mathbf{Y}(\tau) =\left[{\mathbf{0}}_{r \times r},{\mathbf{I}}_{r \times r}\right]^{\top}$, where ${\mathbf{0}}_{r \times r}$ and ${\mathbf{I}}_{r \times r}$ denote the zero and identity matrices, respectively, and
\begin{equation}
    \frac{{\partial {{{\bf{z}}}_1'}\left[ {t,{{\bf{z}}_2}(\tau ,{{\bf{z}}_0})} \right]}}{{\partial {{\bf{z}}_1}\left[ {t,{{\bf{z}}_2}(\tau ,{{\bf{z}}_0})} \right]}} =
    {\begin{bmatrix}
        {{\bf{0}}}_{r \times r}&{{{\bf{I}}_{r \times r}}}\\
        {\frac{{\partial {g_1}\left\{ {{{\bf{z}}_{1,1}}\left[ {t,{{\bf{z}}_2}(\tau ,{{\bf{z}}_0})} \right],{{\bf{z}}_{1,2}}\left[ {t,{{\bf{z}}_2}(\tau ,{{\bf{z}}_0})} \right],{\bf{u}}(t)} \right\}}}{{\partial {{\bf{z}}_{1,1}}\left[ {t,{{\bf{z}}_2}(\tau ,{{\bf{z}}_0})} \right]}}}&{\frac{{\partial {g_1}\left\{ {{{\bf{z}}_{1,1}}\left[ {t,{{\bf{z}}_2}(\tau ,{{\bf{z}}_0})} \right],{{\bf{z}}_{1,2}}\left[ {t,{{\bf{z}}_2}(\tau ,{{\bf{z}}_0})} \right],{\bf{u}}(t)} \right\}}}{{\partial {{\bf{z}}_{1,2}}\left[ {t,{{\bf{z}}_2}(\tau ,{{\bf{z}}_0})} \right]}}}
    \end{bmatrix}}.
    \label{eq:b6}
\end{equation}
Then, $\mathbf{Y}(t)$ can be expressed as
\begin{equation}
    {\bf{Y}}(t) = \exp \left\{ {\int_\tau ^t {\frac{{\partial {{{\bf{z}}}_1'}\left[ {\hat \tau ,{{\bf{z}}_2}(\tau ,{{\bf{z}}_0})} \right]}}{{\partial {{\bf{z}}_1}\left[ {\hat \tau ,{{\bf{z}}_2}(\tau ,{{\bf{z}}_0})} \right]}}{\rm{d}}\hat \tau } } \right\}
    {\begin{bmatrix}
    {{{\bf{0}}_{r \times r}}}\\
    {{{\bf{I}}_{r \times r}}}
    \end{bmatrix}},
    \label{eq:b7}
\end{equation}
where
\begin{equation}
    \exp \left\{ {\int_\tau ^t {\frac{{\partial {{{\bf{z}}}_1'}\left[ {\hat \tau ,{{\bf{z}}_2}(\tau ,{{\bf{z}}_0})} \right]}}{{\partial {{\bf{z}}_1}\left[ {\hat \tau ,{{\bf{z}}_2}(\tau ,{{\bf{z}}_0})} \right]}}{\rm{d}}\hat \tau } } \right\} = \mathop {\lim }\limits_{\scriptstyle\Delta \hat \tau  \to 0\atop\scriptstyle n \to  + \infty } \prod\limits_{i = 1}^n {\exp \left\{ {\frac{{\partial {{{\bf{z}}}_1'}\left[ {{{\hat \tau }_i},{{\bf{z}}_2}(\tau ,{{\bf{z}}_0})} \right]}}{{\partial {{\bf{z}}_1}\left[ {{{\hat \tau }_i},{{\bf{z}}_2}(\tau ,{{\bf{z}}_0})} \right]}}\Delta \hat \tau } \right\}},
    \label{eq:b8}
\end{equation}
$\tau = \hat\tau_0 < \hat\tau_1 < \hat\tau_2 <... < \hat\tau_n = t$, $\Delta \hat \tau = \hat\tau_{i} - \hat\tau_{i-1}$, $i = 1, 2,..., n$, and
\begin{equation}
    \exp \left\{ {\frac{{\partial {{{\bf{z}}}_1'}\left[ {{{\hat \tau }_i},{{\bf{z}}_2}(\tau ,{{\bf{z}}_0})} \right]}}{{\partial {{\bf{z}}_1}\left[ {{{\hat \tau }_i},{{\bf{z}}_2}(\tau ,{{\bf{z}}_0})} \right]}}\Delta \hat \tau } \right\} = \sum\limits_{m = 0}^{ + \infty } {\frac{1}{{m!}}} {\left\{ {\frac{{\partial {{{\bf{z}}}_1'}\left[ {{{\hat \tau }_i},{{\bf{z}}_2}(\tau ,{{\bf{z}}_0})} \right]}}{{\partial {{\bf{z}}_1}\left[ {{{\hat \tau }_i},{{\bf{z}}_2}(\tau ,{{\bf{z}}_0})} \right]}}\Delta \hat \tau } \right\}^m}.
    \label{eq:b9}
\end{equation}
Substituting Eq.~\eqref{eq:b7} into Eq.~\eqref{eq:26}, ${\mathbf{z}}_2\left(t, {\mathbf{z}}_0\right) - {\mathbf{z}}_1\left(t, {\mathbf{z}}_0\right)$ can be calculated as
\begin{equation}
    {{\bf{z}}_2}(t,{{\bf{z}}_0}) - {{\bf{z}}_1}(t,{{\bf{z}}_0}) = \int_0^t {\exp \left\{ {\int_\tau ^t {\frac{{\partial {{{\bf{z}}}_1'}\left[ {\hat \tau ,{{\bf{z}}_2}(\tau ,{{\bf{z}}_0})} \right]}}{{\partial {{\bf{z}}_1}\left[ {\hat \tau ,{{\bf{z}}_2}(\tau ,{{\bf{z}}_0})} \right]}}{\rm{d}}\hat \tau } } \right\}{\bf{\Delta }}g(\tau ,{{\bf{z}}_0}){\rm{d}}\tau},
    \label{eq:b10}
\end{equation}
where
\begin{equation}
    {\bf{\Delta }}g(\tau ,{{\bf{z}}_0}) = 
    {\begin{Bmatrix}
    {{{\bf{0}}_{r \times 1}}}\\{{g_2}\left[ {{{\bf{z}}_{2,1}}(\tau ,{{\bf{z}}_0}),{{\bf{z}}_{2,2}}(\tau ,{{\bf{z}}_0}),{\bf{u}}(\tau )} \right] - {g_1}\left[ {{{\bf{z}}_{2,1}}(\tau ,{{\bf{z}}_0}),{{\bf{z}}_{2,2}}(\tau ,{{\bf{z}}_0}),{\bf{u}}(\tau )} \right]}
    \end{Bmatrix}}.
    \label{eq:b11}
\end{equation}

Since $g_1(\cdot)$ is Lipschitz continuous, from Eq.~\eqref{eq:b6}, it can be obtained
\begin{equation}
    \begin{aligned}
        &\left| {\frac{{\partial {{{\bf{z}}}_1'}\left[ {t,{{\bf{z}}_2}(\tau,{{\bf{z}}_0})} \right]}}{{\partial {{\bf{z}}_1}\left[ {t,{{\bf{z}}_2}(\tau ,{{\bf{z}}_0})} \right]}}
        {\begin{bmatrix}{{{\bf{a}}_1}}\\
        {{{\bf{a}}_2}}\end{bmatrix}}} \right| = \left|{\begin{bmatrix}{{{\bf{a}}_2}}\\
        {\frac{{\partial {g_1}}}{{\partial {{\bf{z}}_{1,1}}}}{{\bf{a}}_1} + \frac{{\partial {g_1}}}{{\partial {{\bf{z}}_{1,2}}}}{{\bf{a}}_2}}\end{bmatrix}}\right| \le \left|{\begin{bmatrix}{\bf{0}}\\
        {\frac{{\partial {g_1}}}{{\partial {{\bf{z}}_{1,1}}}}{{\bf{a}}_1}}\end{bmatrix}}\right| + \left|{\begin{bmatrix}{{{\bf{a}}_2}}\\{\frac{{\partial {g_1}}}{{\partial {{\bf{z}}_{1,2}}}}{{\bf{a}}_2}}
        \end{bmatrix}}\right|\\
        &\;\;\;\;\;\;\;\;\;\;\;\;\;\;\;\;\;\;\;\;\;\;\;\;\;\;\;\;\;\;\; \le \left| {{\begin{bmatrix}{\bf{0}}\\{{L_{{g_1},1}}{{\bf{a}}_1}}\end{bmatrix}}} \right| + \left| {{\begin{bmatrix}{{{\bf{a}}_2}}\\{{L_{{g_1},2}}{{\bf{a}}_2}}\end{bmatrix}}} \right| \le {L_{{g_1}}}\left| {{\begin{bmatrix}{{{\bf{a}}_1}}\\{{{\bf{a}}_2}}\end{bmatrix}}} \right|,
    \end{aligned}
    \label{eq:b12}
\end{equation}
where $L_{g_1} = \max\left[L_{g_1,1},\left(L_{g_1,2} +1\right)\right]$, $L_{g_1,1}$ and $L_{g_1,2}$ are respectively the two Lipschitz constants of $g_1(\cdot)$ with respect to its first and second arguments satisfying
\begin{equation}
    \left| {{g_1}\left( {{{\bf{b}}_1},{{\bf{z}}_{1,2}}\left[ {t,{{\bf{z}}_2}(\tau ,{{\bf{z}}_0})} \right],{\bf{u}}(t)} \right) - {g_1}\left( {{{\bf{b}}_2},{{\bf{z}}_{1,2}}\left[ {t,{{\bf{z}}_2}(\tau ,{{\bf{z}}_0})} \right],{\bf{u}}(t)} \right)} \right| \le {L_{{g_1},1}}\left| {{{\bf{b}}_1} - {{\bf{b}}_2}} \right|,
    \label{eq:b13}
\end{equation}
\begin{equation}
    \left| {{g_1}\left( {{{\bf{z}}_{1,1}}\left[ {t,{{\bf{z}}_2}(\tau ,{{\bf{z}}_0})} \right],{{\bf{b}}_1},{\bf{u}}(t)} \right) - {g_1}\left( {{{\bf{z}}_{1,1}}\left[ {t,{{\bf{z}}_2}(\tau ,{{\bf{z}}_0})} \right],{{\bf{b}}_2},{\bf{u}}(t)} \right)} \right| \le {L_{{g_1},2}}\left| {{{\bf{b}}_1} - {{\bf{b}}_2}} \right|,
    \label{eq:b14}
\end{equation}
and $\mathbf{a}_1$, $\mathbf{a}_2$, $\mathbf{b}_1$, and $\mathbf{b}_2$ are arbitrary \textit{r}-dimensional vectors.

Utilizing Eqs.~\eqref{eq:b8}-\eqref{eq:b12}, $\left|{\mathbf{z}}_2\left(t, {\mathbf{z}}_0\right) - {\mathbf{z}}_1\left(t, {\mathbf{z}}_0\right)\right|$ can be bounded by
\begin{equation}
    \begin{aligned}
        &\left| {{{\bf{z}}_2}(t,{{\bf{z}}_0}) - {{\bf{z}}_1}(t,{{\bf{z}}_0})} \right|\\
        &\;\; \le \int_0^t {\left| {\exp \left\{ {\int_\tau ^t {\frac{{\partial {{{\bf{z}}}_1'}\left[ {\hat \tau ,{{\bf{z}}_2}(\tau ,{{\bf{z}}_0})} \right]}}{{\partial {{\bf{z}}_1}\left[ {\hat \tau ,{{\bf{z}}_2}(\tau ,{{\bf{z}}_0})} \right]}}{\rm{d}}\hat \tau } } \right\}{\bf{\Delta }}g(\tau ,{{\bf{z}}_0})} \right|{\rm{d}}\tau } = \int_0^t {\left| {\mathop {\lim }\limits_{\scriptstyle\Delta \hat \tau  \to 0\atop\scriptstyle n \to  + \infty } \prod\limits_{i = 1}^n {\exp \left\{ {\frac{{\partial {{{\bf{z}}}_1'}\left[ {{{\hat \tau }_i},{{\bf{z}}_2}(\tau ,{{\bf{z}}_0})} \right]}}{{\partial {{\bf{z}}_1}\left[ {{{\hat \tau }_i},{{\bf{z}}_2}(\tau ,{{\bf{z}}_0})} \right]}}\Delta \hat \tau } \right\}} {\bf{\Delta }}g(\tau ,{{\bf{z}}_0})} \right|{\rm{d}}\tau } \\
        &\;\; \le \int_0^t {\mathop {\lim }\limits_{\scriptstyle\Delta \hat \tau  \to 0\atop\scriptstyle n \to  + \infty } \left| {\prod\limits_{i = 1}^n {\left( {\sum\limits_{m = 0}^{ + \infty } {\frac{1}{{m!}}} {{\left\{ {\frac{{\partial {{{\bf{z}}}_1'}\left[ {{{\hat \tau }_i},{{\bf{z}}_2}(\tau ,{{\bf{z}}_0})} \right]}}{{\partial {{\bf{z}}_1}\left[ {{{\hat \tau }_i},{{\bf{z}}_2}(\tau ,{{\bf{z}}_0})} \right]}}\Delta \hat \tau } \right\}}^m}} \right)} {\bf{\Delta }}g(\tau ,{{\bf{z}}_0})} \right|{\rm{d}}\tau } \\
        &\;\; = \int_0^t {\mathop {\lim }\limits_{\scriptstyle\Delta \hat \tau  \to 0\atop\scriptstyle n \to  + \infty } \left| {\left( {\sum\limits_{m = 0}^{ + \infty } {\frac{1}{{m!}}} {{\left\{ {\frac{{\partial {{{\bf{z}}}_1'}\left[ {{{\hat \tau }_n},{{\bf{z}}_2}(\tau ,{{\bf{z}}_0})} \right]}}{{\partial {{\bf{z}}_1}\left[ {{{\hat \tau }_n},{{\bf{z}}_2}(\tau ,{{\bf{z}}_0})} \right]}}\Delta \hat \tau} \right\}}^m}} \right)\prod\limits_{i = 1}^{n - 1} {\left( {\sum\limits_{m = 0}^{ + \infty } {\frac{1}{{m!}}} {{\left\{ {\frac{{\partial {{{\bf{z}}}_1'}\left[ {{{\hat \tau }_i},{{\bf{z}}_2}(\tau ,{{\bf{z}}_0})} \right]}}{{\partial {{\bf{z}}_1}\left[ {{{\hat \tau }_i},{{\bf{z}}_2}(\tau ,{{\bf{z}}_0})} \right]}}\Delta \hat \tau } \right\}}^m}} \right)} {\bf{\Delta }}g(\tau ,{{\bf{z}}_0})} \right|{\rm{d}}\tau } \\
        &\;\;\mathop  \le \limits^{(a)} \int_0^t {\mathop {\lim }\limits_{\scriptstyle\Delta \hat \tau  \to 0\atop\scriptstyle n \to  + \infty } \left[ {\sum\limits_{m = 0}^{ + \infty } {\frac{1}{{m!}}} {{\left( {{L_{{g_1}}}\Delta \hat \tau } \right)}^m}} \right]\left| {\prod\limits_{i = 1}^{n - 1} {\left( {\sum\limits_{m = 0}^{ + \infty } {\frac{1}{{m!}}} {{\left\{ {\frac{{\partial {{{\bf{z}}}_1'}\left[ {{{\hat \tau }_i},{{\bf{z}}_2}(\tau ,{{\bf{z}}_0})} \right]}}{{\partial {{\bf{z}}_1}\left[ {{{\hat \tau }_i},{{\bf{z}}_2}(\tau ,{{\bf{z}}_0})} \right]}}\Delta \hat \tau } \right\}}^m}} \right)} {\bf{\Delta }}g(\tau ,{{\bf{z}}_0})} \right|{\rm{d}}\tau } \\
        &\;\; \le \int_0^t {\mathop {\lim }\limits_{\scriptstyle\Delta \hat \tau  \to 0\atop\scriptstyle n \to +\infty} {{\left[ {\sum\limits_{m = 0}^{ + \infty } {\frac{1}{{m!}}} {{\left( {{L_{{g_1}}}\Delta \hat \tau } \right)}^m}} \right]}^{n}}\left| {{\bf{\Delta }}g(\tau ,{{\bf{z}}_0})} \right|{\rm{d}}\tau } = \int_0^t {\exp \left( {\int_\tau ^t {{L_{{g_1}}}{\rm{d}}\hat \tau } } \right)\left| {{\bf{\Delta }}g(\tau ,{{\bf{z}}_0})} \right|{\rm{d}}\tau } \\
        &\;\; \le {e^{T{L_{{g_1}}}}}T\mathop {\max }\limits_{\tau  \in [0,T]} \left| {{\bf{\Delta }}g(\tau,{{\bf{z}}_0})} \right|,
    \end{aligned}
    \label{eq:b15}
\end{equation}
where Step (\textit{a}) holds from Eq.~\eqref{eq:b12}. Substituting Eq.~\eqref{eq:b11} into Eq.~\eqref{eq:b15}, Eq.~\eqref{eq:27} can be proven.

If the dynamical system governed by $g_1(\cdot)$ is uniformly asymptotically incrementally stable for all $\mathbf{u}(t) \in K$ on the domain $\Omega_{\mathbf{z}}$ in Lemma 5 with a set of stability functions $\beta_k\left(h,t\right)$, $k = 1, 2,..., 2r$, then, from Eq.~\eqref{eq:26}, $\left|{\mathbf{z}}_2\left(t, {\mathbf{z}}_0\right) - {\mathbf{z}}_1\left(t, {\mathbf{z}}_0\right)\right|$ can be bounded by
\begin{equation}
    \begin{aligned}
        &\left| {{{\bf{z}}_2}(t,{{\bf{z}}_0}) - {{\bf{z}}_1}(t,{{\bf{z}}_0})} \right|\\
        &\;\; \le \int_0^t {\left| {\frac{{\partial {{\bf{z}}_1}\left[ {t,{{\bf{z}}_2}(\tau ,{{\bf{z}}_0})} \right]}}{{\partial {{\bf{z}}_2}(\tau ,{{\bf{z}}_0})}}{\begin{Bmatrix}{\mathbf{0}}_{r \times 1}\\{{g_2}\left[ {{{\bf{z}}_{2,1}}(\tau ,{{\bf{z}}_0}),{{\bf{z}}_{2,2}}(\tau ,{{\bf{z}}_0}),{\bf{u}}(\tau )} \right] - {g_1}\left[ {{{\bf{z}}_{2,1}}(\tau ,{{\bf{z}}_0}),{{\bf{z}}_{2,2}}(\tau ,{{\bf{z}}_0}),{\bf{u}}(\tau )} \right]}\end{Bmatrix}}} \right|{\rm{d}}\tau }\\
        &\;\; \le \int_0^t{{\left|{\left\{ {\frac{{\partial {{\bf{z}}_1}\left[ {t,{{\bf{z}}_2}(\tau ,{{\bf{z}}_0})} \right]}}{{\partial {{\bf{z}}_2}(\tau ,{{\bf{z}}_0})}}} \right\}^\top}\right|}_{2,1}{{\left| {{g_2}\left[ {{{\bf{z}}_{2,1}}(\tau ,{{\bf{z}}_0}),{{\bf{z}}_{2,2}}(\tau ,{{\bf{z}}_0}),{\bf{u}}(\tau )} \right] - {g_1}\left[ {{{\bf{z}}_{2,1}}(\tau ,{{\bf{z}}_0}),{{\bf{z}}_{2,2}}(\tau ,{{\bf{z}}_0}),{\bf{u}}(\tau )} \right]} \right|}_2}{\rm{d}}\tau } \\
        &\;\; \le \mathop {\max }\limits_{\tau  \in [0,T]} {\left| {{\bf{\Delta }}g(\tau ,{{\bf{z}}_0})} \right|_2}\int_0^t {{{\left|\left\{ {\frac{{\partial {{\bf{z}}_1}\left[ {t,{{\bf{z}}_2}(\tau ,{{\bf{z}}_0})} \right]}}{{\partial {{\bf{z}}_2}(\tau ,{{\bf{z}}_0})}}} \right\}^\top\right|}_{2,1}}{\rm{d}}\tau },
    \end{aligned}
    \label{eq:b16}
\end{equation}
where
\begin{equation}
    \begin{aligned}
    &{\left|\left\{ {\frac{{\partial {{\bf{z}}_1}\left[ {t,{{\bf{z}}_2}(\tau ,{{\bf{z}}_0})} \right]}}{{\partial {{\bf{z}}_2}(\tau ,{{\bf{z}}_0})}}} \right\}^\top\right|_{2,1}} = \sum\limits_{k = 1}^{2r} {{{\left| {\frac{{\partial {z_{1,k}}\left( {t,{{\bf{z}}_2}} \right)}}{{\partial {{\bf{z}}_2}}}} \right|}_2}}  = \sum\limits_{k = 1}^{2r} {\left| {\frac{{\partial {z_{1,k}}\left( {t,{{\bf{z}}_2}} \right)}}{{\partial {{\bf{z}}_2}}} \cdot {{\bf{\Delta }}_k}} \right|} \\
    &\;\;\;\;\;\;\;\;\;\;\;\;\;\;\;\;\;\;\;\;\;\;\;\;\;\;\;\;\;\;\;\;\;\; = \sum\limits_{k = 1}^{2r} {\mathop {\lim }\limits_{v \to 0} \frac{{\left| {{z_{1,k}}\left( {t,{{\bf{z}}_2} + v{{\bf{\Delta }}_k}} \right) - {z_{1,k}}\left( {t,{{\bf{z}}_2}} \right)} \right|}}{{{{\left| {v{{\bf{\Delta }}_k}} \right|}_2}}}}\\
    &\;\;\;\;\;\;\;\;\;\;\;\;\;\;\;\;\;\;\;\;\;\;\;\;\;\;\;\;\;\;\;\;\;\; \mathop \le \limits^{(a)} \sum\limits_{k = 1}^{2r} {\mathop {\lim }\limits_{v \to 0} \frac{{{\beta _k}({{\left| {v{{\bf{\Delta }}_k}} \right|}_2},t-\tau )}}{{{{\left| {v{{\bf{\Delta }}_k}} \right|}_2}}}}  = {\left. {\sum\limits_{k = 1}^{2r} {\frac{\partial }{{\partial h}}{\beta _k}(h,t-\tau)} } \right|_{h = {0^ + }}},
    \end{aligned}
    \label{eq:b17}
\end{equation}
$z_{1,k}\left(t,{\mathbf{z}}_2\right)$ denotes the $k^{\text{th}}$ element of ${\mathbf{z}}_{1}\left(t,{\mathbf{z}}_2\right)$, ${\mathbf{\Delta}}_k$ is the unit vector in the direction of the partial derivative of $z_{1,k}\left(t,{\mathbf{z}}_2\right)$ with respect to ${\mathbf{z}}_2$, and Step (\textit{a}) holds from Eq.~\eqref{eq:24} for all ${\mathbf{z}}_2 \in \Omega_{\mathbf{z}}$. Substituting Eq.~\eqref{eq:b17} into Eq.~\eqref{eq:b16}, $\left|{\mathbf{z}}_2\left(t, {\mathbf{z}}_0\right) - {\mathbf{z}}_1\left(t, {\mathbf{z}}_0\right)\right|$ can be bounded by
\begin{equation}
    \left| {{{\bf{z}}_2}(t,{{\bf{z}}_0}) - {{\bf{z}}_1}(t,{{\bf{z}}_0})} \right| \le \mathop {\max }\limits_{\tau  \in [0,T]} {\left| {{\bf{\Delta }}g(\tau ,{{\bf{z}}_0})} \right|_2}\int_0^t {{\left. {\sum\limits_{k = 1}^{2r} {\frac{\partial }{{\partial h}}{\beta _k}(h,t-\tau)}}\right|_{h = {0^+}}}{\rm{d}}\tau } \mathop \le \limits^{(a)} \mathop {\max }\limits_{\tau  \in [0,T]} {\left| {{\bf{\Delta }}g(\tau ,{{\bf{z}}_0})} \right|_2}{B_\beta },
    \label{eq:b18}
\end{equation}
for all $\mathbf{u}(t) \in K$ and all $t \in [0,T]$, where Step (\textit{a}) holds from $\left.\partial \beta_k(h, t-\tau) / \partial h\right|_{h = 0^+} \ge 0$ and $B_\beta$ is the stability bound in Eq.~\eqref{eq:25}. Eq.~\eqref{eq:28} is proven by substituting Eq.~\eqref{eq:b11} into Eq.~\eqref{eq:b18}.
\hfill$\blacksquare$

\subsection{Proof of Theorem 2}
\label{appB.3}
{\bf Proof:} The difference between ${\mathbf{y}}(t)$ and ${\hat{\mathbf{y}}}(t)$ can be bounded by
\begin{equation}
    \left| {{\bf{y}}(t) - {\hat{\bf y}}(t)} \right| \le \left| {h\left[ {{\bf{x}}(t)} \right] - h\left[ {{\hat{\bf x}}(t)} \right]} \right| + \left| {{\it{\Pi}} \left[ {{\bf{x}}(t),{\bf{u}}(0),t} \right] - h\left[ {{\bf{x}}(t)} \right]} \right|,
    \label{eq:b19}
\end{equation}
where $\mathbf{x}(t)$ is the solution of the second-order ODE in Eq.~\eqref{eq:1}.

First, the first part on the right side of Eq.~\eqref{eq:b19} is considered. According to the assumptions in Theorem 2, {\color{black} for every compact domain $\Omega \subset {\mathbb{R}}^{2r+p}$, each $g_{j}(\cdot)$ of $g(\cdot)$ has a Barron function extension $\tilde{g}_{j,\Omega}(\cdot) = g_j(\cdot)$ on $\Omega$, and the spectral Barron norms $\|{g}_{j}\|_{\mathcal{B},\Omega} =  \int_{{\mathbb{R}^{2r + p}}} {\left(1+\left| \boldsymbol{\upomega} \right|\right)} \left| {{\mathcal{H}_{{{\tilde g}_{j,\Omega}}}}(\boldsymbol{\upomega})} \right|{\rm{d}}\boldsymbol{\upomega}$ \citep{siegel2023characterization} of all extensions are uniformly bounded, where $\mathcal{H}_{{\tilde{g}}_{j,\Omega}}\left(\boldsymbol{\upomega}\right)$ is the Fourier transform of ${\tilde{g}}_{j,\Omega}(\cdot)$ and $\boldsymbol{\upomega} = [\omega_1, \omega_2,\dots, \omega_{2r+p}]^{\top}$ is a radian frequency vector. Thus, $g(\cdot)$ is bounded and continuously differentiable with bounded derivatives over $\mathbb{R}^{2r+p}$ \citep{barron1993universal}.} The solutions ${\hat{\mathbf{x}}}(t)$ and ${\hat{\mathbf{x}}}'(t)$ to Eq.~\eqref{eq:29} driven by all $\mathbf{u}(t) \in K$, with the initial conditions ${\hat{\mathbf{x}}}(0) = {\hat{\mathbf{x}}}'(0) = \mathbf{0}$, are unique and they can be bounded by two constants $B_{\hat{\mathbf{x}}}$ and $B_{\hat{\mathbf{x}}'}$, respectively, that is $\left|{\hat{x}_j}(t)\right| \leq  B_{\hat{\mathbf{x}}}$ and $\left|{\hat{x}_j'}(t)\right| \leq  B_{\hat{\mathbf{x}}'}$ for all $\mathbf{u}(t) \in K$, all $t \in [0,T]$, and $j = 1, 2,..., r$. Since the dynamical system in Eq.~\eqref{eq:29} is uniformly asymptotically incrementally stable for all $\mathbf{u}(t) \in K$ on a domain $\Omega\left(B_{\mathbf{x}}^{r}, B_{{\mathbf{x}}'}^{r}, B_K^p\right) = \left[-B_{\mathbf{x}}, B_{\mathbf{x}}\right]^{r} \times \left[-B_{{\mathbf{x}}'}, B_{{\mathbf{x}}'}\right]^{r} \times \left[-B_K, B_K\right]^{p}$, where $B_K$ is the bound of all $\mathbf{u}(t) \in K$ in Assumption 2, $B_{{\mathbf{x}}} = \alpha B_{\beta{g}} + B_{\hat{\mathbf{x}}}$, $B_{{\mathbf{x}
}'} = \alpha B_{\beta{g}} + B_{\hat{\mathbf{x}}'}$, $\alpha$ is a positive coefficient, and $B_{\beta{g}}$ is the stability bound of the dynamical system in Eq.~\eqref{eq:29}, which is calculated by Eq.~\eqref{eq:25}, if the difference between the MLP ${\it{\Gamma}}(\cdot)$ in Eq.~\eqref{eq:1} and $g(\cdot)$ in Eq.~\eqref{eq:29} over $\Omega\left(B_{\mathbf{x}}^{r}, B_{{\mathbf{x}}'}^{r}, B_K^p\right)$ is bounded by
\begin{equation}
    {\left| {{\it{\Gamma}} ({\bf{x}},{\bf{x'}},{\bf{u}}) - g({\bf{x}},{\bf{x'}},{\bf{u}})} \right|_2} \le \alpha,
    \label{eq:b20}
\end{equation}
then, according to Eq.~\eqref{eq:28}, ${{\mathbf{x}}}(t)$ and ${{\mathbf{x}}}'(t)$ from Eq.~\eqref{eq:1} and ${\hat{\mathbf{x}}}(t)$ and ${\hat{\mathbf{x}}}'(t)$ from Eq.~\eqref{eq:29}, driven by the same arbitrary $\mathbf{u}(t) \in K$, with the same initial conditions ${{\mathbf{x}}}(0) = \mathbf{x}'(0) ={\hat{\mathbf{x}}}(0) = {\hat{\mathbf{x}}}'(0) = \mathbf{0}$, satisfy
\begin{equation}
    \left| {{\bf{x}}(t) - {\hat{\bf x}}(t)} \right| + \left| {{\bf{x'}}(t) - {\hat{\bf x}'}(t)} \right| \le \mathop {\max }\limits_{\tau  \in [0,T]} {\left| {{\it{\Gamma}} \left[ {{\bf{x}}(\tau ),{\bf{x'}}(\tau ),{\bf{u}}(\tau )} \right] - g\left[ {{\bf{x}}(\tau ),{\bf{x'}}(\tau ),{\bf{u}}(\tau )} \right]} \right|_2}{B_{{\beta _g}}} \le \alpha{B_{{\beta _g}}},
    \label{eq:b21}
\end{equation}
and the values of ${{\mathbf{x}}}(t)$ and ${{\mathbf{x}}}'(t)$ are contained in $\Omega\left(B_{\mathbf{x}}^{r}, B_{{\mathbf{x}}'}^{r}, B_K^p\right)$ for all $\mathbf{u}(t) \in K$ and all $t \in [0,T]$.

%%%%
% Since each $g_j(\cdot)$ of $g(\cdot)$ {\color{black} has continuous partial derivatives of order $\floor{r+0.5p}+3$ over $\mathbb{R}^{2r+p}$}, there exists a corresponding ${\tilde{g}}_j(\cdot)$ equaling $g_j(\cdot)$ over $\Omega\left(B_{\mathbf{x}}^{r}, B_{{\mathbf{x}}'}^{r}, B_K^p\right)$ and the first moment of the Fourier transform $\mathcal{H}_{{\tilde{g}}_j}\left(\boldsymbol{\upomega}\right)$ of ${\tilde{g}}_j(\cdot)$ is finite \citep{barron1993universal}, that is
% \begin{equation}
%     \int_{{\mathbb{R}^{2r + p}}} {\left| \boldsymbol{\upomega} \right|_2} \left| {{\mathcal{H}_{{{\tilde g}_j}}}(\boldsymbol{\upomega})} \right|{\rm{d}}\boldsymbol{\upomega} <  + \infty ,\;\;j = 1,2, \ldots ,r,
%     \label{eq:b22}
% \end{equation}
% where $\boldsymbol{\upomega} = \left[\omega_1, \omega_2,..., \omega_{2r+p}\right]^{\top}$. 
%%%%

{\color{black} Since each $g_{j}(\cdot)$ of $g(\cdot)$ has a Barron function extension $\tilde{g}_{j,\Omega}(\cdot) = g_j(\cdot)$ on $\Omega\left(B_{\mathbf{x}}^{r}, B_{{\mathbf{x}}'}^{r}, B_K^p\right)$ with a finite spectral Barron norm,} based on Theorem 2.2 in \citet{yukich2002sup} and the fact that four ReLU activation functions can form a total variation activation function that has a bounded support, e.g., $\left[\sigma_{\mathrm{ReLU}}(x) - \sigma_{\mathrm{ReLU}}(x - 1)\right] - \left[\sigma_{\mathrm{ReLU}}(x - 2) - \sigma_{\mathrm{ReLU}}(x - 3)\right]$, for an arbitrary $\varepsilon_{1}$ satisfying $0 < \varepsilon_{1} < \alpha/r$, there exist $r$ one-hidden-layer MLPs ${\it{\Gamma}}_j(\cdot)$ with $\sigma_{\mathrm{ReLU}}(\cdot)$, $j = 1, 2,...,r$, the widths of whose input and output layers are respectively $2r + p$ and 1, and the width of the hidden layer for each ${\it{\Gamma}}_j(\cdot)$ is at most
\begin{equation}
    {w_{{{\it{\Gamma}} _j}}} \le 8\left\lceil {C_{{{\it{\Gamma}} _j}}^2\varepsilon _1^{ - 2}} \right\rceil,
    \label{eq:b23}
\end{equation}
such that the difference between ${\it{\Gamma}}_j(\cdot)$ and $g_j(\cdot)$ over $\Omega\left(B_{\mathbf{x}}^{r}, B_{{\mathbf{x}}'}^{r}, B_K^p\right)$ is bounded by
\begin{equation}
    \left| {{{\it{\Gamma}} _j}({\bf{x}},{\bf{x'}},{\bf{u}}) - {g_j}({\bf{x}},{\bf{x'}},{\bf{u}})} \right| \le {\varepsilon_1}.
    \label{eq:b24}
\end{equation}
In Eq.~\eqref{eq:b23}, $C_{{\it{\Gamma}}_j} = C_{{\it{\Gamma}}_j}\left[\mathcal{H}_{{\tilde{g}}_j}\left(\boldsymbol{\upomega}\right), \sigma_{\mathrm{ReLU}}(\cdot),2r+p, \Omega\left(B_{\mathbf{x}}^{r}, B_{{\mathbf{x}}'}^{r}, B_K^p\right)\right]$ is a constant only depending on the Fourier transform $\mathcal{H}_{{\tilde{g}}_j}\left(\boldsymbol{\upomega}\right)$ of ${\tilde{g}}_j(\cdot)$ corresponding to $g_j(\cdot)$, the domain $\Omega\left(B_{\mathbf{x}}^{r}, B_{{\mathbf{x}}'}^{r}, B_K^p\right)$, the activation function $\sigma_{\mathrm{ReLU}}(\cdot)$, and the number $2r+p$ of the arguments of $g_j(\cdot)$.

Then, from {\color{black}Eqs.~\eqref{eq:b20}-\eqref{eq:b23}}, for an arbitrary $\varepsilon_{1}$ satisfying $0 < \varepsilon_{1} < \alpha/r$, there exists a one-hidden-layer MLP ${\it{\Gamma}}(\cdot)$ consisting of above ${\it{\Gamma}}_1(\cdot),{\it{\Gamma}}_2(\cdot),...,{\it{\Gamma}}_r(\cdot)$, the widths of the input and output layers of ${\it{\Gamma}}(\cdot)$ are respectively $2r + p$ and $r$, and the width of its hidden layer is at most
\begin{equation}
    {w_{{{\it{\Gamma}}}}} \le 8r\left\lceil {C_{{{\it{\Gamma}}}}^2\varepsilon _1^{-2}} \right\rceil,
    \label{eq:b25}
\end{equation}
where ${C_{\it{\Gamma}} } = \max \left( {{C_{{{\it{\Gamma}}_1}}},{C_{{{\it{\Gamma}}_2}}}, \ldots ,{C_{{{\it{\Gamma}}_r}}}} \right)$, such that for all $\mathbf{u}(t) \in K$, the corresponding solutions $\mathbf{x}(t)$ from Eq.~\eqref{eq:1} and ${\hat{\mathbf{x}}}(t)$ from Eq.~\eqref{eq:29}, with the initial conditions $\mathbf{x}(0) = {\mathbf{x}}'(0) = {\hat{\mathbf{x}}}(0) = {{\hat{\mathbf{x}}}}'(0) = \mathbf{0}$, satisfy
\begin{equation}
    \begin{aligned}
        &\left| {{\bf{x}}(t) - {\hat{\bf x}}(t)} \right| \le \mathop {\max }\limits_{\tau  \in [0,T]} {\left| {{\it{\Gamma}} \left[ {{\bf{x}}(\tau ),{\bf{x}}(\tau ),{\bf{u}}(\tau )} \right] - g\left[ {{\bf{x}}(\tau ),{\bf{x}}(\tau ),{\bf{u}}(\tau )} \right]} \right|_2}{B_{{\beta _g}}}\\
        &\;\;\;\;\;\;\;\;\;\;\;\;\;\;\;\;\; \le {B_{{\beta _g}}}\sqrt {\sum\limits_{j = 1}^r {\mathop {\max }\limits_{\tau  \in [0,T]} \left\{ {{{\it{\Gamma}}_j}\left[ {{\bf{x}}(\tau ),{\bf{x}}(\tau ),{\bf{u}}(\tau )} \right] - {g_j}\left[ {{\bf{x}}(\tau ),{\bf{x}}(\tau ),{\bf{u}}(\tau )} \right]} \right\}^2} }  \le {B_{{\beta _g}}}\sqrt r {\varepsilon _1}
    \end{aligned}
    \label{eq:b26}
\end{equation}
for all $t \in [0,T]$, where Eq.~\eqref{eq:b20} is satisfied by
\begin{equation}
    \begin{aligned}
        &\mathop {\max }\limits_{\tau  \in [0,T]} {\left| {{\it{\Gamma}}\left[ {{\bf{x}}(\tau ),{\bf{x}}(\tau ),{\bf{u}}(\tau )} \right] - g\left[ {{\bf{x}}(\tau ),{\bf{x}}(\tau ),{\bf{u}}(\tau )} \right]} \right|_2}\\
        &\;\;\;\; \le \sqrt {\sum\limits_{j = 1}^r {\mathop {\max }\limits_{\tau  \in [0,T]} \left\{ {{{\it{\Gamma}}_j}\left[ {{\bf{x}}(\tau ),{\bf{x}}(\tau ),{\bf{u}}(\tau )} \right] - {g_j}\left[ {{\bf{x}}(\tau ),{\bf{x}}(\tau ),{\bf{u}}(\tau )} \right]} \right\}^2} }  \le \sqrt r {\varepsilon _1} \le \alpha{r^{ - 0.5}} \le \alpha.
    \end{aligned}
    \label{eq:b27}
\end{equation}

{\color{black} Since each $h_j(\cdot)$ of $h(\cdot)$, $j = 1, 2,..., q$, has a Barron function extension ${\tilde{h}}_{j,\Omega}(\cdot) = h_j(\cdot)$ on $\Omega\left(B_{\mathbf{x}}^{r}\right) = \left[-B_{\mathbf{x}}, B_{\mathbf{x}}\right]^{r}$, and the spectral Barron norm $\|{h}_{j}\|_{\mathcal{B},\Omega} =  \int_{{\mathbb{R}^r}} {\left(1+\left| \boldsymbol{\upomega} \right|\right)} \left| {{\mathcal{H}_{{{\tilde h}_{j,\Omega}}}}(\boldsymbol{\upomega})} \right|{\rm{d}}\boldsymbol{\upomega}$ of each ${\tilde{h}}_{j,\Omega}(\cdot)$ is finite, $h(\cdot)$ is Lipschitz continuous with a Lipschitz constant $L_h$.} The first term on the right side of Eq.~\eqref{eq:b19} can be bounded by
\begin{equation}
    \left| {h\left[ {{\hat{\bf x}}(t)} \right] - h\left[ {{\bf{x}}(t)} \right]} \right| \le {L_h}\left| {{\hat{\bf x}}(t) - {\bf{x}}(t)} \right| \le {L_h}{B_{{\beta _g}}}\sqrt r {\varepsilon _1}
    \label{eq:b28}
\end{equation}
for all $t \in [0,T]$.

Subsequently, the second term on the right side of Eq.~\eqref{eq:b19} is taken into account. {\color{black} Using the same method based on the Barron class \citep{barron1993universal}, Theorem 2.2 in \citet{yukich2002sup}, and the fact that four ReLU activation functions can form a total variation activation function that has a bounded support,} for an arbitrary positive $\varepsilon_2$, there exists \textit{q} one-hidden-layer MLPs ${\it{\Pi}}_j(\cdot)$ with $\sigma_{\mathrm{ReLU}}(\cdot)$, the parameters in each ${\it{\Pi}}_j(\cdot)$ with respect to $\mathbf{u}(0)$ and $t$ are zero, the widths of the input and output layers of each ${\it{\Pi}}_j(\cdot)$ are respectively $r + p + 1$ and 1, the width of the hidden layer for each ${\it{\Pi}}_j(\cdot)$ is at most
\begin{equation}
    {w_{{{\it{\Pi}} _j}}} \le 8\left\lceil {C_{{{\it{\Pi}}_j}}^2\varepsilon _2^{ - 2}} \right\rceil,
    \label{eq:b29}
\end{equation}
such that the difference between ${\it{\Pi}}_j\left[\mathbf{x},\mathbf{u}(0),t\right]$ and $h_j\left(\mathbf{x}\right)$ over $\mathbf{x} \in \Omega\left(B_{\mathbf{x}}^{r}\right)$ is bounded by
\begin{equation}
    \left| {{{\it{\Pi}} _j}\left[ {{\bf{x}},{\bf{u}}(0),t} \right] - {h_j}({\bf{x}})} \right| \le {\varepsilon _2}.
    \label{eq:b30}
\end{equation}
In Eq.~\eqref{eq:b29}, $C_{{{\it{\Pi}}_j}} = \left[\mathcal{H}_{{\tilde{h}}_j}\left(\boldsymbol{\upomega}\right),\sigma_{\mathrm{ReLU}}(\cdot),r, \Omega\left(B_{\mathbf{x}}^{r}\right)  \right]$ is a constant only depending on the Fourier transform $\mathcal{H}_{{\tilde{h}}_j}\left(\boldsymbol{\upomega}\right)$ of ${\tilde{h}}_j(\cdot)$ corresponding to ${{h}}_j(\cdot)$, the domain $\Omega\left(B_{\mathbf{x}}^{r}\right)$, the activation function $\sigma_{\mathrm{ReLU}}(\cdot)$, and the number \textit{r}. 

Then, for an arbitrary positive $\varepsilon_{2}$, there exists a one-hidden-layer MLP ${\it{\Pi}}(\cdot)$ consisting of above \textit{q} MLPs, ${\it{\Pi}}_1(\cdot),{\it{\Pi}}_2(\cdot)$ $,\dots,{\it{\Pi}}_q(\cdot)$, the widths of the input and output layers of ${\it{\Pi}}(\cdot)$ are respectively $r + p + 1$ and $q$, and the width of its hidden layer is at most
\begin{equation}
    {w_{{{\it{\Pi}}}}} \le 8q\left\lceil {C_{{{\it{\Pi}}}}^2\varepsilon _2^{ - 2}} \right\rceil,
    \label{eq:b31}
\end{equation}
where ${C_{\it{\Pi}} } = \max \left( {{C_{{{\it{\Pi}}_1}}},{C_{{{\it{\Pi}}_2}}}, \ldots ,{C_{{{\it{\Pi}}_q}}}} \right)$, such that for all $\mathbf{u}(t) \in K$, the corresponding solutions $\mathbf{x}(t)$ from Eq.~\eqref{eq:1}, whose ${\it{\Gamma}}(\cdot)$ satisfying Eqs.~\eqref{eq:b25} to \eqref{eq:b27}, with the initial conditions $\mathbf{x}(0) = {\mathbf{x}}'(0) = \mathbf{0}$, satisfy
\begin{equation}
    \left|{{\it{\Pi}} \left[ {{\bf{x}}(t),{\bf{u}}(0),t} \right] - h\left[ {{\bf{x}}(t)} \right]} \right| \le q{\varepsilon _2}
    \label{eq:b32}
\end{equation}
for all $t \in [0,T]$. Substituting Eqs.~\eqref{eq:b28} and \eqref{eq:b32} into Eq.~\eqref{eq:b19}, Eq.~\eqref{eq:30} is proven.
\hfill$\blacksquare$

\section{Explanation of the time dependence of \texorpdfstring{$\Psi\left({\mathbf{x}},{\mathbf{0}},t\right)$}{Ψ(\textbf{x}, \textbf{0}, \textit{t})}}
\label{appC}
Assume that $\mathbf{u}(0) = \mathbf{0}$ holds for all $\mathbf{u}(t) \in K$, then, the continuous mapping $\Psi \left( {\bf{x}},{\bf{0}},t \right)$ in Lemma 3 is now defined as $\Psi \left( {\mathbf{x}},{\mathbf{0}},t \right) = \Psi \left[ {\mathbf{x}}(t),{\bf{0}},t \right] = \Phi[{\hat{\mathbf{u}}}_t(\tau)](t)$, where the {\color{black} approximation} ${\hat{\mathbf{u}}}_t(\tau)$ of $\mathbf{u}(t)$ is calculated as $\hat{\mathbf{u}}_{t}(\tau) = \sum_{n = 1}^M {{\alpha _n}{{\bf{x}}_{n}}(t)\sin \left[ {{\omega _n}(t - \tau ) - {\theta _n}} \right]}$ and $\mathbf{x}_{n}(t) = \mathcal{L}_t\mathbf{u}(\omega_n)$ in Eq.~\eqref{eq:a17}. For $\forall\mathbf{u}_1(t) \in K$ with a positive $\Delta t$, its shifted version $\mathbf{u}_2(t)$ is 
\begin{equation}
    \mathbf{u}_2(t) = \left\{
    \begin{aligned}
    &\mathbf{u}_1(t-\Delta t),\;\;\; t \geq \Delta t\\ &\mathbf{0},\;\;\;\;\;\;\;\;\;\;\;\;\;\;\;\;\text{otherwise}
    \end{aligned}.\right.
    \label{eq:c1}
\end{equation}
For $\forall t_1 \geq 0$ and $t_2 = t_1 +\Delta t \leq T$, it is satisfied $\mathbf{x}_{1,n}(t_1) = \mathcal{L}_{t_1}\mathbf{u}_1(\omega_n) = \mathcal{L}_{t_2}\mathbf{u}_2(\omega_n) = \mathbf{x}_{2,n}(t_2)$, which is denoted as $\mathbf{x}_{1,n}(t_1) = \mathbf{x}_{2,n}(t_2) = \mathbf{x}_{n}$. Then, the {\color{black} approximations} ${\hat{\mathbf{u}}_{1,t_1}}(\tau)$ and ${\hat{\mathbf{u}}_{2,t_2}}(\tau)$ of $\mathbf{u}_1(t)$ and $\mathbf{u}_2(t)$ are respectively ${\hat{\mathbf{u}}_{1,t_1}}(\tau ) = \sum_{n = 1}^M {{\alpha _n}{{\mathbf{x}}_{n}}\sin \left[ {{\omega _n}(t_1 - \tau ) - {\theta _n}} \right]}$ and ${\hat{\mathbf{u}}_{2,t_2}}(\tau ) = \sum_{n = 1}^M {{\alpha _n}{{\bf{x}}_{n}}\sin \left[ {{\omega _n}(t_2 - \tau ) - {\theta _n}} \right]}$. The related $\Psi\left( {\mathbf{x}}_n,{\mathbf{0}},t_1 \right)$ is $\Psi\left({\mathbf{x}}_n,{\mathbf{0}},t_1 \right)$ $= \Phi\left[\sum_{n = 1}^M {{\alpha_n}{{\mathbf{x}}_{n}}\sin \left[ {{\omega_n}(t_1 - \tau ) - {\theta _n}} \right]}\right](t_1)$ and $\Psi \left( {\mathbf{x}}_n,{\mathbf{0}},t_2 \right)$ is $\Psi\left({\mathbf{x}}_n,{\mathbf{0}},t_2 \right) =\Phi\left[ \sum_{n = 1}^M {{\alpha_n}{{\mathbf{x}}_{n}}\sin \left[ {{\omega_n}(t_2 - \tau ) - {\theta _n}} \right]}\right](t_2)$. Although $\Phi$ is assumed to be time-autonomous, thereby $\Phi\left[\mathbf{u}_1(\tau)\right](t_1) = \Phi\left[\mathbf{u}_2(\tau)\right](t_2)$, $\Psi ({\mathbf{x}}_n,{\mathbf{0}},$ $t_1)$ does not equal $\Psi \left( {\mathbf{x}}_n,{\mathbf{0}},t_2 \right)$. This is because the valid range of $\tau$ in $\sin \left[ {{\omega _n}(t_1 - \tau ) - {\theta _n}} \right]$ is $[0,t_1]$ but that of $\sin \left[ {{\omega _n}(t_2 - \tau) - {\theta _n}} \right]$ is $[0,t_2]$. Thus, the time dependence of $\Psi \left({\mathbf{x}},{\mathbf{0}},t \right)$ cannot be ignored even if $\Phi$ is time-autonomous.

An alternative {\color{black} approximation} ${\hat{\mathbf{u}}_{2,t_2,\Delta t}}(\tau)$ of $\mathbf{u}_2(t)$ is
\begin{equation}
    {\hat{\mathbf{u}}_{2,t_2,\Delta t}}(\tau) = 
    \left\{
    \begin{aligned}
        &\sum\limits_{n = 1}^M {{\alpha _n}{{\bf{x}}_{n}}\sin \left[ {{\omega _n}(t_2 - \tau ) - {\theta _n}} \right]},\;\;\;\tau \in[\Delta t,t_2]\; \text{and}\; \Delta t \leq t_2 \leq T\\ &\mathbf{0},\;\;\;\;\;\;\;\;\;\;\;\;\;\;\;\;\;\;\;\;\;\;\;\;\;\;\;\;\;\;\;\;\;\;\;\;\;\;\;\;\;\;\;\;\;\;\;\;\tau \in\left[0,\Delta t\right)
    \end{aligned}.\right.
    \label{eq:c2}
\end{equation}
It is satisfied $\Phi\left[\hat{\mathbf{u}}_{1,t_1}(\tau) \right](t_1) = \Phi\left[\hat{\mathbf{u}}_{2,t_2,\Delta t}(\tau)\right](t_2)$. However, the {\color{black} approximation} in Eq.~\eqref{eq:c2} cannot apply to the input functions with non-zero values over $[0,\Delta t)$, and thus it is not applicable to all functions in \textit{K}.

\section{Proof of Eq.~\eqref{eq:27} with the Grönwall's inequality}
\label{appD}
{\bf Proof:} $\left|{\mathbf{z}}_2\left(t, {\mathbf{z}}_0\right) - {\mathbf{z}}_1\left(t, {\mathbf{z}}_0\right)\right|$, which are driven by the same $\mathbf{u}(t) \in K$ with the same initial condition ${\mathbf{z}}_0$, can be bounded by
\begin{equation}
    \begin{aligned}
    &\left| {\mathbf{z}}_2\left(t, {\mathbf{z}}_0\right) - {\mathbf{z}}_1\left(t, {\mathbf{z}}_0\right) \right|\\
    &\;\;\; = \left| {\int_0^t {{{{\bf{z}}}_2'}(\tau ){\rm{d}}\tau }  - \int_0^t {{{{\bf{z}}}_1'}(\tau ){\rm{d}}\tau } } \right|\\
    &\;\;\;= \left| {\int_0^t {{{\bf{z}}_{2,2}}(\tau ){\rm{d}}\tau }  - \int_0^t {{{\bf{z}}_{1,2}}(\tau ){\rm{d}}\tau } } \right| + \left| {\int_0^t {{g_1}\left[ {{{\bf{z}}_{1,1}}(\tau ),{{\bf{z}}_{1,2}}(\tau ),{\bf{u}}(\tau )} \right]} {\rm{d}}\tau  - \int_0^t {{g_2}\left[ {{{\bf{z}}_{2,1}}(\tau ),{{\bf{z}}_{2,2}}(\tau ),{\bf{u}}(\tau )} \right]{\rm{d}}\tau } } \right|\\
    &\;\;\; \le \int_0^t {\left| {{{\bf{z}}_{2,2}}(\tau ) - {{\bf{z}}_{1,2}}(\tau )} \right|{\rm{d}}\tau } + \int_0^t {\left| {{g_1}\left[ {{{\bf{z}}_{1,1}}(\tau ),{{\bf{z}}_{1,2}}(\tau ),{\bf{u}}(\tau )} \right] - {g_2}\left[ {{{\bf{z}}_{2,1}}(\tau ),{{\bf{z}}_{2,2}}(\tau ),{\bf{u}}(\tau )} \right]} \right|} {\rm{d}}\tau \\
    &\;\;\; \le \int_0^t {\left| {{{\bf{z}}_{2,2}}(\tau ) - {{\bf{z}}_{1,2}}(\tau )} \right|{\rm{d}}\tau } + \int_0^t {\left| {{g_1}\left[ {{{\bf{z}}_{1,1}}(\tau ),{{\bf{z}}_{1,2}}(\tau ),{\bf{u}}(\tau )} \right] - {g_1}\left[ {{{\bf{z}}_{2,1}}(\tau ),{{\bf{z}}_{1,2}}(\tau ),{\bf{u}}(\tau )} \right]} \right|} {\rm{d}}\tau \\
    &\;\;\;\;\;\;\; + \int_0^t {\left| {{g_1}\left[ {{{\bf{z}}_{2,1}}(\tau ),{{\bf{z}}_{1,2}}(\tau ),{\bf{u}}(\tau )} \right] - {g_1}\left[ {{{\bf{z}}_{2,1}}(\tau ),{{\bf{z}}_{2,2}}(\tau ),{\bf{u}}(\tau )} \right]} \right|} {\rm{d}}\tau \\
    &\;\;\;\;\;\;\; + \int_0^t {\left| {{g_1}\left[ {{{\bf{z}}_{2,1}}(\tau ),{{\bf{z}}_{2,2}}(\tau ),{\bf{u}}(\tau )} \right] - {g_2}\left[ {{{\bf{z}}_{2,1}}(\tau ),{{\bf{z}}_{2,2}}(\tau ),{\bf{u}}(\tau )} \right]} \right|} {\rm{d}}\tau.
    \end{aligned}
    \label{eq:d1}
\end{equation}
Since $g_1(\cdot)$ is Lipschitz continuous, $\left|{\mathbf{z}}_2\left(t, {\mathbf{z}}_0\right) - {\mathbf{z}}_1\left(t, {\mathbf{z}}_0\right)\right|$ can be bounded by
\begin{equation}
    \begin{aligned}
        &\left| {\mathbf{z}}_2\left(t, {\mathbf{z}}_0\right) - {\mathbf{z}}_1\left(t, {\mathbf{z}}_0\right) \right|\\
        &\;\;\; \le \int_0^t {\left| {{{\bf{z}}_{2,2}}(\tau ) - {{\bf{z}}_{1,2}}(\tau )} \right|{\rm{d}}\tau } + {L_{{g_1},1}}\int_0^t {\left| {{{\bf{z}}_{2,1}}(\tau ) - {{\bf{z}}_{1,1}}(\tau )} \right|} {\rm{d}}\tau  + {L_{{g_1},2}}\int_0^t {\left| {{{\bf{z}}_{1,2}}(\tau ) - {{\bf{z}}_{2,2}}(\tau )} \right|} {\rm{d}}\tau \\
        &\;\;\;\;\;\;\; + \int_0^t {\left| {{g_1}\left[ {{{\bf{z}}_{2,1}}(\tau ),{{\bf{z}}_{2,2}}(\tau ),{\bf{u}}(\tau )} \right] - {g_2}\left[ {{{\bf{z}}_{2,1}}(\tau ),{{\bf{z}}_{2,2}}(\tau ),{\bf{u}}(\tau )} \right]} \right|} {\rm{d}}\tau \\
        &\;\;\; \le {L_{{g_1}}}\int_0^t {\left| {{{\bf{z}}_1}(\tau ) - {{\bf{z}}_2}(\tau )} \right|} {\rm{d}}\tau + \mathop {\max }\limits_{\tau  \in [0,T]} \left| {{g_1}\left[ {{{\bf{z}}_{2,1}}(\tau ),{{\bf{z}}_{2,2}}(\tau ),{\bf{u}}(\tau )} \right] - {g_2}\left[ {{{\bf{z}}_{2,1}}(\tau ),{{\bf{z}}_{2,2}}(\tau ),{\bf{u}}(\tau )} \right]} \right|T,\\
        &\;\;\;\mathop  \le \limits^{(a)} {e^{T{L_{{g_1}}}}}T\mathop {\max }\limits_{\tau  \in [0,T]} \left| {{g_1}\left[ {{{\bf{z}}_{2,1}}(\tau ),{{\bf{z}}_{2,2}}(\tau ),{\bf{u}}(\tau )} \right] - {g_2}\left[ {{{\bf{z}}_{2,1}}(\tau ),{{\bf{z}}_{2,2}}(\tau ),{\bf{u}}(\tau )} \right]} \right|,
    \end{aligned}
    \label{eq:d2}
\end{equation}
where $L_{g_1} = \max\left[L_{g_1,1},\left(L_{g_1,2} +1\right)\right]$, $L_{g_1,1}$ and $L_{g_1,2}$ are respectively the two Lipschitz constants of $g_1(\cdot)$ with respect to its first and second arguments, and Step (\textit{a}) holds from the Grönwall's inequality \citep{ames1997inequalities}. Eq.~\eqref{eq:d2} is the same as Eq.~\eqref{eq:27}.

\section{Approach for Training the Neural Oscillator}
\label{appE}
By introducing the state space vector $\mathbf{z}(t) = \left[\mathbf{z}_1^\top(t),\mathbf{z}_2^\top(t)\right]^\top = \left[\mathbf{x}^\top(t),\mathbf{x}'^\top(t)\right]^\top$, Eq.~\eqref{eq:1} over the time interval $\left[0,T\right]$ can be rewritten in the state function form
\begin{equation}
    \left\{
    \begin{aligned}
        &\mathbf{z}_1'(t) = \mathbf{z}_2(t)\\
        &\mathbf{z}_2'(t) = \itGamma\left[\mathbf{z}_1(t),\mathbf{z}_2(t),\mathbf{u}(t)\right]\\
        &\mathbf{y}(t) = \itPi\left[\mathbf{z}_1(t),\mathbf{u}(0),t\right]
    \end{aligned}.\right.
    \label{eq:e1}
\end{equation}
Utilizing a second-order Runge-Kutta scheme \citep{griffiths2010numerical}, a time-discretization form of Eq.~\eqref{eq:e1} is
\begin{equation}
    \left\{
    \begin{aligned}
        &\mathbf{y}(t_{i+1}) = \itPi\left[\mathbf{z}_1(t_{i+1}),\mathbf{u}(0),t_{i+1}\right]\\
        &\mathbf{z}(t_{i+1}) = 0.5\Delta{t_i}\left(\mathbf{k}_2+\mathbf{k}_1\right)\\
        &\mathbf{k}_2 =
        \begin{Bmatrix}
            \mathbf{z}_2(t_i)+\Delta{t_i}\mathbf{k}_{12}\\
            \itGamma\left[\mathbf{z}_1(t_i)+\Delta{t_i}\mathbf{k}_{11},\mathbf{z}_2(t_i)+\Delta{t_i}\mathbf{k}_{12},\mathbf{u}(t_{i+1})\right]
        \end{Bmatrix}\\
        &\mathbf{k}_1=
        \begin{bmatrix}
            \mathbf{k}_{11}\\
            \mathbf{k}_{12}
        \end{bmatrix}=
        \begin{Bmatrix}
            \mathbf{z}_2(t_i)\\
            \itGamma\left[\mathbf{z}_1(t_i),\mathbf{z}_2(t_i),\mathbf{u}(t_i)\right]
        \end{Bmatrix}
    \end{aligned},\right.
    \label{eq:e2}
\end{equation}
where $\mathbf{z}(0) = \mathbf{0}$, $\mathbf{y}(0)=\itPi\left[\mathbf{z}_1(0),\mathbf{u}(0),t\right]$, $\Delta{t_i} = t_{i+1} - t_i$, $0 = t_0 < t_1 < t_2 < \dots < t_I = T$. Given available output function samples {\color{black} $\hat{\mathbf{y}}_l(t_i) = \left[\hat{y}_{l,1}(t_i),\hat{y}_{l,2}(t_i),\dots,\hat{y}_{l,q}(t_i)\right]^\top$} caused by the input samples $\mathbf{u}_l(t_i)$, $l = 1, 2, \dots, L$, the {\color{black} $r^{\text{th}}$} power loss function $\ell_r$ is used to learn the trainable parameters of the MLPs $\itGamma(\cdot)$ and $\itPi(\cdot)$ in Eq.~\eqref{eq:e1},
\begin{equation}
    \ell_r = \frac{1}{LIq}{\sum\limits_{l = 1}^L {\sum\limits_{i = 0}^{\color{black}I-1}\sum\limits_{j = 1}^q {{\left|{y_{l,j}}(t_i) - \hat{y}_{l,j}(t_i)\right|}^r}}},
    \label{eq:e3}
\end{equation}
where $\mathbf{y}_l(t_i) = \left[y_{l,1}(t_i),y_{l,2}(t_i),\dots,y_{l,q}(t_i)\right]^\top$ represents the output computed from the neural oscillator driven by $\mathbf{u}_l(t_i)$ using Eq.~\eqref{eq:e2}. In the numerical cases of this study, the Adam method \citep{kingma2014adam} implemented using the PyTorch \citep{paszke2019pytorch} is utilized to train the neural oscillator using the Runge-Kutta scheme in Eq.~\eqref{eq:e2} and the loss function $\ell_r$ in Eq.~\eqref{eq:e3}.

\bibliographystyle{elsarticle-harv} 
\bibliography{references}




% %% For authoryear reference style

% \bibitem[Lamport(1994)]{lamport94}
%   \textit{\LaTeX: a document preparation system},
%   Addison Wesley, Massachusetts,

% \end{thebibliography}
\end{document}